\documentclass[a4paper,11pt]{article}

\input{macros.tex}

\hypersetup{
  pdftitle={Byzantine-Robust Muon: Resilient Momentum Orthogonalization for Distributed Learning},
  pdfauthor={Grigory Malinovsky, Yury Demidovich, Peter Richt\'arik}}

\title{
\toptitlebar
{{\center\baselineskip 18pt
                      {\Large\bf Byzantine-Robust Muon: Resilient Momentum Orthogonalization for Distributed Learning}}
}
\bottomtitlebar}
\date{}
\author{
Grigory Malinovsky,
Yury Demidovich,
Peter Richt\'arik
}
\affil{
King Abdullah University of Science and Technology (KAUST), Saudi Arabia
}

\begin{document}

\maketitle

\def\thefootnote{\arabic{footnote}}

\begin{abstract}
Optimization methods, including the Muon optimizer based on the Linear Minimization Oracle (LMO) and momentum orthogonalization, have shown strong empirical performance in large-scale and distributed learning, yet their robustness to adversarial behavior and data heterogeneity is not well understood. We study these methods within a unified first-order trust-region framework and analyze Muon-based updates with robust aggregation for Byzantine-robust distributed and federated learning. Our analysis does not rely on the i.i.d. assumption and introduces a non-Euclidean notion of heterogeneity consistent with the trust-region geometry. Under this framework, we establish convergence guarantees for both general non-convex and star-convex objectives. Our results provide a theoretical characterization of Byzantine-robust Muon-based optimization, with an additive heterogeneity error floor consistent with known impossibility results, under explicit aggregation assumptions. Under an explicit additive oracle-error condition, the guarantees also cover inexact orthogonalization, with the non-convex Byzantine terms in the bound's numerator unchanged but a smaller progress denominator. In simulated federated training with ResNet-18 and ViT-Tiny on CIFAR-10, additional image benchmarks, and a 38M-parameter byte-level language model on TinyStories, the tested Muon configuration attains higher attacked accuracy (lower perplexity) than the tested SGD and normalized-SGD configurations, and than sign updates on CIFAR-10.
 Fixed-honest controls further show a substantial RFA advantage over averaging under a bounded RFA-query attack, higher attacked accuracy for Muon-NS than tensorwise Euclidean normalization from shared checkpoints, and lower held-out TinyStories perplexity than a tuned AdamW baseline.

\end{abstract}

\section{Introduction}

Deep learning has become a dominant paradigm in modern machine learning \citep{lecun2015deep}, particularly with the remarkable success of large language models (LLMs) \citep{zhao2023survey}. Training such models requires substantial computational and memory resources, so optimization methods play a central role: they can significantly accelerate training and improve stability \citep{le2011optimization}. In practice, adaptive methods, most notably Adam \citep{kingma2014adam} and AdamW \citep{loshchilov2017decoupled}, have become the standard choice due to their fast and reliable convergence across a wide range of architectures \citep{achiam2023gpt,  liu2024deepseek, grattafiori2024llama, crawshaw2025exploration}.

Recently, a new class of optimizers, such as Muon \citep{jordan6muon} and Scion \citep{pethick2025training}, has been proposed, leveraging geometry-aware adaptation via momentum orthogonalization, with notable improvements on large language model (LLM) tasks \citep{liu2025muon}. At its core, Muon can be viewed as stochastic spectral descent (SSD) \citep{carlson2015stochastic, carlson2015stochastic1, carlson2015preconditioned} --- steepest descent with respect to the spectral norm \citep{bernstein2024old} --- augmented with momentum, where the singular value decomposition required for exact spectral descent is replaced by efficient Newton--Schulz iterations \citep{kovarik1970some, bjorck1971iterative, bernstein2025deriving, amsel2025polar, shulgin2025beyond, pethick2025training}.

From a theoretical perspective, the Muon optimizer admits several interpretations. One line of work connects Muon to linear minimization oracle (LMO) methods, such as Frank--Wolfe \citep{jaggi2013revisiting}: \citet{pethick2025training} propose a general framework for norm-constrained optimization with a unified analysis of non-Euclidean stochastic gradient descent with momentum under arbitrary norms, recovering SignSGD \citep{bernstein2018signsgd}, Signum \citep{sun2023momentum}, and Muon, with clear connections to adaptive methods; subsequent works extend the framework to gradient clipping and generalized smoothness \citep{pethick2025generalized, riabinin2025gluon}.

A related interpretation and generalization is proposed in \citet{kovalev2025understanding}, which analyzes Muon within the framework of non-Euclidean trust-region methods \citep{demidovich2026distance}, explains why orthogonalizing the momentum rather than the raw gradient \citep{tuddenham2022orthogonalising} is beneficial, recovers normalized SGD with momentum \citep{cutkosky2020momentum}, SignSGD, Signum, and Muon, and naturally incorporates weight decay; subsequent work demonstrates advantages of non-Euclidean optimization arising from the structure of stochastic gradient noise \citep{kovalev2025non}. Recent works \citep{xie2025structured, khirirat2025better, qian2025muon} develop variance-reduced variants along with explicit connections between Muon and adaptive methods \citep{gupta2017unified} such as ASGO/One-sided Shampoo \citep{an2025asgo} and AdaGrad \citep{streeter2010less, li2019convergence}.


The core idea of Muon has been widely adopted and adapted \citep{gruntkowska2025non}, e.g., in MuonBP for model-parallel training \citep{khaled2025muonbp} and in Drop-Muon, which updates only a random subset of layers per step \citep{gruntkowska2025drop}. One of the most significant application areas is distributed and federated learning, where multiple workers or clients collaboratively optimize a global model \citep{kairouz2021advances} and several Muon-type methods improve convergence speed, communication efficiency, and scalability \citep{therien2025muloco, ahn2025dion, gruntkowska2025error, qian2026communication}. Despite these advances, existing approaches largely overlook robustness to workers that behave maliciously or send adversarial updates \citep{sun2019can}, which can significantly degrade the performance of distributed or federated optimization \citep{he2020secure}.

Distributed and federated learning are commonly viewed as forms of collaborative training whose objective is to learn a single shared model \citep{mammen2021federated, diskin2021distributed}. In these settings, however, the learning process is vulnerable to \emph{Byzantine} workers \citep{lamport2019byzantine, su2016fault} --- participants that deviate from the prescribed protocol due to malicious intent or system failures, transmitting corrupted information such as adversarial gradients to slow down or completely derail training \citep{baruch2019little}; they may collude and possess partial knowledge of other users' private data. Because robustness is typically required in the worst-case sense, much of the literature places no restrictions on the adversary's capabilities \citep{gorbunov2022variance, malinovsky2024byzantine}. Under these conditions, methods relying on simple aggregation, such as averaging stochastic gradients, are fundamentally fragile: even a single Byzantine worker can inject an arbitrarily large update and drive the iterates arbitrarily far from the desired solution \citep{bagdasaryan2020backdoor, wang2020attack}.

Consequently, such settings require secure, \emph{Byzantine-robust} optimization methods \citep{alistarh2018byzantine, allen2020byzantine}. Designing methods with provable guarantees is nontrivial: stochastic gradients of honest workers are inherently diverse due to randomness and data heterogeneity \citep{zhu2021broadcast, allouah2023fixing}, which makes benign updates hard to distinguish from malicious ones, while adversaries can craft messages resembling legitimate gradients or inject small, temporally correlated perturbations \citep{xie2020fall}. As demonstrated by \citet{baruch2019little, karimireddy2021learning}, such attacks bypass many widely used defenses combining robust aggregation with distributed SGD \citep{blanchard2017machine, damaskinos2019aggregathor, yin2018byzantine, Pillutla2022}. Moreover, for broad classes of heterogeneous problems, achieving arbitrary accuracy is fundamentally impossible under Byzantine failures \citep{karimireddy2021learning, Karimireddy2020ByzantineRobustLO}.

The lower bounds of \citet{karimireddy2021learning} demonstrate limitations of memoryless, permutation-invariant aggregation-based optimization under heterogeneous Byzantine data. These results motivate exploiting temporal structure, such as worker momentum or error correction \citep{islamov2026byzantine}; our guarantees are established under the explicit assumptions given below rather than claiming that all memoryless algorithms fail on every problem.

Robust aggregation of worker momentum followed by a normalized server update predates our work: \citet{yang2024effect} analyze the Euclidean method with i.i.d. workers and study batch-size effects. \citet{bolatov2026byzantine} extend the Euclidean analysis to heterogeneous data under $(L_0,L_1)$-smoothness and explicitly identify an arbitrary-norm LMO, including operator-norm Byzantine-robust Muon, as a possible extension. Our contribution is a non-Euclidean convergence analysis under relative first-moment heterogeneity, including non-convex and star-convex guarantees and extensions to extrapolation and inexact LMO steps.
For fixed good-worker messages $\{z_i\}_{i\in\mathcal G}$ in the same dual norm, let $G=|\mathcal G|\ge2$, $\bar z=G^{-1}\sum_{i\in\mathcal G}z_i$, and define
\begin{align*}
D_G &:= \frac{1}{G}\sum_{i\in\mathcal G}\|z_i-\bar z\|_*,\qquad
r_G := \frac{1}{G(G-1)}\sum_{\substack{i,l\in\mathcal G\\i\ne l}}\|z_i-z_l\|_*,\\
\frac12 r_G &\le D_G \le \frac{G-1}{G}r_G.
\end{align*}
The last inequality follows from Jensen's inequality and the triangle inequality and also holds after taking expectations. The pairwise dispersion in Definition~\ref{def:aragg} and the mean deviation used by \citet{bolatov2026byzantine} are thus equivalent up to constants when measured in the same norm; comparing different norms introduces the corresponding norm-equivalence factors. Table~\ref{tab:prior-comparison} summarizes the distinct proved guarantees.

\begin{table}[H]
\centering\small
\caption{Comparison of the main theoretical guarantees for robustly aggregated worker momentum. The arbitrary-norm LMO of \citet{bolatov2026byzantine} is discussed as an extension, rather than covered by their theorem.}\label{tab:prior-comparison}
\setlength{\tabcolsep}{2pt}
\begin{tabular}{@{}llllll@{}}
\toprule
Work & \makecell[l]{Geometry and\\smoothness} & Heterogeneity & \makecell[l]{Aggregation\\premise} & \makecell[l]{Update\\oracle} & \makecell[l]{Convergence\\criterion} \\
\midrule
\citet{yang2024effect} & \makecell[l]{Euclidean\\$L$-smooth} & i.i.d. & \makecell[l]{Euclidean\\second moment} & \makecell[l]{exact\\normalization} & \makecell[l]{average\\$\mathbb E\|\nabla f\|_2$} \\
\addlinespace[2pt]
\citet{bolatov2026byzantine} & \makecell[l]{Euclidean\\$(L_0,L_1)$-smooth} & \makecell[l]{additive\\second moment} & \makecell[l]{Euclidean\\first moment} & \makecell[l]{exact\\normalization} & \makecell[l]{minimum\\$\mathbb E\|\nabla f\|_2$} \\
\addlinespace[2pt]
This work & \makecell[l]{arbitrary norm\\$L$-smooth} & \makecell[l]{relative/additive\\first moment} & \makecell[l]{dual-norm\\first moment} & \makecell[l]{exact/inexact\\LMO} & \makecell[l]{average $\mathbb E\|\nabla f\|_*$\\star-convex gap} \\
\bottomrule
\end{tabular}
\end{table}

\section{Contribution}
We make the following contributions.
\begin{itemize}
    \item We study \textbf{Byzantine-Robust Muon}, a distributed Muon-type method using robust aggregation in place of standard averaging while preserving the momentum-orthogonalization mechanism of Muon. The method maintains worker-level momentum buffers, following the worker-momentum design of \citet{karimireddy2021learning,farhadkhani2022byzantine}, and therefore uses temporal information that is absent from memoryless permutation-invariant schemes.
    \item We formulate the method in a non-Euclidean trust-region framework, which separates the robust aggregation step from the geometry of the update, supports arbitrary norms, and specializes to Muon under the operator/nuclear-norm geometry; other norm choices recover normalized SGD- and sign-type momentum methods.
    \item We prove convergence guarantees under Byzantine workers and heterogeneous data: a bound on the average gradient dual norm for general non-convex objectives, and a linear contraction to a stochastic and Byzantine noise neighborhood for star-convex objectives with weight decay, under a non-Euclidean heterogeneity condition aligned with the trust-region geometry.
    \item We extend both guarantees under a separate second-moment aggregation premise, with concrete bucketed rules subject to their stated conditions, improve the non-convex rate under second-order smoothness via gradient extrapolation ($\mathcal{O}(\varepsilon^{-3.5})$ at $\delta = 0$; the star-convex counterpart is discussed in Remark~\ref{rem:second-order}), and extend all results to an \emph{inexact} Muon update satisfying an additive oracle-error condition with parameter $\bar{\delta}$ \citep{shulgin2025beyond}: progress deflates by $(1-\bar{\delta})$ while the non-convex Byzantine terms in the numerator remain unchanged (Theorem~\ref{thm:main-inexact}).
    \item We evaluate the same robust aggregation protocol with ResNet-18 and ViT-Tiny on CIFAR-10, a 38M-parameter byte-level language model on TinyStories, and additional image benchmarks. Five-seed comparisons on CIFAR-10 (five attacks) and TinyStories (four attacks), and three-seed comparisons on the transfer benchmarks (three attacks), cover clean and attacked training; on CIFAR-10 ResNet-18 we additionally ablate the Byzantine fraction, heterogeneity, worker momentum weight, aggregator, bucketing, and nearest-neighbor mixing (Appendix~\ref{apx:experiment-details}).
\end{itemize}

\section{Preliminaries}
This section presents the formal problem setup, along with the main definitions and assumptions used throughout the analysis. We consider the following distributed finite-sum optimization problem:

\begin{equation}
\label{eq:main}
\min_{x\in\mathcal{X}}\left[f(x) := \frac{1}{G} \sum_{i \in \mathcal{G}} \{f_i(x) := \mathbb{E}_{\xi_i}[F_i(x; \xi_i)]\}\right],
\end{equation}
where $\mathcal{X}$ is a finite-dimensional vector space endowed with the inner product $\langle\cdot,\cdot\rangle: \mathcal{X}\times\mathcal{X}\to\mathbb{R},$ $f$ is a differentiable loss function bounded from below, with $f_{\inf} := \inf_{x\in\mathcal{X}} f(x) > -\infty$, and $\mathcal{G}$ denotes the set of regular clients, with cardinality $G := |\mathcal{G}|$. We assume that the set of all participating clients is $[n] := \{1,2,\ldots,n\}$, with $\mathcal{G} \subseteq [n]$.
The complementary set $\mathcal{B} := [n] \setminus \mathcal{G}$ consists of Byzantine clients. Let $|\mathcal{B}| = \delta_{\mathrm{real}} n$, where $\delta_{\mathrm{real}}$ denotes the true fraction of Byzantine clients and $\delta$ is a known upper bound such that $|\mathcal{B}| \le \delta n$; we assume throughout that at least two clients are regular, $G \ge 2$. (We reserve the symbol $B$ for the relative heterogeneity constant of Assumption~\ref{asn:bounded-heterogeneity} below.)
We further assume that $0 \le \delta_{\mathrm{real}} \le \delta < \tfrac{1}{2}$, since if Byzantine clients form a majority, the problem becomes unsolvable in general. In the context of distributed learning, $f_i : \mathcal{X} \to \mathbb{R}$ represents the loss function associated with the data of a regular client~$i$.






\paragraph{Robust aggregation.} One of the central components of our approach is the concept of a $(\delta, c)$-robust aggregator. Because our analysis measures all momentum errors by \emph{first moments} of the \emph{dual norm} $\|\cdot\|_*$, we state the aggregation property directly in that geometry. It is a first-moment variant of the definitions proposed by \citet{gorbunov2022variance}, which extend and generalize earlier notions introduced by \citet{karimireddy2021learning, Karimireddy2020ByzantineRobustLO}.

\begin{definition}[($\delta, c$)-Robust Aggregator]\label{def:aragg}
Let $\delta < \nicefrac{1}{2}$ and $c \ge 0$. Assume that $\{z_1, \dots, z_n\} \subset \mathcal{X}$ are random vectors such that there exists a subset $\mathcal{G} \subseteq [n]$ of size $G := |\mathcal{G}| \geq (1 - \delta)n$ whose average pairwise dispersion is bounded as
\begin{equation}
\frac{1}{G(G-1)} \sum_{\substack{i,l \in \mathcal{G}\\ i \neq l}} \mathbb{E}\|z_i - z_l\|_* \le r
\end{equation}
for some $r \ge 0$. We say that $\hat{z}$ is a $(\delta, c)$-Robust Aggregator and write $\hat{z} = \mathrm{ARAgg}(z_1, \dots, z_n)$ if
\begin{equation}
\mathbb{E} \|\hat{z} - \bar{z}\|_* \le c\delta\, r,
\qquad \text{where } \bar{z} = \frac{1}{G} \sum_{i \in \mathcal{G}} z_i,
\end{equation}
and $\hat{z}$ is computed \textit{without knowledge of the dispersion level $r$} (the aggregator is \emph{agnostic}).
\end{definition}
The definition asks the aggregator to approximate the ideal good-worker average (unattainable exactly, since Byzantine identities are unknown) with an error proportional to the dispersion of the honest inputs; the input $\delta$ must satisfy $\delta_{\mathrm{real}} \le \delta < \nicefrac{1}{2}$. The product $c\delta$ is best read as a single aggregation-quality factor: in this first-moment form it cannot in general be made small (worst case $c\delta \ge \nicefrac{1}{2}$; Remark~\ref{rem:aggregators}), unlike the second-moment setting of \citet{karimireddy2021learning}, where error of order $\delta$ times the squared dispersion is achievable and optimal. Under the corruption, bucket-size, and dimension conditions of their respective results, bucketed Krum, geometric median, and coordinate-wise median provide second-moment Euclidean aggregation guarantees \citep{Karimireddy2020ByzantineRobustLO, gorbunov2022variance}. Centered clipping requires a dispersion-dependent radius in the cited analysis and is not an agnostic realization of this premise. The second-moment property is distinct from Definition~\ref{def:aragg}: Jensen and Assumption~\ref{asn:norm-equivalence} convert its output error to the dual norm, while Appendix~\ref{apx:second-moment} derives separate guarantees under second-moment heterogeneity.

\noindent\textbf{Non-Euclidean Trust Region Framework.}
We use the first-order non-Euclidean trust-region viewpoint of~\citet{kovalev2025understanding}. Given a norm $\|\cdot\|$ with dual norm $\|\cdot\|_*$ and a radius $\eta>0$, the update generated by a model vector $v_k$ is
\[
    x^{k+1}\in
    \arg\min_{x\in\mathcal{X}}
    \left\{ \langle v_k,x\rangle:
    \|x-x^k\|\le \eta \right\}.
\]
This is a first-order trust-region step: rather than minimizing a quadratic majorizer, the method minimizes a linear model over a norm ball. The geometry of the norm determines the steepest-descent direction, while the scalar radius $\eta$ controls the step length. In the Euclidean case, choosing $v_k=\nabla f(x^k)$ recovers normalized gradient descent~\citep{shor2012minimization,grimmer2019convergence}.

The same template becomes Muon when applied to matrix-valued parameters with the operator norm as the primal norm and the nuclear norm as the dual norm: the steepest-descent direction is the polar/orthogonal factor solving the associated linear minimization oracle, which Muon applies to a momentum buffer and approximates by Newton--Schulz iterations~\citep{jordan6muon,pethick2025training,kovalev2025understanding}. The trust-region formulation thus covers normalized SGD, sign-based methods, and Muon-style momentum orthogonalization in a single framework, which is particularly useful in Byzantine settings: robust aggregation controls the submitted momentum vectors, whereas the trust-region subproblem controls the geometry and magnitude of the final server update.


\paragraph{Assumptions.} Throughout the paper we employ several standard assumptions. In what follows, $\mathcal{F}^{k}$ denotes the $\sigma$-algebra generated by the first $k+1$ stochastic gradient draws $\{\xi_i^{s}\}_{i \in \mathcal{G}, s \le k}$ of the algorithms of Section~\ref{sec:method}, together with the Byzantine messages and aggregator randomness through $m^{k}$ and the selection randomness of the trust-region solver through $x^{k+1}$; the round-$(k+1)$ gradient draws are \emph{not} $\mathcal{F}^{k}$-measurable. We set $\mathcal{F}^{-1}$ trivial; in particular, $x^{k+1}$ is $\mathcal{F}^{k}$-measurable.
\begin{assumption}[Stochastic gradient estimator]\label{asn:stoch-grad}
    For any good worker $i\in \mathcal{G},$ let $g_i(x; \xi)$ be a stochastic gradient estimator, and abbreviate $g_i^k=g_i(x^k;\xi_i^k)$. We assume conditional unbiasedness and bounded variance for all $k\geq 0$, including initialization:
\begin{align*}
\mathbb E[g_i^k\mid\mathcal F^{k-1}]&=\nabla f_i(x^k),\\
\mathbb E[\|g_i^k-\nabla f_i(x^k)\|_2^2\mid\mathcal F^{k-1}]&\leq\sigma^2.
\end{align*}
where $\sigma > 0$ is a positive variance parameter and $\|\cdot\|_2$ is the standard Euclidean norm; moreover, the noises $\{g_i(x^{k}; \xi_i^{k}) - \nabla f_i(x^{k})\}_{i \in \mathcal{G}}$ are conditionally independent across good workers given $\mathcal{F}^{k-1}$.
\end{assumption}
These standard assumptions are widely used to analyze stochastic gradient-based algorithms~\citep{ghadimi2013, ghadimi2016, cutkosky2020momentum, sun2023momentum, demidovich2023guide}
\begin{assumption}[Non-Euclidean norm and Lipschitz continuous gradient]\label{asn:lipschitz}
    We assume that vector space $\mathcal{X}$ is equipped with a norm $\|\cdot\|: \mathcal{X} \rightarrow \mathbb{R}_{+}$, which possibly does not coincide with the Euclidean norm $\|\cdot\|_2$. In addition, we assume that, for any good worker $i\in \mathcal{G},$ the gradient $\nabla f_i(\cdot)$ is Lipschitz continuous with respect to the norm $\|\cdot\|:$
    \begin{equation*}
    \|\nabla f_i(x) - \nabla f_i(x')\|_* \leq L\|x - x'\|, \quad \forall\; x, x' \in \mathcal{X},
    \end{equation*}
    where $L > 0$ is the gradient Lipschitz constant, and $\|\cdot\|_*:$ $\mathcal{X} \rightarrow \mathbb{R}_{+}$ is the dual norm associated with $\|\cdot\|$, i.e., $\|x\|_* = \sup_{\|x'\| \leq 1} \langle x, x' \rangle$ for all $x \in \mathcal{X}.$
\end{assumption}
Assuming the gradient is Lipschitz continuous is standard practice, consistent with prior work on first-order methods~\citep{ghadimi2013, cutkosky2020momentum, gorbunov-sgd-2020}. By the triangle inequality, $f = \frac{1}{G}\sum_{i\in\mathcal{G}} f_i$ inherits the same property, and the standard integral argument with the pairing $\langle a, b\rangle \le \|a\|_*\|b\|$ then yields the two-sided quadratic bounds $|f(y) - f(x) - \langle \nabla f(x), y - x\rangle| \le \frac{L}{2}\|y-x\|^2$, used repeatedly in the proofs.
\begin{assumption}[Norm equivalence]\label{asn:norm-equivalence}
    For a finite dimensional space $\mathcal{X},$ there exists a constant $\kappa \ge 1,$ such that $\dnorm{z}\le\kappa\norm{z}_2,$ for all $z\in\mathcal{X}.$
\end{assumption}
Since the space is finite-dimensional, the norm equivalence theorem guarantees the existence of such a constant. For instance, for matrix parameters $\mathcal{X} = \mathbb{R}^{d_1 \times d_2}$ with the operator norm as the primal norm, the dual norm is the nuclear norm and one may take $\kappa = \sqrt{\min\{d_1, d_2\}}$. We reserve the symbol $\kappa$ exclusively for this constant.
For the tensorwise Muon implementation in Section~\ref{sec:experiments}, let $U_b$ denote matrix blocks (with convolutional tensors reshaped to matrices) and $u_j$ one-dimensional parameter blocks. The corresponding product norm and its dual are
\[
\|(U,u)\| = \max\!\left\{\max_b\|U_b\|_{\mathrm{op}},\;\max_j\|u_j\|_2\right\},
\qquad
\|(Z,z)\|_* = \sum_b\|Z_b\|_{\mathrm{nuc}}+\sum_j\|z_j\|_2.
\]
Writing $r_b$ for the smaller dimension of matrix block $b$ and $J$ for the number of one-dimensional blocks, Assumption~\ref{asn:norm-equivalence} holds with $\kappa=\sqrt{\sum_b r_b+J}$ by the nuclear/Frobenius inequality and Cauchy--Schwarz. The exact LMO over this product ball applies the negative polar factor to each nonzero matrix block and the negative normalized direction to each nonzero vector block; a zero block may receive a zero direction. The five Newton--Schulz iterations used in the experiments approximate this LMO, but do not by themselves certify the oracle-error condition of Section~\ref{sec:inexact-main}.
\begin{assumption}[Star-convexity]
\label{asn:star-cvx}
A function $f(x)$ is star-convex if $f(\beta x^* + (1 - \beta)x) \leq \beta f(x^*) + (1 - \beta)f(x)$ for all $x \in \mathcal{X}$ and for all $\beta \in [0, 1];$ $x^*\in \mathcal{X}$ is a solution of problem~\eqref{eq:main}.
\end{assumption}
Controlling data heterogeneity is essential for Byzantine-robust analysis: without any restriction, it is impossible to distinguish a malicious update from an honest one computed on different data, so guarantees cannot be established in fully arbitrary heterogeneous environments. A common approach is to bound the average deviation of local gradients from the global gradient \citep{wu2020federated, gorbunov2022variance, allouah2023privacy}; in the non-Euclidean setting, such assumptions are naturally formulated in the dual norm of the chosen geometry.

\begin{assumption}[$(B,\zeta)$-heterogeneity]
\label{asn:bounded-heterogeneity}
There exist constants $\zeta \ge 0$ and $B \ge 0$ such that for all $x \in \mathcal{X}$, the gradients of the good workers $i\in \mathcal{G}$ satisfy:
\[
\frac{1}{G}\sum_{i\in \mathcal{G}}\|\nabla f_i(x) - \nabla f(x)\|_* \le  B\|\nabla f(x)\|_* +\zeta.
\]
\end{assumption}
The condition $4c\delta B < 1$ imposed by our non-convex theorem constrains only the \emph{relative} constant $B$; the absolute component $\zeta$ is unconstrained, and the two trade off against each other, so the condition can be met on any bounded region of interest by decreasing $B$ and compensating with a larger $\zeta$. We caution that on an unbounded domain this trade-off is not available in general (a finite $\zeta$ cannot absorb unbounded gradient disagreement), so Assumption~\ref{asn:bounded-heterogeneity} with a finite pair $(B, \zeta)$ is a genuine restriction; for the weight-decay variant, however, the iterates provably remain in the bounded set $\{x : \|x\| \le \max\{\|x^0\|, \eta/\beta\}\}$ (Lemma~\ref{lem:aux-inequalities}), where the trade-off applies.

\section{Robust Non-Euclidean Optimization}
\label{sec:method}
In this section, we investigate the optimization problem in heterogeneous environments with Byzantine workers, assuming access to a generic robust aggregator.

\subsection{Algorithm Description}
We propose \textbf{Byzantine-Robust Muon} (Algorithm~\ref{alg:byz-muon}). The method addresses two distinct sources of error. Robust aggregation limits the influence of Byzantine clients, with an aggregation error controlled by the dispersion of honest workers' updates, while local momentum reduces stochastic gradient variance. Each honest worker updates a local momentum buffer
$ m_i^{k+1}=(1-\alpha)m_i^k+\alpha g_i^{k+1}$ and sends this buffer, rather than a raw stochastic gradient, to the server; Byzantine workers may send arbitrary vectors at every round. The server applies an agnostic robust aggregator to the submitted buffers and then performs a non-Euclidean trust-region step using the aggregated momentum. Momenta are initialized from a single stochastic gradient at $x^0$ (robustly aggregated on the server), which the analysis uses.

A key feature of the algorithm is that robustness and geometry are modular. The aggregation rule determines how the server estimates a reliable momentum vector in the presence of adversarial messages, whereas the trust-region subproblem determines the geometry of the actual update. By setting $\mathcal{X}=\mathbb{R}^{d\times n}$, $\|\cdot\|=\|\cdot\|_{\mathrm{op}}$, and $\|\cdot\|_*=\|\cdot\|_{\mathrm{nuc}}$, the update specializes to the Muon geometry. Other norm choices recover normalized SGD or sign-type methods with momentum. This separation is useful because the same robust aggregation analysis can be paired with different non-Euclidean update geometries.

\begin{algorithm}[t]
\caption{Byzantine-Robust Muon}
\label{alg:byz-muon}
\begin{algorithmic}[1]
    \STATE {\bfseries Input:} $(\delta,c)$-ARAgg, trust-region radius $\eta > 0$, worker momentum weight $\alpha \in (0,1]$, initial point $x^0 \in \mathcal{X}$
    \STATE Each good worker $i \in \mathcal{G}$ initializes $m_i^{0} \leftarrow g_i(x^0;\,\xi_i^{0})$; each Byzantine worker $i \in \mathcal{B}$ sends an arbitrary vector $m_i^0$
    \STATE Server computes $m^{0} \leftarrow \mathrm{ARAGG}(m_1^{0},\dots,m_n^{0})$
    \FOR{$k = 0,1,\dots, K-1$}
        \STATE Server solves trust-region linear subproblem:
        \STATE \hskip1em $x^{k+1} \;=\; \underset{x \in \mathcal{X}}{\operatorname{argmin}}\ \langle m^{k}, x \rangle \quad\text{s.t.}\quad \|x - x^k\| \le \eta.$
        \FOR{worker $i \in [n]$ {\bfseries in parallel}}
            \STATE Good worker $i \in \mathcal{G}$: compute $g_i^{k+1} \leftarrow g_i(x^{k+1};\,\xi_i^{k+1})$, update $m_i^{k+1} \leftarrow (1-\alpha)\,m_i^{k} + \alpha\,g_i^{k+1}$, send $m_i^{k+1}$
            \STATE Byzantine worker $i \in \mathcal{B}$: send an arbitrary vector $m_i^{k+1} \in \mathcal{X}$
        \ENDFOR
        \STATE Server computes robust aggregate momentum: $m^{k+1} \leftarrow \mathrm{ARAGG}(m_1^{k+1},\dots,m_n^{k+1})$
    \ENDFOR
    \STATE {\bfseries Output:} final model $x^K$
\end{algorithmic}
\end{algorithm}

\paragraph{Weight decay.}
For the star-convex analysis we also use a decoupled weight-decay variant, shown in Algorithm~\ref{alg:byz-muon-decay}. The only change is the center of the trust region: instead of searching around $x^k$, the server searches around $(1-\beta)x^k$. This gives the same Muon-type LMO direction while adding the contraction induced by weight decay. In the operator/nuclear-norm geometry this corresponds to Muon with decoupled weight decay; with other norms, it recovers normalized or sign-type momentum methods with weight decay.

\begin{algorithm}[t]
\caption{Byzantine-Robust Muon with Weight Decay}
\label{alg:byz-muon-decay}
\begin{algorithmic}[1]
    \STATE {\bfseries Input:} $(\delta,c)$-ARAgg, trust-region radius $\eta > 0$, worker momentum weight $\alpha \in (0,1]$, weight decay $\beta \in (0,1)$, initial point $x^0 \in \mathcal{X}$
    \STATE Each good worker $i \in \mathcal{G}$ initializes $m_i^{0} \leftarrow g_i(x^0;\,\xi_i^{0})$; each Byzantine worker $i \in \mathcal{B}$ sends an arbitrary vector $m_i^0$
    \STATE Server computes $m^{0} \leftarrow \mathrm{ARAGG}(m_1^{0},\dots,m_n^{0})$
    \FOR{$k = 0,1,\dots, K-1$}
        \STATE Server solves trust-region linear subproblem:
        \STATE \hskip1em $x^{k+1} \;=\; \underset{x \in \mathcal{X}}{\operatorname{argmin}}\ \langle m^{k}, x \rangle \quad\text{s.t.}\quad \|x - (1-\beta)x^k\| \le \eta.$
        \FOR{worker $i \in [n]$ {\bfseries in parallel}}
            \STATE Good worker $i \in \mathcal{G}$: compute $g_i^{k+1} \leftarrow g_i(x^{k+1};\,\xi_i^{k+1})$, update $m_i^{k+1} \leftarrow (1-\alpha)\,m_i^{k} + \alpha\,g_i^{k+1}$, send $m_i^{k+1}$
            \STATE Byzantine worker $i \in \mathcal{B}$: send an arbitrary vector $m_i^{k+1} \in \mathcal{X}$
        \ENDFOR
        \STATE Server computes robust aggregate momentum: $m^{k+1} \leftarrow \mathrm{ARAGG}(m_1^{k+1},\dots,m_n^{k+1})$
    \ENDFOR
    \STATE {\bfseries Output:} final model $x^K$
\end{algorithmic}
\end{algorithm}

\subsection{Convergence Guarantees}
\label{sec:theory}
We now give convergence guarantees on non-IID data; all proofs, with explicit constants, are deferred to Appendices~\ref{apx:nonconvex-proof} and~\ref{apx:starconvex-proof}. Throughout, $q := 1 - \alpha$, and we use the two noise scales $V_G := \kappa\sigma\sqrt{\nicefrac{\alpha}{G(2-\alpha)}}$ and $V_P := \kappa\sigma\sqrt{\nicefrac{2\alpha}{2-\alpha}}$, which bound the averaged and the pairwise stochastic noise carried by the worker momenta; both are damped by a factor $\sqrt{\alpha}$ relative to the raw noise level $\kappa\sigma$ --- the variance-reduction benefit of aggregating momenta rather than raw gradients. First, we consider Algorithm~\ref{alg:byz-muon} in the stochastic non-convex case.
\begin{theorem}[Non-convex case]\label{thm:main-nonconvex}
Let Assumptions~\ref{asn:stoch-grad}, \ref{asn:lipschitz}, \ref{asn:norm-equivalence}, and~\ref{asn:bounded-heterogeneity} hold, let the aggregator satisfy Definition~\ref{def:aragg}, and let $4c\delta B < 1$. Then the iterates of Algorithm~\ref{alg:byz-muon} satisfy
\begin{align*}
    \frac1K\sum_{k=0}^{K-1}\mathbb E\|\nabla f(x^k)\|_*\\
    {}\leq \frac{\Delta_0/(\eta K)+C_1L\eta/2+C_2+C_0/(\alpha K)}{1-4c\delta B},
\end{align*}
where $\Delta_0 = f(x^0) - f_{\inf}$ and
\begin{gather*}
C_1 = 1 + \frac{4(1-\alpha)}{\alpha}, \qquad
C_2 = 2 V_G + 2c\delta V_P + 4 c \delta \zeta, \\
C_0 = \frac{2\kappa\sigma}{\sqrt{G}} + 2\sqrt{2}\, c\delta\kappa\sigma + 4c\delta B \|\nabla f(x^0)\|_*.
\end{gather*}
The same bound holds for $\min_{0 \le k < K}\mathbb{E}\|\nabla f(x^k)\|_*$ and for a uniformly sampled iterate.
\end{theorem}
For this non-convex guarantee, the theoretical output is $x^R$ with $R$ sampled uniformly from $\{0,\ldots,K-1\}$; the bound does not assert the same stationarity guarantee for the final iterate $x^K$.
For fixed parameter values, the bound decreases at rate $O(1/K)$ toward a stationarity neighborhood of size $(C_1L\eta/2+C_2)/(1-4c\delta B)$, whose numerator consists of an optimization term $O(L\eta/\alpha)$, a statistical term $2V_G=O(\kappa\sigma\sqrt{\alpha/G})$, and a Byzantine term $O(c\delta(\sqrt{\alpha}\kappa\sigma+\zeta))$. When $\delta = 0$, choosing $\eta = \Theta(K^{-3/4})$, $\alpha = \Theta(K^{-1/2})$ yields the $\mathcal{O}(K^{-1/4})$ tuned rate of normalized and trust-region SGD with momentum \citep{cutkosky2020momentum, kovalev2025understanding}. For fixed $\delta>0$ and $\zeta>0$, the residual term $\frac{4c\delta\zeta}{1 - 4c\delta B}$ cannot be eliminated merely by choosing $\eta$ and $\alpha$; it is zero when $\zeta=0$. Nonzero heterogeneity-dependent error is unavoidable in general Byzantine optimization \citep{Karimireddy2020ByzantineRobustLO} (see Remark~\ref{rem:heterogeneity-floor}), although the cited lower bound does not assert that this particular constant is optimal. Momentum does not attenuate persistent heterogeneity, though it does damp the Byzantine \emph{variance} term $2c\delta V_P = \mathcal{O}(c\delta\sqrt{\alpha}\,\kappa\sigma)$.

\begin{remark}[Approximate aggregation]\label{rem:approx-aggregation}
The first-moment analysis also permits an additive aggregation residual: if $\mathbb{E}\|m^k-\bar m^k\|_*\le c\delta r_k+\varepsilon_k^{\mathrm{agg}}$, with $r_k$ as in Lemma~\ref{lem:agg_error_rigorous}, then Theorem~\ref{thm:main-nonconvex} gains $2K^{-1}\sum_{k=0}^{K-1}\varepsilon_k^{\mathrm{agg}}$ in its numerator, while the condition $4c\delta B<1$ is unchanged. The star-convex bound gains $2\eta\sum_{k=0}^{K-1}(1-\beta/2)^{K-1-k}\varepsilon_k^{\mathrm{agg}}$; a uniformly bounded residual therefore enlarges its limiting neighborhood by at most $4\eta\sup_k\varepsilon_k^{\mathrm{agg}}/\beta$. The same substitution adds $2K^{-1}\sum_{k<K}\varepsilon_k^{\mathrm{agg}}$ to the numerator of the inexact non-convex bound. A concrete residual for a smoothed geometric median with certified objective accuracy is given in Lemma~\ref{lem:smoothed-median}.
\end{remark}

\paragraph{Second-order smoothness.} If $f$ has an $H$-Lipschitz Hessian, evaluating worker gradients at extrapolated points $\bar{x}^{k} = x^{k-1} + \frac{1}{\alpha}(x^{k} - x^{k-1})$ \citep{cutkosky2020momentum, kovalev2025understanding, sadiev2025improved} replaces the drift $\frac{2(1-\alpha)L\eta}{\alpha}$ above by $\frac{(1-\alpha)H\eta^2}{\alpha^2}$, plus a Byzantine cross term $\frac{4c\delta B(1-\alpha)L\eta}{\alpha}$ of the same order in $\eta/\alpha$ as the removed drift; the tuned complexity improves at $\delta = 0$ to $\mathcal{O}(\varepsilon^{-3.5})$ with the Byzantine floor unchanged (Theorem~\ref{thm:second-order}, Appendix~\ref{apx:second-order}).

Next, we consider Algorithm~\ref{alg:byz-muon-decay} in the stochastic star-convex case. Let $\Delta_k := \mathbb{E}[f(x^{k}) - f(x^*)]$.
\begin{theorem}[Star-convex case]\label{thm:main-starconvex}
    Let Assumptions~\ref{asn:stoch-grad}, \ref{asn:lipschitz}, \ref{asn:norm-equivalence}, \ref{asn:star-cvx}, and~\ref{asn:bounded-heterogeneity} hold, let the aggregator satisfy Definition~\ref{def:aragg}, and let the parameters satisfy $0 < \beta < \alpha \le 1$ and $\eta \ge \beta\max\{\|x^0\|, \|x^*\|\}.$ Then, the iterates of Algorithm~\ref{alg:byz-muon-decay} satisfy
    \begin{align*}
        \Delta_K \leq \left(1-\tfrac{\beta}{2}\right)^K \left(\Delta_0 + C_3\right) + \frac{2}{\beta}\left( C_4 + \frac{L K_B^2}{\beta}\right),
    \end{align*}
    where the constants are defined as
    \begin{gather*}
    C_3 = \frac{2\eta\kappa\sigma}{(\alpha-\beta/2)\sqrt G}\\
    {}+\frac{2\sqrt2\eta\kappa\sigma c\delta}{\alpha-\beta/2}\\
    {}+\frac{4c\delta\eta B\|\nabla f(x^0)\|_*}{\alpha-\beta/2}, \\
    C_4 = \frac{4L\eta^2}{\alpha} + 2\eta V_G\\
    {}+4c\delta\eta\zeta + 2c\delta\eta V_P,\\
    K_B = \frac{4c\delta\eta\,\alpha B(1-\beta)}{\alpha - \beta}.
    \end{gather*}
\end{theorem}
Under the stronger assumption of star-convexity, Algorithm~\ref{alg:byz-muon-decay} achieves \textit{linear convergence}, at contraction rate $(1 - \beta/2)$, to a noise neighborhood of size $\frac{2}{\beta}C_4 + \frac{2LK_B^2}{\beta^2}$ (formal version: Theorem~\ref{thm:starconvex-apx}). Larger weight decay $\beta$ accelerates the contraction but inflates the neighborhood and must satisfy $\beta < \alpha$: a trade-off, not a free acceleration. Unlike the non-convex case, \emph{no} upper bound on $\delta B$ is required, since the relative-heterogeneity history is absorbed by the weight-decay contraction and enters only through the quadratic term $LK_B^2/\beta^2$; at $\delta = 0$ the guarantee recovers the structure of the centralized trust-region result of \citet{kovalev2025understanding} up to constants. In both theorems, the neighborhood combines momentum-damped stochastic noise $\mathcal{O}(\kappa\sigma\sqrt{\alpha})$ with a Byzantine term $\mathcal{O}(a(\zeta + \sqrt{\alpha}\kappa\sigma))$, where $a:=c\delta$ is the first-moment aggregation factor and need not scale linearly with $\delta$. Under the joint scaling $(\eta,\beta) \to (t\eta, t\beta)$, the optimization term $\mathcal{O}(L\eta^2/(\alpha\beta))$ vanishes at the price of a slower contraction, while the statistical and Byzantine terms scale as $(\eta/\beta)\big(V_G + c\delta(\zeta + V_P)\big)$ (and $2LK_B^2/\beta^2$ persists, $\propto (c\delta B\,\eta/\beta)^2$) with $\eta/\beta \ge \max\{\|x^0\|,\|x^*\|\}$ under the theorem's condition; they are controlled by the momentum damping $\sqrt{\alpha}$ and the corruption level $\delta$ rather than by the radius alone.

\subsection{Inexact Muon}\label{sec:inexact-main}
In practice Muon computes the orthogonalized direction only approximately, by a few Newton--Schulz iterations \citep{jordan6muon, amsel2025polar}. Following \citet{shulgin2025beyond}, we model this by an additive-error LMO oracle: $x^{k+1} = (1-\beta)x^{k} + \eta\,\hat{d}^{\,k}$ ($\beta = 0$ for Algorithm~\ref{alg:byz-muon}), where $\|\hat{d}^{\,k} - d^{k}\| \le \bar{\delta} < 1$ for an exact LMO direction $d^{k} \in \arg\min_{\|d\| \le 1}\langle m^{k}, d\rangle$, $\hat{d}^{\,k}$ possibly infeasible (Assumption~\ref{asn:inexact}). All results of this paper extend to this oracle (Appendix~\ref{apx:inexact}); we state the non-convex case, the remaining counterparts being Theorems~\ref{thm:inexact-starconvex}--\ref{thm:inexact-second-order}.
\begin{theorem}[Inexact non-convex case]\label{thm:main-inexact}
Let the assumptions of Theorem~\ref{thm:main-nonconvex} hold with the inexact update above in place of the exact trust-region step, and let $4c\delta B < 1 - \bar{\delta}$. Then the resulting iterates satisfy
\begin{align*}
\frac1K\sum_{k=0}^{K-1}\mathbb E\|\nabla f(x^k)\|_*\\
{}\leq\frac{\Delta_0/(\eta K)+C_{1,\bar\delta}L\eta/2+C_2+C_0/(\alpha K)}{1-\bar\delta-4c\delta B},
\end{align*}
where $C_{1,\bar{\delta}} = (1+\bar{\delta})^2 + \frac{4(1+\bar{\delta})(1-\alpha)}{\alpha}$, and $C_0, C_2$ are \emph{exactly} as in Theorem~\ref{thm:main-nonconvex}.
\end{theorem}
Inexactness deflates progress by $(1-\bar\delta)$ and leaves the numerator's adversarial constants $C_2$ and $C_0$ unchanged, but it also tightens the admissible condition to $4c\delta B<1-\bar\delta$. Thus the \emph{overall} adversarial floor can increase through the denominator. The second-moment route (Theorem~\ref{thm:inexact-second-moment}) remains condition-free. The star-convex and second-order extensions appear in Appendix~\ref{apx:inexact}. These are conditional guarantees: using five Newton--Schulz steps in our experiments does not itself verify a small $\bar\delta$.

\section{Experiments}\label{sec:experiments}
We test whether the proposed combination of worker momentum, robust aggregation, and Muon geometry is effective on neural-network benchmarks, and whether the dependence on corruption and temporal averaging accords with the theory. All results below are from completed runs. Unless stated otherwise, we report the mean $\pm$ sample standard deviation over five seeds; the transfer experiments use three seeds, and the shared-checkpoint geometry control uses ten. The experimental protocol is described in Appendix~\ref{apx:experiment-details}; compact per-seed summaries and run configurations accompany the source.

\begin{figure}[t]
\centering
\includegraphics[width=0.58\textwidth]{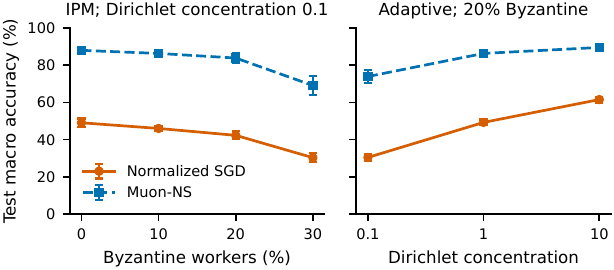}\hfill
\includegraphics[width=0.40\textwidth]{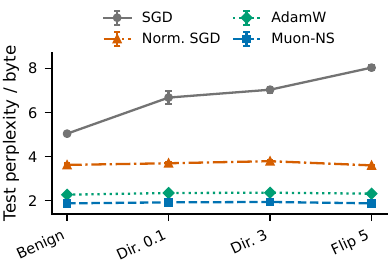}
\caption{ResNet-18 test macro-accuracy for normalized SGD and Muon-NS versus Byzantine fraction under IPM and Dirichlet concentration under the adaptive attack (left two panels); independent TinyStories byte-level test perplexity of the final round-1200 model (right). Points show five-seed means with sample standard deviations. The Byzantine fractions on the left axis are nominal (10\% and 30\% round to two and eight of 25 workers).}\label{fig:vision-trends}
\end{figure}

\paragraph{Protocol.} In the primary simulated federated vision benchmark, 25 workers train CIFAR-10 ResNet-18~\citep{he2016deep} (11.17M parameters) or ViT-Tiny~\citep{dosovitskiy2021image,touvron2021training} (5.36M), with Dirichlet-$0.1$ non-IID partitions~\citep{hsu2019measuring}. Five of the 25 workers are Byzantine in attacked runs unless the Byzantine fraction is varied (Figure~\ref{fig:vision-trends}); the training data are partitioned across the honest workers only (Appendix~\ref{apx:experiment-details}). We compare ordinary SGD, globally normalized SGD, sign updates, and Muon with five Newton--Schulz iterations (Muon-NS), keeping worker momentum ($\alpha=0.1$) and the bucketed~\citep{Karimireddy2020ByzantineRobustLO} robust federated averaging (RFA, smoothed geometric median)~\citep{Pillutla2022} aggregator (bucket size 2) fixed. The SGD row is therefore the worker-momentum method of \citet{karimireddy2021learning} (see also \citealp{farhadkhani2022byzantine}) with the bucketed robust aggregation of \citet{Karimireddy2020ByzantineRobustLO}, RFA in place of centered clipping, so the comparison isolates the update geometry on top of an existing robust baseline. Ordinary SGD applies the aggregated momentum directly; the other three geometries apply steepest-descent directions under a norm~\citep{bernstein2024old} to the aggregated momentum: Muon-NS orthogonalizes each weight matrix separately (convolution kernels are reshaped to out $\times$ (in $\cdot k\cdot k$); one-dimensional parameters are $\ell_2$-normalized per tensor), the sign geometry uses $\operatorname{sign}(m)/\sqrt{\mathrm{numel}}$ per tensor, and globally normalized SGD divides the concatenated update by its single Euclidean norm. This benchmark runs 500 communication rounds; learning rates are selected on validation data (SGD/sign/Muon-NS: $0.01$; normalized SGD: $0.03$), and for the normalized, sign, and Muon-NS geometries the learning rate is the trust-region radius $\eta$ of Algorithm~\ref{alg:byz-muon-decay} (weight decay $10^{-4}$ is applied as decoupled multiplicative decay for every geometry; Appendix~\ref{apx:experiment-details}). The metric in this benchmark is held-out test macro-accuracy at a validation-selected checkpoint. Attacks include sign-flip, ALIE~\citep{baruch2019little}, inner-product manipulation (IPM)~\citep{xie2020fall}, mimic~\citep{Karimireddy2020ByzantineRobustLO}, and an adaptive directional attack; sign-flip and IPM share the same construction, a negative multiple of the honest mean, at two different scales. These are bounded implementations of specified attacks, not all possible Byzantine strategies. Appendix~\ref{apx:experiment-details} defines each attack and describes the learning-rate and checkpoint selection protocol.

\begin{table}[t]
\centering\small
\caption{CIFAR-10 test macro-accuracy (\%, higher is better), mean $\pm$ sample standard deviation over five seeds. Same bucketed RFA aggregation and worker momentum in every row; 20\% Byzantine workers except Clean.}\label{tab:vision-main}
\setlength{\tabcolsep}{3.2pt}
\resizebox{\textwidth}{!}{\begin{tabular}{lrrrrrr}
\toprule
Geometry & Clean & Sign-flip & ALIE & IPM & Mimic & Adaptive \\
\midrule
ResNet-18: SGD & 39.8 $\pm$ 2.0 & 29.2 $\pm$ 2.3 & 27.1 $\pm$ 1.4 & 32.9 $\pm$ 4.2 & 29.7 $\pm$ 5.5 & 25.1 $\pm$ 2.3 \\
ResNet-18: Normalized SGD & 49.0 $\pm$ 2.3 & 34.8 $\pm$ 2.7 & 34.4 $\pm$ 1.8 & 42.3 $\pm$ 2.1 & 44.3 $\pm$ 1.7 & 30.4 $\pm$ 1.7 \\
ResNet-18: Sign & 54.3 $\pm$ 1.8 & 41.6 $\pm$ 3.7 & 43.3 $\pm$ 1.7 & 47.2 $\pm$ 2.0 & 49.0 $\pm$ 1.1 & 41.3 $\pm$ 2.3 \\
ResNet-18: \textbf{Muon-NS} & 88.0 $\pm$ 1.4 & 78.4 $\pm$ 4.3 & 86.4 $\pm$ 0.8 & 83.9 $\pm$ 2.5 & 84.3 $\pm$ 1.2 & 74.0 $\pm$ 3.3 \\
\midrule
ViT-Tiny: SGD & 35.5 $\pm$ 2.0 & 26.6 $\pm$ 1.6 & 37.1 $\pm$ 1.2 & 29.7 $\pm$ 1.2 & 30.8 $\pm$ 1.8 & 28.4 $\pm$ 1.5 \\
ViT-Tiny: Normalized SGD & 42.4 $\pm$ 1.4 & 34.2 $\pm$ 1.2 & 42.5 $\pm$ 1.0 & 37.3 $\pm$ 1.0 & 39.4 $\pm$ 1.1 & 32.9 $\pm$ 1.5 \\
ViT-Tiny: Sign & 46.2 $\pm$ 2.2 & 35.7 $\pm$ 1.1 & 45.8 $\pm$ 1.2 & 40.3 $\pm$ 0.4 & 42.5 $\pm$ 1.5 & 35.0 $\pm$ 2.0 \\
ViT-Tiny: \textbf{Muon-NS} & 64.1 $\pm$ 2.1 & 53.8 $\pm$ 2.2 & 62.7 $\pm$ 1.5 & 58.2 $\pm$ 1.1 & 59.3 $\pm$ 1.4 & 47.1 $\pm$ 1.1 \\
\bottomrule

\end{tabular}}
\end{table}

\paragraph{Vision.} Muon-NS is strongest in every ResNet-18 and ViT-Tiny column of Table~\ref{tab:vision-main}; for the adaptive attack its respective accuracies are $74.0\%$ and $47.1\%$, versus $41.3\%$ and $35.0\%$ for sign, the next strongest listed geometry. The advantage does not imply universal protection: the adaptive attack reduces Muon-NS ResNet-18 accuracy by about 14 points from its clean value, and worst-class accuracy is substantially lower than macro-accuracy (Table~\ref{tab:worst-class}): for Muon-NS on ResNet-18 it is $72.4\%$ against a macro-accuracy of $88.0\%$ when clean and $25.1\%$ against $74.0\%$ under the adaptive attack, with large seed variance. Figure~\ref{fig:vision-trends} shows sensitivity to Byzantine fraction and non-IID severity for normalized SGD and Muon-NS. Accuracy deteriorates at the nominal 30\% corruption level (eight of 25 workers), and the drop is largest under strong heterogeneity: at this level of IPM corruption Muon-NS reaches $69.1\%$ at Dirichlet concentration $0.1$ but $86.9\%$ and $90.1\%$ at concentrations 1 and 10 (five-seed means from the same sweep; the latter two points are not plotted). This is consistent with a nonzero Byzantine/heterogeneity floor but not a direct numerical verification of the theorem's constants.

\begin{table}[t]
\centering\small
\caption{Fixed-honest CIFAR-10 test macro-accuracy (\%, higher is better), mean $\pm$ sample standard deviation over five seeds. Both methods use bucketed RFA and 500 final-round updates with 20 honest workers and five benign or Byzantine messages. Learning rates are selected on separate validation seeds.}\label{tab:vision-adamw}
\setlength{\tabcolsep}{4pt}
\begin{tabular}{lrrr}
\toprule
Model and method & Benign & Dir. $3$ & Flip $5$ \\
\midrule
ResNet-18: AdamW & 66.7 $\pm$ 2.0 & 17.2 $\pm$ 4.7 & 21.1 $\pm$ 5.9 \\
ResNet-18: \textbf{Muon-NS} & 88.1 $\pm$ 0.8 & 30.4 $\pm$ 6.9 & 44.4 $\pm$ 5.6 \\
\midrule
ViT-Tiny: AdamW & 64.7 $\pm$ 1.1 & 18.0 $\pm$ 3.2 & 21.5 $\pm$ 1.9 \\
ViT-Tiny: \textbf{Muon-NS} & 66.5 $\pm$ 1.1 & 25.3 $\pm$ 3.4 & 31.7 $\pm$ 4.3 \\
\bottomrule

\end{tabular}
\end{table}

\paragraph{Comparison with AdamW.} Table~\ref{tab:vision-adamw} adds server-side AdamW after the same robust aggregation. On ViT-Tiny, Muon-NS reaches $66.5\%$ versus $64.7\%$ without attack, $25.3\%$ versus $18.0\%$ under directional strength $3$, and $31.7\%$ versus $21.5\%$ under sign-flip strength $5$. Its attack-induced loss is smaller by $5.6$ and $8.4$ points respectively (paired 95\% intervals: $0.8$--$10.3$ and $4.9$--$12.0$). Muon-NS also has higher attacked accuracy on ResNet-18, but its $88.1\%$ benign accuracy is far above AdamW's $66.7\%$; this comparison does not by itself isolate an attack-sensitivity advantage. Appendix~\ref{apx:experiment-details} gives the optimizer and tuning details.

\begin{table}[H]
\centering\small
\caption{Fixed-honest ResNet-18/CIFAR-10 test macro-accuracy (\%, higher is better) under an attack chosen after four queries to bucketed RFA, mean $\pm$ sample standard deviation over six paired seeds. The same candidate-selection rule is applied with both aggregators. $\Delta$ is the seedwise attack-minus-benign accuracy change.}\label{tab:oracle-control}
\setlength{\tabcolsep}{5pt}
\begin{tabular}{lrrr}
\toprule
Aggregator & Benign & RFA-query attack & $\Delta$ \\
\midrule
Mean & 88.7 $\pm$ 1.0 & 10.0 $\pm$ 0.0 & -78.7 $\pm$ 1.0 \\
\textbf{RFA} & 88.1 $\pm$ 0.7 & 36.5 $\pm$ 6.6 & -51.7 $\pm$ 6.2 \\
\bottomrule

\end{tabular}
\end{table}

\paragraph{Robust aggregation under an RFA-query attack.} In the original mild-attack cells, bucketed mean matches or exceeds RFA (Table~\ref{tab:aggregators}). In the stronger fixed-honest control of Table~\ref{tab:oracle-control}, the two methods start at $88.7\%$ and $88.1\%$ without attack. The attacker chooses among four forged-message candidates after querying the current bucketed RFA output. Under this common attack rule, mean falls to $10.0\%$ while RFA retains $36.5\%$; RFA reduces the attack-induced loss by $27.0$ points (paired 95\% interval: $20.3$--$33.8$). The attack is chosen against RFA, making this a particularly direct test of the aggregator, although the remaining $51.7$-point RFA drop rules out a general security claim. The earlier directional sweep is retained in Appendix~\ref{apx:experiment-details}.

\begin{table}[H]
\centering\small
\caption{Shared-checkpoint ResNet-18/CIFAR-10 test macro-accuracy (\%, higher is better), mean $\pm$ sample standard deviation over ten paired seeds. Both geometries start from the same 500-round benign Muon-NS checkpoint, then run 200 further rounds under bucketed RFA with matched honest-worker streams. $\Delta$ is attack-minus-benign accuracy within a geometry and seed.}\label{tab:geometry-shared}
\setlength{\tabcolsep}{3.1pt}
\resizebox{\textwidth}{!}{\begin{tabular}{lrrrrr}
\toprule
Geometry & Benign & Dir. $3$ & Flip $5$ & $\Delta$ Dir. $3$ & $\Delta$ Flip $5$ \\
\midrule
Tensorwise $\ell_2$ & 90.8 $\pm$ 0.4 & 82.3 $\pm$ 3.4 & 82.4 $\pm$ 3.9 & -8.5 $\pm$ 3.4 & -8.3 $\pm$ 3.7 \\
\textbf{Muon-NS} & 91.4 $\pm$ 0.3 & 83.7 $\pm$ 2.2 & 85.0 $\pm$ 2.8 & -7.7 $\pm$ 2.2 & -6.4 $\pm$ 2.7 \\
\bottomrule

\end{tabular}}
\end{table}

\paragraph{Matrix geometry from a common checkpoint.} The earlier tensorwise $\ell_2$ control starts far below Muon-NS without attack ($50.8\%$ versus $88.6\%$; Table~\ref{tab:robustness-controls}). The matched continuation in Table~\ref{tab:geometry-shared} closes this benign gap to $0.6$ points: Muon-NS reaches $91.4\%$ and tensorwise $\ell_2$ $90.8\%$. Muon-NS remains ahead under directional strength $3$ ($83.7\%$ versus $82.3\%$) and sign-flip strength $5$ ($85.0\%$ versus $82.4\%$). More importantly, the attack-induced loss is smaller for Muon-NS by $0.8$ points under directional and $1.9$ points under sign-flip; the paired 95\% intervals are $-0.4$--$1.9$ and $0.6$--$3.3$, respectively. The common start and matched benign performance make the directional accuracy gain more informative than in the earlier control, while the sign-flip reduction in sensitivity is the clearer geometry-specific effect.

\paragraph{Language model.} We train a 38.15M-parameter, 12-layer byte-level decoder on TinyStories~\citep{eldan2023tinystories} for 1200 rounds with seven honest workers and two additional benign or Byzantine messages, fixed unbucketed RFA aggregation, and five seeds. Learning rates are chosen on a validation-only stream; the final models are evaluated on 100 fixed windows from a stream of 4000 disjoint, previously unused stories. Table~\ref{tab:language-main} reports independent test perplexity per byte, not token-level perplexity. Muon-NS reaches $1.875$ without attack versus $2.268$ for AdamW, the strongest additional baseline, and remains lower under directional strength $3$ ($1.938$ versus $2.358$) and sign-flip strength $5$ ($1.876$ versus $2.315$). The directional attack raises Muon-NS cross-entropy by $0.033$ from its own benign result, versus $0.039$ for AdamW; this paired difference is small and uncertain, so the robust conclusion is Muon-NS's better absolute test perplexity under the tested attacks. The longer run and independent test remove the short-budget validation-only limitations of the earlier 300-round study (Table~\ref{tab:language-full}); SGD's selected rate still lies on the upper edge of its expanded grid.

\begin{table}[H]
\centering\small
\caption{TinyStories byte-level perplexity on the independent story-disjoint test split (lower is better), mean $\pm$ sample standard deviation over five seeds at round 1200. The same seven honest workers and two benign or Byzantine messages are used in every condition.}\label{tab:language-main}
\setlength{\tabcolsep}{3.4pt}
\begin{tabular}{lrrrr}
\toprule
Method & Benign & Dir. $0.1$ & Dir. $3$ & Flip $5$ \\
\midrule
SGD & 5.038 $\pm$ 0.080 & 6.676 $\pm$ 0.304 & 7.034 $\pm$ 0.128 & 8.035 $\pm$ 0.121 \\
Normalized SGD & 3.620 $\pm$ 0.064 & 3.695 $\pm$ 0.042 & 3.789 $\pm$ 0.043 & 3.599 $\pm$ 0.009 \\
AdamW & 2.268 $\pm$ 0.053 & 2.345 $\pm$ 0.075 & 2.358 $\pm$ 0.061 & 2.315 $\pm$ 0.047 \\
\textbf{Muon-NS} & 1.875 $\pm$ 0.003 & 1.920 $\pm$ 0.006 & 1.938 $\pm$ 0.004 & 1.876 $\pm$ 0.004 \\
\bottomrule

\end{tabular}
\end{table}

\paragraph{Temporal mechanism.} Varying the worker momentum weight $\alpha$ on ResNet-18 reveals a non-monotone trade-off: under the adaptive attack, Muon-NS reaches $51.0\%$ at $\alpha=0.03$, $74.0\%$ at $0.1$, $82.7\%$ at $0.3$, and $73.4\%$ without temporal averaging ($\alpha=1$). This supports the practical importance of temporal state, but also shows that more smoothing is not uniformly better. The full sweep for SGD, normalized SGD, and Muon-NS is in Table~\ref{tab:temporal} and Figure~\ref{fig:momentum}; the controlled theorem-facing tests are in Appendix~\ref{apx:experiment-details}.

\paragraph{Certified aggregation.} A separate matrix-quadratic control certifies the additive error of unbucketed smoothed RFA in Lemma~\ref{lem:smoothed-median}; the original eight Weiszfeld iterations satisfy the objective certificate on all tested rounds (Table~\ref{tab:certified-rfa}, Appendix~\ref{apx:experiment-details}).

\paragraph{Scope and limitations.} The theory isolates stochastic noise, non-Euclidean heterogeneity, robust aggregation, and inexact directions under explicit assumptions; the neural-network results neither measure the theorem constants nor verify those assumptions. Five Newton--Schulz iterations are not certified to have small oracle error, and better optimization at a fixed budget does not establish a faster asymptotic rate. The attacks are specified reproducible strategies. The new RFA-query adversary uses four candidates and the current bucketed RFA output, but is not an exhaustive adaptive search or a worst-case adversary. In all twelve shared cells under the original mild attacks, mean matched or exceeded RFA (Table~\ref{tab:aggregators}); the stronger fixed-honest and RFA-query controls reverse that ordering for their specified attacks, yet RFA still loses substantial accuracy. The Byzantine-fraction sweep changes honest data volume and partitions, whereas the fixed-honest controls do not. From a shared checkpoint, Muon-NS has higher directional attacked accuracy and a smaller mean attack-induced drop than tensorwise $\ell_2$, but the paired interval for the latter difference includes zero; the sign-flip drop difference is clearer. The AdamW comparison also changes temporal state ($\alpha=1$ for AdamW versus $0.1$ for Muon-NS), and the ResNet-18 benign gap precludes an intrinsic robustness interpretation there. The independent TinyStories test removes validation reuse from the new language comparison, but SGD's rate remains at the tuning-grid boundary, the model is 38.15M parameters, and only five training seeds were tested. Macro-accuracy can mask low worst-class accuracy, and the implemented smoothed Euclidean RFA is not the exact dual-norm geometric median described in Remark~\ref{rem:aggregators}. These are mechanism checks on simulated workers, not deployment-level Byzantine-security guarantees.

\section{Conclusion}
We analyzed Byzantine-robust distributed optimization with non-Euclidean trust-region updates and local momentum. The guarantees cover non-convex and star-convex objectives, second-moment aggregation under its stated assumptions, extrapolation, and inexact Muon directions. Across the tested image and language settings, Muon-NS achieves higher attacked accuracy or lower attacked perplexity than the tested SGD and globally normalized SGD configurations, and outperforms the additional AdamW baseline on an independent TinyStories test. Fixed-honest ResNet-18 controls show a substantial RFA advantage over mean under a bounded attack selected after querying RFA. From common benign checkpoints, Muon-NS retains higher directional and sign-flip accuracy than tensorwise Euclidean normalization with nearly matched benign performance; its reduction in attack-induced loss is clearest for sign-flip. The ViT-Tiny comparison with AdamW likewise shows smaller losses under both strong attacks. These experiments probe the proposed mechanism in simulated federated training but do not establish deployment-level worst-case security or the theorem constants.

\section*{Acknowledgements}
The research reported in this publication was supported by funding from King Abdullah University of Science and Technology (KAUST): i) KAUST Baseline Research Scheme, ii) CRG Grant ORFSCRG12-2024-6460, and iii) Center of Excellence for Generative AI, under award number 5940.

\bibliography{references}
\bibliographystyle{plainnat}

\newpage
\appendix
\counterwithin{figure}{section}
\counterwithin{table}{section}

\vbox{
  {\hrule height 2pt \vskip 0.15in \vskip -\parskip}
  \centering
  {\LARGE\bf Appendix\par}
  {\vskip 0.2in \vskip -\parskip \hrule height 0.5pt \vskip 0.09in}
}

\newcommand\invisiblepart[1]{%
  \refstepcounter{part}%
  \addcontentsline{toc}{part}{\protect\numberline{\thepart}#1}%
}

\invisiblepart{Appendix}
\setcounter{tocdepth}{2}
\localtableofcontents

\appendixtrue

\section{Experimental Protocol and Additional Results}\label{apx:experiment-details}

\paragraph{Data and models.} CIFAR-10 and CIFAR-100~\citep{krizhevsky2009learning} use their standard test splits; a fixed stratified 10\% of each vision training set is reserved for validation and never used for fitting. The transfer experiments also use MNIST~\citep{lecun1998gradient} and Fashion-MNIST~\citep{xiao2017fashion}. Remaining examples are split by a Dirichlet label distribution~\citep{hsu2019measuring} across the honest workers only (all 25 workers in the original Clean runs; 20 of 25 at the default Byzantine fraction, and 23 or 17 in the Byzantine-fraction sweep), redrawn until every honest worker holds at least 32 examples; the primary concentration is $0.1$. Byzantine workers hold no data of their own and forge their messages from the honest momenta, so the data volume per honest worker differs between clean and attacked runs. CIFAR-10 and CIFAR-100 training batches use random crops (4-pixel reflect padding) and horizontal flips with probability $1/2$; MNIST and Fashion-MNIST use no augmentation, and evaluation batches are never augmented. The ResNet-18~\citep{he2016deep} has 11,173,962 parameters and ViT-Tiny~\citep{dosovitskiy2021image,touvron2021training} 5,362,762. In the original CIFAR-10 vision benchmark, each worker uses batch size 128, weight decay $10^{-4}$ (applied as decoupled multiplicative decay $x\leftarrow(1-\beta)x$ before each step with $\beta=10^{-4}$, for every geometry including SGD), 500 rounds, evaluation every 25 rounds, and a validation-selected checkpoint (the evaluated checkpoint with highest validation macro-accuracy). Muon-NS uses five Newton--Schulz iterations. The transfer runs keep the same batch size, weight decay, partitioning, validation hold-out, checkpoint selection, worker momentum, aggregator, and attack settings but use 300 (MNIST), 400 (Fashion-MNIST), and 700 (CIFAR-100) rounds with evaluation every 50 rounds and three seeds. MNIST and Fashion-MNIST use a small CNN (two $3\times3$ convolution layers with 32 and 64 channels, each followed by ReLU and $2\times2$ max-pooling, then a 128-unit fully connected layer and classifier; 421,642 parameters) and CIFAR-100 a ResNet-18 (11,220,132 parameters). All vision and language runs compute the forward pass and loss under bfloat16 autocast. The vision experiments use an NVIDIA H200 GPU on an internal research cluster with independent single-GPU trials. GPU time and memory per run are recorded in the per-seed summaries.

\paragraph{Learning-rate selection.} For the original vision benchmark, learning rates were tuned per geometry on ResNet-18/CIFAR-10 with seed 0 for 300 rounds under clean training and IPM, over the grids $\{0.003,0.01,0.03,0.1,0.3\}$ for SGD and globally normalized SGD and $\{0.001,0.003,0.01,0.03,0.1\}$ for sign and Muon-NS, selecting the rate that maximizes the minimum of the clean and IPM final validation macro-accuracy (ties broken by the clean value). The selected values are $0.01$ for SGD, sign, and Muon-NS and $0.03$ for globally normalized SGD; each lies strictly inside its grid. These rates were reused without retuning for ViT-Tiny, the momentum sweep, and the transfer runs. For normalized SGD, sign, and Muon-NS the learning rate is the trust-region radius $\eta$. For the earlier 300-round language comparison, the grids were $\{0.0003,0.001,0.003,0.01,0.03\}$ for SGD and globally normalized SGD and $\{0.0001,0.0003,0.001,0.003,0.01\}$ for Muon-NS, tuned with seed 0 for 100 rounds under clean training and IPM, minimizing the maximum of clean and IPM validation loss (ties broken by the clean value). The selected rates, $0.03$ for SGD and globally normalized SGD and $0.01$ for Muon-NS, are the largest value of each grid, so the language comparison may be affected by the tuning range. The tuning runs are not part of the per-seed summaries.

The earlier language study uses TinyStories~\citep{eldan2023tinystories} and a 12-layer, width-512, eight-head byte-level decoder (38,154,752 parameters), context length 512, nine workers (the training stream is sharded across nine honest workers in clean runs and seven under attack), batch size eight, gradient clipping at 1, weight decay $10^{-4}$ (decoupled), 300 rounds with one minibatch gradient per worker per round, and unbucketed RFA (bucket size 1). Every 20 rounds it evaluates 20 batches on a fixed validation stream; the reported perplexity is that of the final (round-300) model on this stream, and no checkpoint selection is performed for language. The same stream was used for learning-rate selection, so the comparison is validation-only and uses byte-level perplexity; the expanded independent-test protocol is described below. The four attacks are sign-flip, ALIE, IPM, and adaptive directional; Table~\ref{tab:language-full} gives the per-attack perplexities.

\paragraph{Adversary and statistics.} In attacked CIFAR-10 and transfer settings the default Byzantine fraction is $0.2$ (five of 25 workers); Figure~\ref{fig:vision-trends} additionally sweeps nominal fractions $0.1$ and $0.3$, which round to two and eight of 25 workers (8\% and 32\%). In language it is two of nine workers. Honest worker momentum buffers are aggregated by RFA/geometric median, bucketed with bucket size 2 in vision and unbucketed (bucket size 1) in language and controlled-quadratic experiments; attacks replace Byzantine messages before bucketing and aggregation. Each vision round freshly permutes the 25 messages, averages consecutive groups, and gives each resulting bucket one vote in the aggregator; bucket size 2 therefore produces 12 pairs and one singleton, and size 4 produces six groups of four and one singleton. The default forged messages are chosen before this permutation, whereas the RFA-query attack observes the next permutation through its four probes. At 20\% corruption, five messages can contaminate five buckets; at the eight-worker setting of the fraction sweep, eight of the 13 size-two buckets can be contaminated, so a clean-bucket majority is not guaranteed there. We use seeds 0--4 in the main vision and language comparisons, six seeds for the RFA-query control, ten for the shared-checkpoint geometry control, and three for transfer. Every $\pm$ in tables and error bars denotes the sample standard deviation across seeds, not a confidence interval. We distinguish held-out test accuracy for vision from validation perplexity in the earlier language study and independent test perplexity in the extended language study. Compact run configurations and per-seed metrics accompany the editable draft.

\paragraph{Attacks.} Let $\bar g$ be the mean and $\sigma$ the biased coordinate-wise standard deviation of the honest workers' momentum messages in the current round, and let $G$ be the number of honest workers. Sign-flip sends $-\bar g$ (the IPM construction of \citealp{xie2020fall} with $\epsilon=1$); IPM sends $-0.1\bar g$; ALIE~\citep{baruch2019little} sends $\bar g-0.1\sigma$, a mild setting; mimic~\citep{Karimireddy2020ByzantineRobustLO} copies the message of a fixed honest worker; and the adaptive directional attack sends $\bar g-0.1\sqrt G\operatorname{sign}(\bar g)\odot\sigma$ elementwise. All Byzantine workers send identical copies of the forged message. In the original benchmark the adversary sees current honest messages but does not query the aggregator; the new RFA-query control below additionally uses aggregator feedback. Controlled-quadratic suites use strength $0.1$ for all attacks.

\paragraph{Fixed-honest aggregation and geometry controls.} The additional ResNet-18/CIFAR-10 study in Table~\ref{tab:robustness-controls} uses the same 20 honest worker partitions, batch workload, and seeds $0$--$4$ in every condition, with five further messages supplied to the aggregator. In the benign control each of these messages equals the current honest mean $\bar g$; this preserves the honest computation and message count without an adversarial shift and is distinct from the 25-honest-worker Clean column of Table~\ref{tab:vision-main}. We cross bucketed mean and eight-iteration RFA (bucket size 2, no nearest-neighbor mixing) with Muon-NS and tensorwise Euclidean normalization. The latter divides each tensor by its Frobenius norm and multiplies a reshaped matrix by $\sqrt{\min\{m,n\}}$, matching the Frobenius norm of an ideal full-rank $m\times n$ polar factor; vector parameters retain unit $\ell_2$ norm. It does not match the singular-vector allocation or exact finite-iteration magnitude of Muon-NS. Both aggregators use the same learning rate within a geometry. Independent seed-$99$ validation runs of 300 rounds select the learning rate by average benign final macro-accuracy across the two aggregators, using strength-$1$ directional accuracy only to break ties. The grids include $\{0.003,0.01,0.03\}$ for both geometries and, after the Euclidean baseline favored the low boundary, $\{0.001,0.0003,0.0001,0.00003\}$ for that baseline; selection gives $0.01$ for Muon-NS and $0.0003$ for tensorwise $\ell_2$. The full evaluation uses 500 rounds, validation-selected checkpoints, and held-out test accuracy. Directional strengths $s\in\{0.1,1,3\}$ send $\bar g-s\sqrt G\operatorname{sign}(\bar g)\odot\sigma$; the diverse variant at $s=3$ adds $0.5s(g_i-\bar g)$ to the $i$th Byzantine copy, cycling through honest messages $g_i$; strong sign-flip sends $-5\bar g$. The adversary sees current honest momenta but does not optimize against a queried aggregator. Table~\ref{tab:robustness-controls} reports five-seed means and sample standard deviations; paired 95\% intervals in the text use a $t$ interval over seedwise differences and are exploratory, without correction for multiple comparisons. Compact configurations and per-seed test summaries accompany the editable draft.

\begin{table}[H]
\centering\small
\caption{Earlier fixed-honest ResNet-18/CIFAR-10 control, test macro-accuracy (\%, higher is better), mean $\pm$ sample standard deviation over five paired seeds. Every run uses 20 honest workers and five additional messages; the benign messages copy the honest mean. This 500-round protocol selects the best validation checkpoint.}\label{tab:robustness-controls}
\setlength{\tabcolsep}{3.2pt}
\resizebox{\textwidth}{!}{\begin{tabular}{lrrrrrr}
\toprule
Geometry and aggregator & Benign & Dir. $0.1$ & Dir. $1$ & Dir. $3$ & Diverse $3$ & Flip $5$ \\
\midrule
Muon-NS: Mean & 89.1 $\pm$ 0.5 & 79.4 $\pm$ 2.3 & 19.4 $\pm$ 1.4 & 10.0 $\pm$ 0.0 & 10.0 $\pm$ 0.0 & 10.0 $\pm$ 0.0 \\
Muon-NS: RFA & 88.6 $\pm$ 1.1 & 74.0 $\pm$ 3.3 & 34.8 $\pm$ 5.9 & 32.9 $\pm$ 4.7 & 35.4 $\pm$ 6.6 & 45.1 $\pm$ 5.2 \\
\midrule
Tensorwise $\ell_2$: Mean & 49.1 $\pm$ 0.7 & 37.7 $\pm$ 2.1 & 11.5 $\pm$ 1.6 & 10.0 $\pm$ 0.0 & 10.0 $\pm$ 0.0 & 10.0 $\pm$ 0.0 \\
Tensorwise $\ell_2$: RFA & 50.8 $\pm$ 1.0 & 32.7 $\pm$ 1.9 & 16.7 $\pm$ 2.7 & 16.3 $\pm$ 2.8 & 14.5 $\pm$ 2.5 & 20.5 $\pm$ 3.4 \\
\bottomrule

\end{tabular}}
\end{table}

\begin{figure}[H]
\centering
\includegraphics[width=\textwidth]{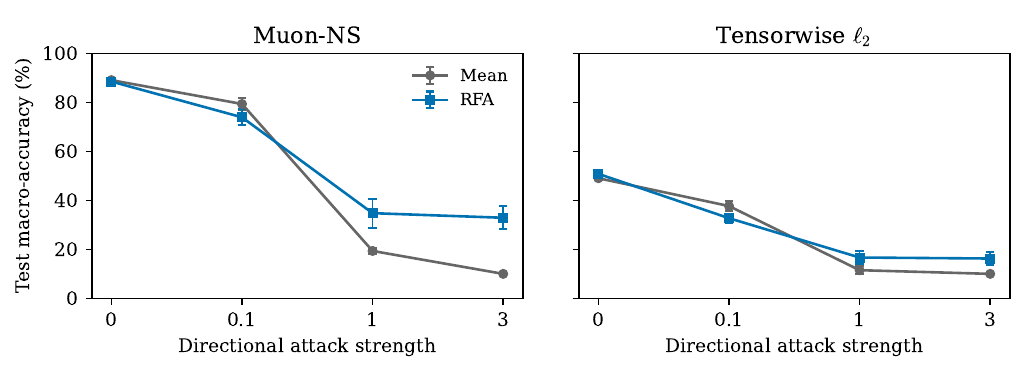}
\caption{Earlier fixed-honest ResNet-18 test macro-accuracy under the directional attack. Strength zero denotes five benign copies of the honest mean, with the same 20 honest workers and data partitions as in attacked runs. Curves show five-seed means with sample standard deviations.}\label{fig:robustness-controls}
\end{figure}

\paragraph{RFA-query aggregation control.} Table~\ref{tab:oracle-control} fixes 20 honest ResNet-18 workers, their partitions and batch streams, five additional messages, learning rate $0.01$, and 500 final-round updates. Six seeds cross bucketed mean and RFA with benign copies of the honest mean and a bounded attack. At each round the attacker observes the current honest momenta and queries eight-iteration bucketed RFA, using the same next random bucket permutation, on four candidate forged messages: directional strengths 1 and 3, diverse directional strength 3, and sign-flip strength 5. It selects the candidate whose RFA output has the smallest inner product with the honest mean, divided by the mean's squared norm. The same candidate-selection rule is used in the mean and RFA runs, with current messages recomputed along each training trajectory. Thus the attack has current-message and RFA-output feedback, but searches only these four candidates rather than the entire Byzantine message space. We report held-out test macro-accuracy of the final model, without checkpoint selection. The paired difference of attack-minus-benign drops uses a Student $t$ interval over six seeds.

\paragraph{Shared-checkpoint matrix geometry control.} The ten evaluation seeds of Table~\ref{tab:geometry-shared} start from corresponding 500-round benign Muon-NS/RFA ResNet-18 checkpoints. From each identical model and honest-worker momentum state, Muon-NS and tensorwise $\ell_2$ each continue for 200 rounds under benign messages, directional strength 3, or sign-flip strength 5. The continuation uses the same worker partitions, matched honest-worker batch streams and bucket seeds across branches, and final-round held-out test evaluation. Two independent tuning checkpoints (seeds 99 and 100) select rates by mean benign final validation macro-accuracy over 200-round continuations. The common two-seed grids are $\{0.0005,0.001,0.002,0.003,0.005,0.01\}$ for Muon-NS and $\{0.0001,0.0003,0.0005,0.001\}$ for tensorwise $\ell_2$; the selected rates are $0.001$ and $0.0005$. These are separate from the earlier 500-round rates in Table~\ref{tab:robustness-controls}. The paired attack-minus-benign differences and 95\% Student $t$ intervals use ten seeds and are exploratory, without multiplicity correction.

\paragraph{AdamW and extended language protocols.} Table~\ref{tab:vision-adamw} fixes 20 honest CIFAR-10 workers and five additional benign or Byzantine messages, bucketed RFA, and 500 final-round updates. Server-side AdamW uses $\beta_1=0.9$, $\beta_2=0.999$, $\epsilon=10^{-8}$ and no internal weight decay; the same multiplicative $10^{-4}$ decay is applied to both methods before each update. AdamW uses current honest gradients ($\alpha=1$), whereas Muon-NS uses worker momentum ($\alpha=0.1$); this is an optimizer comparison, not a geometry-only ablation. AdamW rates $0.0001$ (ResNet-18) and $0.001$ (ViT-Tiny), and the ViT-Tiny Muon-NS rate $0.01$, maximize clean final validation macro-accuracy after 300 rounds over seeds 99 and 100; AdamW grids are $\{0.00003,0.0001,0.0003,0.001,0.003\}$ and the ViT-Tiny Muon-NS grid is $\{0.001,0.003,0.01,0.03,0.06\}$. The ResNet-18 Muon-NS rate $0.01$ is inherited from the original benchmark.

For the extended TinyStories study, all conditions use seven honest training workers and two additional benign or Byzantine messages, unbucketed RFA, batch size eight, context length 512, gradient clipping at 1, and 1200 final-round updates. The official validation stories are split into a tuning stream and 4000 held-out stories, excluding the 252 stories touched by the earlier fixed evaluation windows from the held-out set. The test stream is not read during tuning; final test perplexity uses 100 fixed windows and no checkpoint selection. Clean 600-round tuning on seeds 99 and 100 selects rates by final validation loss from grids $\{0.03,0.06,0.12,0.24,0.48\}$ for SGD and globally normalized SGD, $\{0.005,0.01,0.02,0.04,0.08\}$ for Muon-NS, and $\{0.00003,0.0001,0.0003,0.001,0.003\}$ for AdamW. The selected rates are $0.48$, $0.06$, $0.02$, and $0.0003$, respectively. AdamW uses the same server moments as above, and its worker momenta are disabled; the other methods use worker momentum weight $\alpha=0.1$. SGD is the only selected rate on a grid boundary. The byte-level decoder has 38,154,752 parameters. The split manifest, compact per-seed summaries, tuning grids, and GPU telemetry accompany the editable source.

\paragraph{Transfer datasets.} Table~\ref{tab:transfer} reports the three transfer datasets (three seeds; sign was not run). The ordering of the three geometries is the same as on CIFAR-10 in every column; on CIFAR-100 under the adaptive attack Muon-NS, normalized SGD, and SGD reach $61.0\%$, $10.9\%$, and $4.4\%$, while on MNIST all three remain above $65\%$. The learning rates were transferred from the CIFAR-10 ResNet-18 tuning, so baselines were not retuned.

\begin{table}[H]
\centering\small
\caption{Earlier 300-round TinyStories validation perplexity per byte; five seeds. This fixed validation stream was also used for learning-rate selection (final-round model, no checkpoint selection). The independent-test extension is in Table~\ref{tab:language-main}.}\label{tab:language-full}
\resizebox{\textwidth}{!}{\begin{tabular}{lrrrrr}
\toprule
Geometry & Clean & Sign-flip & ALIE & IPM & Adaptive \\
\midrule
SGD & 11.62 $\pm$ 0.04 & 11.96 $\pm$ 0.05 & 11.63 $\pm$ 0.04 & 11.96 $\pm$ 0.05 & 11.69 $\pm$ 0.06 \\
Normalized SGD & 9.56 $\pm$ 0.02 & 9.60 $\pm$ 0.04 & 9.61 $\pm$ 0.05 & 9.59 $\pm$ 0.04 & 9.63 $\pm$ 0.02 \\
Muon-NS & 2.55 $\pm$ 0.01 & 2.71 $\pm$ 0.01 & 2.71 $\pm$ 0.01 & 2.70 $\pm$ 0.01 & 2.99 $\pm$ 0.02 \\
\bottomrule

\end{tabular}}
\end{table}

\begin{table}[H]
\centering\small
\caption{CIFAR-10 test worst-class accuracy (\%), mean $\pm$ sample standard deviation over five seeds, for the same runs as Table~\ref{tab:vision-main}. Worst-class accuracy is far below macro-accuracy for every geometry and has large seed variance; the baselines stay below 7\% under every attack, so the gap is not specific to Muon-NS.}\label{tab:worst-class}
\resizebox{\textwidth}{!}{\begin{tabular}{lrrrrrr}
\toprule
Geometry & Clean & Sign-flip & ALIE & IPM & Mimic & Adaptive \\
\midrule
ResNet-18: SGD & 3.9 $\pm$ 5.3 & 0.0 $\pm$ 0.0 & 0.0 $\pm$ 0.0 & 0.0 $\pm$ 0.0 & 0.0 $\pm$ 0.0 & 0.0 $\pm$ 0.0 \\
ResNet-18: Normalized SGD & 14.6 $\pm$ 10.3 & 0.0 $\pm$ 0.1 & 0.0 $\pm$ 0.0 & 1.5 $\pm$ 2.9 & 1.9 $\pm$ 2.2 & 0.0 $\pm$ 0.0 \\
ResNet-18: Sign & 19.9 $\pm$ 9.6 & 2.7 $\pm$ 5.6 & 0.5 $\pm$ 0.9 & 1.4 $\pm$ 2.3 & 6.5 $\pm$ 10.3 & 0.1 $\pm$ 0.1 \\
ResNet-18: Muon-NS & 72.4 $\pm$ 4.8 & 26.1 $\pm$ 23.7 & 69.5 $\pm$ 7.4 & 57.3 $\pm$ 14.2 & 59.2 $\pm$ 10.8 & 25.1 $\pm$ 17.0 \\
\midrule
ViT-Tiny: SGD & 4.8 $\pm$ 4.5 & 0.1 $\pm$ 0.3 & 2.4 $\pm$ 3.9 & 0.8 $\pm$ 1.8 & 0.2 $\pm$ 0.4 & 0.1 $\pm$ 0.2 \\
ViT-Tiny: Normalized SGD & 13.1 $\pm$ 8.3 & 0.4 $\pm$ 0.7 & 5.4 $\pm$ 9.7 & 3.0 $\pm$ 5.6 & 2.3 $\pm$ 4.7 & 0.2 $\pm$ 0.4 \\
ViT-Tiny: Sign & 13.9 $\pm$ 9.9 & 0.1 $\pm$ 0.3 & 6.7 $\pm$ 8.1 & 2.0 $\pm$ 2.3 & 2.4 $\pm$ 2.6 & 1.4 $\pm$ 2.8 \\
ViT-Tiny: Muon-NS & 36.5 $\pm$ 5.7 & 15.0 $\pm$ 11.5 & 31.1 $\pm$ 14.0 & 30.0 $\pm$ 6.2 & 26.1 $\pm$ 6.3 & 7.4 $\pm$ 11.1 \\
\bottomrule

\end{tabular}}
\end{table}

\begin{table}[H]
\centering\small
\caption{Transfer test macro-accuracy (\%), mean $\pm$ sample standard deviation over three seeds. Attack settings match the main vision study; learning rates were transferred from CIFAR-10 ResNet-18 tuning rather than retuned.}\label{tab:transfer}
\resizebox{\textwidth}{!}{\begin{tabular}{lrrrr}
\toprule
Geometry & Clean & Sign-flip & IPM & Adaptive \\
\midrule
MNIST/CNN: SGD & 86.4 $\pm$ 5.3 & 65.6 $\pm$ 10.0 & 80.5 $\pm$ 3.9 & 72.0 $\pm$ 5.5 \\
MNIST/CNN: Normalized SGD & 95.3 $\pm$ 1.0 & 87.5 $\pm$ 2.3 & 92.2 $\pm$ 1.4 & 87.5 $\pm$ 0.7 \\
MNIST/CNN: Muon-NS & 99.0 $\pm$ 0.0 & 97.4 $\pm$ 0.4 & 98.9 $\pm$ 0.1 & 98.5 $\pm$ 0.6 \\
\midrule
Fashion-MNIST/CNN: SGD & 73.9 $\pm$ 1.3 & 61.1 $\pm$ 4.3 & 65.9 $\pm$ 3.8 & 62.5 $\pm$ 2.4 \\
Fashion-MNIST/CNN: Normalized SGD & 79.4 $\pm$ 1.2 & 72.5 $\pm$ 2.1 & 77.1 $\pm$ 0.9 & 71.1 $\pm$ 1.6 \\
Fashion-MNIST/CNN: Muon-NS & 90.9 $\pm$ 0.3 & 88.7 $\pm$ 0.9 & 89.9 $\pm$ 0.1 & 89.8 $\pm$ 0.2 \\
\midrule
CIFAR-100/ResNet-18: SGD & 19.1 $\pm$ 0.0 & 6.2 $\pm$ 0.5 & 9.6 $\pm$ 0.2 & 4.4 $\pm$ 1.1 \\
CIFAR-100/ResNet-18: Normalized SGD & 29.8 $\pm$ 0.6 & 15.9 $\pm$ 0.4 & 21.2 $\pm$ 0.4 & 10.9 $\pm$ 1.0 \\
CIFAR-100/ResNet-18: Muon-NS & 65.4 $\pm$ 0.5 & 61.2 $\pm$ 0.5 & 64.4 $\pm$ 0.1 & 61.0 $\pm$ 0.5 \\
\bottomrule

\end{tabular}}
\end{table}
\paragraph{Aggregators, bucketing, and nearest-neighbor mixing.} Table~\ref{tab:aggregators} varies the aggregator on ResNet-18 at the primary setting (Dirichlet $0.1$, 20\% Byzantine workers, five seeds per cell). The upper block compares plain averaging, RFA, Krum~\citep{blanchard2017machine}, coordinate-wise median (CM)~\citep{yin2018byzantine}, and centered clipping (CClip)~\citep{karimireddy2021learning}, each with bucket size 2 and no nearest-neighbor mixing; this block was run for clean training, sign-flip, IPM, and mimic only, and not for sign geometry, ALIE, or the adaptive attack. Bucketed averaging matched or exceeded bucketed RFA in all twelve cells where both were run, and Muon-NS is the strongest geometry under every aggregator. Sign-flip and IPM send negative scalar multiples of the honest mean, so with 20\% identical Byzantine copies the plain average is still a positive multiple of the honest mean ($0.6\bar g$ and $0.78\bar g$, up to bucketing), which scale-invariant updates absorb; averaging nevertheless has no worst-case guarantee, and the gaps in Table~\ref{tab:vision-main} under these two attacks are largely attributable to the shared robust aggregator. The lower block crosses RFA and CM with bucket size $b\in\{1,2,4\}$~\citep{Karimireddy2020ByzantineRobustLO} and nearest-neighbor mixing (NNM)~\citep{allouah2023fixing} for Muon-NS. Without NNM, larger buckets improve both aggregators under all three attacks, most strongly for CM, which is weak when unbucketed; with NNM the two aggregators become nearly indistinguishable at every bucket size. NNM helps unbucketed RFA under IPM and mimic but lowers RFA accuracy under sign-flip at every bucket size, and for bucketed RFA under IPM and mimic it is roughly neutral at $b=2$ (0.5--1.2 points) and costs 2--3 points at $b=4$. No setting of either robust aggregator reaches plain averaging under these three attacks.

\begin{table}[H]
\centering\footnotesize
\caption{ResNet-18 test macro-accuracy (\%), mean $\pm$ sample standard deviation over five seeds, versus aggregator (Dirichlet $0.1$; 20\% Byzantine workers except Clean). Upper block: bucket size 2, no NNM. Lower block: Muon-NS with RFA or CM, bucket size $b$, with and without nearest-neighbor mixing; -- marks configurations not run.}\label{tab:aggregators}
\resizebox{\textwidth}{!}{\begin{tabular}{lrrrr}
\toprule
Geometry and aggregator & Clean & Sign-flip & IPM & Mimic \\
\midrule
SGD: Mean & 40.6 $\pm$ 2.5 & 39.3 $\pm$ 2.8 & 39.7 $\pm$ 3.2 & 33.1 $\pm$ 2.4 \\
SGD: RFA & 39.8 $\pm$ 2.0 & 29.2 $\pm$ 2.3 & 32.9 $\pm$ 4.2 & 29.7 $\pm$ 5.5 \\
SGD: Krum & 16.8 $\pm$ 1.1 & 14.1 $\pm$ 1.8 & 15.3 $\pm$ 1.3 & 15.9 $\pm$ 2.4 \\
SGD: CM & 19.9 $\pm$ 1.3 & 11.3 $\pm$ 1.1 & 16.9 $\pm$ 2.7 & 17.0 $\pm$ 3.4 \\
SGD: CClip & 18.2 $\pm$ 1.8 & 16.8 $\pm$ 2.4 & 18.4 $\pm$ 2.6 & 14.6 $\pm$ 1.8 \\
\midrule
Normalized SGD: Mean & 49.4 $\pm$ 1.9 & 46.3 $\pm$ 1.6 & 48.7 $\pm$ 2.1 & 46.7 $\pm$ 3.0 \\
Normalized SGD: RFA & 49.0 $\pm$ 2.3 & 34.8 $\pm$ 2.7 & 42.3 $\pm$ 2.1 & 44.3 $\pm$ 1.7 \\
Normalized SGD: Krum & 21.8 $\pm$ 3.0 & 14.8 $\pm$ 2.1 & 13.6 $\pm$ 2.2 & 21.2 $\pm$ 3.9 \\
Normalized SGD: CM & 22.7 $\pm$ 2.7 & 13.2 $\pm$ 2.1 & 19.2 $\pm$ 1.7 & 18.7 $\pm$ 2.6 \\
Normalized SGD: CClip & 25.5 $\pm$ 2.3 & 25.2 $\pm$ 2.1 & 31.8 $\pm$ 2.0 & 22.4 $\pm$ 2.3 \\
\midrule
Muon-NS: Mean & 89.2 $\pm$ 0.6 & 88.3 $\pm$ 0.9 & 88.8 $\pm$ 0.7 & 88.2 $\pm$ 0.7 \\
Muon-NS: RFA & 88.0 $\pm$ 1.4 & 78.4 $\pm$ 4.3 & 83.9 $\pm$ 2.5 & 84.3 $\pm$ 1.2 \\
Muon-NS: Krum & 54.9 $\pm$ 2.7 & 25.8 $\pm$ 4.0 & 22.7 $\pm$ 3.8 & 46.7 $\pm$ 7.2 \\
Muon-NS: CM & 70.3 $\pm$ 2.6 & 28.0 $\pm$ 1.3 & 60.3 $\pm$ 3.9 & 64.9 $\pm$ 4.1 \\
Muon-NS: CClip & 76.3 $\pm$ 2.7 & 69.6 $\pm$ 2.0 & 81.8 $\pm$ 1.8 & 73.2 $\pm$ 2.6 \\
\midrule
Muon-NS: RFA, $b=1$ & -- & 76.2 $\pm$ 5.1 & 77.4 $\pm$ 4.0 & 78.1 $\pm$ 6.0 \\
Muon-NS: RFA+NNM, $b=1$ & -- & 67.1 $\pm$ 4.9 & 83.4 $\pm$ 2.9 & 84.7 $\pm$ 1.8 \\
Muon-NS: RFA, $b=2$ & 88.0 $\pm$ 1.4 & 78.4 $\pm$ 4.3 & 83.9 $\pm$ 2.5 & 84.3 $\pm$ 1.2 \\
Muon-NS: RFA+NNM, $b=2$ & -- & 73.6 $\pm$ 4.3 & 82.7 $\pm$ 2.5 & 83.8 $\pm$ 2.1 \\
Muon-NS: RFA, $b=4$ & -- & 79.5 $\pm$ 2.6 & 85.5 $\pm$ 1.9 & 85.3 $\pm$ 0.6 \\
Muon-NS: RFA+NNM, $b=4$ & -- & 77.3 $\pm$ 2.1 & 83.0 $\pm$ 1.8 & 82.2 $\pm$ 1.7 \\
Muon-NS: CM, $b=1$ & -- & 19.4 $\pm$ 2.2 & 22.3 $\pm$ 1.8 & 29.9 $\pm$ 2.4 \\
Muon-NS: CM+NNM, $b=1$ & -- & 66.9 $\pm$ 4.4 & 83.6 $\pm$ 3.3 & 84.8 $\pm$ 2.3 \\
Muon-NS: CM, $b=2$ & 70.3 $\pm$ 2.6 & 28.0 $\pm$ 1.3 & 60.3 $\pm$ 3.9 & 64.9 $\pm$ 4.1 \\
Muon-NS: CM+NNM, $b=2$ & -- & 72.2 $\pm$ 4.6 & 82.4 $\pm$ 3.2 & 82.1 $\pm$ 2.3 \\
Muon-NS: CM, $b=4$ & -- & 55.9 $\pm$ 5.3 & 73.6 $\pm$ 2.7 & 75.0 $\pm$ 1.5 \\
Muon-NS: CM+NNM, $b=4$ & -- & 76.6 $\pm$ 3.0 & 82.2 $\pm$ 2.5 & 82.3 $\pm$ 1.2 \\
\bottomrule

\end{tabular}}
\end{table}

\begin{figure}[H]
\centering\includegraphics[width=0.7\textwidth]{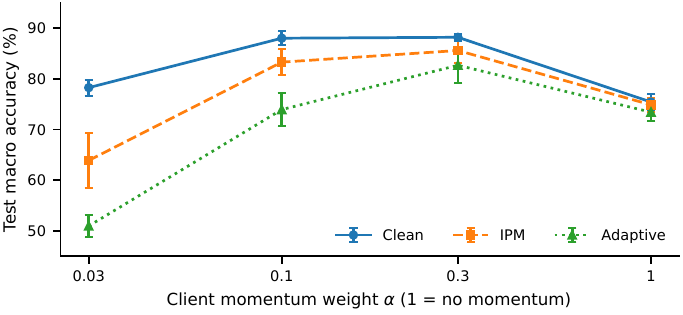}
\caption{ResNet-18 Muon-NS test macro-accuracy versus worker momentum weight (the Muon-NS rows of Table~\ref{tab:temporal}; five-seed means with sample standard deviations). The $\alpha=0.1$ runs use the same configuration as Table~\ref{tab:vision-main}; the small IPM difference (83.3 versus 83.9) comes from three seeds that were resumed from a checkpoint and re-executed, not from a configuration change.}\label{fig:momentum}
\end{figure}

\paragraph{Worker momentum sweep.} Table~\ref{tab:temporal} and Figure~\ref{fig:momentum} vary the worker momentum weight $\alpha$ for SGD, normalized SGD, and Muon-NS under clean training, IPM, and the adaptive attack (five seeds; all other settings as in Table~\ref{tab:vision-main}). The $\alpha=0.1$ column re-executes the primary configuration with identical attack settings; the clean values of all three geometries, all normalized-SGD values, and the Muon-NS adaptive value coincide exactly with Table~\ref{tab:vision-main}, while the SGD IPM (31.8 versus 32.9), SGD adaptive (24.8 versus 25.1), and Muon-NS IPM (83.3 versus 83.9) values differ slightly because some seeds in one suite were resumed from a checkpoint and partially re-executed, i.e., run-to-run nondeterminism rather than a configuration change.

\begin{table}[H]
\centering\small
\caption{ResNet-18 test macro-accuracy (\%) versus worker momentum weight $\alpha$, mean $\pm$ sample standard deviation over five seeds ($\alpha=1$ is no temporal averaging).}\label{tab:temporal}
\resizebox{\textwidth}{!}{\begin{tabular}{llrrrr}
\toprule
Geometry & Attack & $\alpha=0.03$ & $\alpha=0.1$ & $\alpha=0.3$ & $\alpha=1$ \\
\midrule
SGD & Clean & 17.8 $\pm$ 1.8 & 39.8 $\pm$ 2.0 & 50.6 $\pm$ 1.0 & 47.1 $\pm$ 0.9 \\
SGD & IPM & 16.4 $\pm$ 1.9 & 31.8 $\pm$ 3.1 & 42.5 $\pm$ 1.6 & 43.7 $\pm$ 1.1 \\
SGD & Adaptive & 15.1 $\pm$ 1.6 & 24.8 $\pm$ 2.8 & 33.9 $\pm$ 2.4 & 40.9 $\pm$ 1.6 \\
\midrule
Normalized SGD & Clean & 28.6 $\pm$ 1.9 & 49.0 $\pm$ 2.3 & 52.4 $\pm$ 1.5 & 42.3 $\pm$ 1.1 \\
Normalized SGD & IPM & 20.3 $\pm$ 1.7 & 42.3 $\pm$ 2.1 & 46.2 $\pm$ 1.8 & 41.5 $\pm$ 1.5 \\
Normalized SGD & Adaptive & 17.3 $\pm$ 1.7 & 30.4 $\pm$ 1.7 & 37.2 $\pm$ 0.9 & 36.7 $\pm$ 2.5 \\
\midrule
Muon-NS & Clean & 78.3 $\pm$ 1.6 & 88.0 $\pm$ 1.4 & 88.2 $\pm$ 0.7 & 75.5 $\pm$ 1.6 \\
Muon-NS & IPM & 63.9 $\pm$ 5.4 & 83.3 $\pm$ 2.6 & 85.6 $\pm$ 2.5 & 74.9 $\pm$ 1.3 \\
Muon-NS & Adaptive & 51.0 $\pm$ 2.2 & 74.0 $\pm$ 3.3 & 82.7 $\pm$ 3.5 & 73.4 $\pm$ 1.8 \\
\bottomrule

\end{tabular}}
\end{table}

\paragraph{Interpretation limits.} The experiments simulate multi-worker learning on one GPU per trial, not deployment over independent institutions. In Table~\ref{tab:aggregators}, bucketed mean matches or exceeds RFA in every shared mild-attack cell; the stronger fixed-honest and RFA-query controls favor RFA, but it still suffers large losses. Averaging has no worst-case Byzantine guarantee. The older geometry control has a large benign gap, whereas the shared-checkpoint continuation largely removes it; only the sign-flip difference in attack-induced drops has a paired interval excluding zero. AdamW and Muon-NS differ in temporal state, and the ResNet-18 AdamW benign gap limits attribution. Worst-class accuracy can be low even when macro-accuracy is favorable (Table~\ref{tab:worst-class}). The theoretical inexact-oracle guarantee is conditional on an oracle-error bound; five Newton--Schulz iterations alone do not establish that the error is negligible. We report empirical superiority under these budgets and attacks, not universal Byzantine security or a direct measurement of theorem constants.

\paragraph{Controlled theory-facing experiment.} We also evaluate the same Muon-NS update on a $512\times512$ matrix quadratic with 25 simulated workers, heterogeneity parameter $0.25$, stochastic noise $0.02$, 300 rounds, unbucketed RFA, and an adaptive directional attack. Table~\ref{tab:theory-extrapolation} compares ordinary and extrapolated gradient evaluation at matched radius $\eta=0.01$. The final nuclear norm of the true gradient decreases with extrapolation in this controlled setting, including at 20\% corruption. This probes the mechanism behind the second-order extension, but a single finite-budget quadratic cannot establish the asymptotic $\varepsilon$-complexity rate. One-factor-at-a-time sweeps of corruption, heterogeneity, momentum, radius $\eta$, weight decay, and inexact oracle steps around a base configuration (20\% IPM corruption, heterogeneity $0.25$, noise $0.02$, $\alpha=0.1$, $\eta=0.01$, five Newton--Schulz steps; with a few two-factor crosses), together with a momentum $\times$ noise $\times$ corruption grid on a smaller $256\times256$ instance run for 2000 rounds, are preserved as per-seed summaries and available from the authors on request.

\begin{table}[H]
\centering\small
\caption{Final gradient nuclear norm (lower is better) on the controlled matrix quadratic; five seeds, mean $\pm$ sample standard deviation. Byzantine fractions are nominal: $0.1$ corresponds to two of 25 workers (8\%).}\label{tab:theory-extrapolation}
\begin{tabular}{lrr}
\toprule
Byzantine fraction & Ordinary & Extrapolated \\
\midrule
0\% & 18.15 $\pm$ 0.04 & 9.06 $\pm$ 0.02 \\
10\% & 18.28 $\pm$ 0.03 & 9.43 $\pm$ 0.02 \\
20\% & 19.18 $\pm$ 0.03 & 13.80 $\pm$ 0.01 \\
\bottomrule

\end{tabular}
\end{table}

\paragraph{Certified aggregation control.} We additionally test Lemma~\ref{lem:smoothed-median} on the same $512\times512$ quadratic family, with 20 honest workers, five benign copies of their mean or five adaptive directional messages (strength $1$), additive center heterogeneity $0.25$, coordinate-wise Gaussian noise standard deviation $0.001$, worker momentum $\alpha=0.1$, radius $\eta=0.001$, and an exact polar update for 300 rounds. All honest Hessians are the identity, so $B=0$ and the operator/nuclear smoothness constant may be taken as $L=512$. The server uses unbucketed Euclidean smoothed RFA with $\nu=10^{-6}$: it performs the original eight Weiszfeld iterations and continues only if the objective-gap certificate of Lemma~\ref{lem:smoothed-median} exceeds the threshold corresponding to $\varepsilon_k^{\mathrm{agg}}=10^{-3}$ after conversion with $\kappa=\sqrt{512}$. Over five seeds per condition, all 3000 round outputs pass after exactly eight iterations and coincide with the original eight-iteration RFA output on the same messages. The largest certified additive residual is $1.90\times10^{-5}$ in the nuclear norm; Table~\ref{tab:certified-rfa} reports the objective certificates and final gradient norms. The complete round summaries and the stopping-rule implementation accompany the editable source.

\begin{table}[H]
\centering\small
\caption{Certified unbucketed smoothed RFA on the controlled matrix quadratic. Final gradient nuclear norm is the mean $\pm$ sample standard deviation over five seeds; $\tau_k$ and $\varepsilon_k^{\mathrm{agg}}$ are maxima over all 1500 rounds per condition. Every round uses eight Weiszfeld iterations.}\label{tab:certified-rfa}
\begin{tabular}{lrrrr}
\toprule
Messages & Final $\|\nabla f(x^K)\|_{\mathrm{nuc}}$ & Max. $\tau_k$ & Max. $\varepsilon_k^{\mathrm{agg}}$ & Iterations \\
\midrule
Benign & $1.411 \pm 0.001$ & $1.53\times 10^{-12}$ & $1.89\times 10^{-5}$ & 8 \\
Directional & $1.942 \pm 0.003$ & $1.15\times 10^{-7}$ & $1.90\times 10^{-5}$ & 8 \\

\end{tabular}
\end{table}

\section{Analysis for Non-Convex Functions}\label{apx:nonconvex-proof}
In this appendix we prove Theorem~\ref{thm:main-nonconvex} for Algorithm~\ref{alg:byz-muon}. The analysis follows the non-Euclidean trust-region viewpoint of~\citet{kovalev2025understanding} and the momentum-based Byzantine-robust framework of~\citet{karimireddy2021learning, Karimireddy2020ByzantineRobustLO}.

\paragraph{Notation and filtration.}
Throughout the appendices we write $q := 1-\alpha \in [0,1)$ and, for a good worker $i\in\mathcal{G}$ and $k \ge 0$, we denote the individual and averaged stochastic gradient noises by
\[
\xi_i^{k} := g_i(x^{k};\xi_i^{k}) - \nabla f_i(x^{k}), \qquad
\bar{\xi}^{k} := \frac{1}{G}\sum_{i\in\mathcal{G}} \xi_i^{k}
\]
(we slightly abuse notation and write $\xi_i^k$ both for the random seed and for the induced noise vector). Recall that $\mathcal{F}^{k}$ is the $\sigma$-algebra generated by the gradient draws $\{\xi_i^{s}\}_{i\in\mathcal{G}, s \le k}$, the Byzantine messages and aggregator randomness through $m^{k}$, and the selection randomness of the (measurably selected) trust-region solver through $x^{k+1}$, with $\mathcal{F}^{-1}$ trivial; thus $x^{k+1}$ is $\mathcal{F}^{k}$-measurable, while the round-$(k+1)$ draws $\xi_i^{k+1}$ are not. Assumption~\ref{asn:stoch-grad} states that, for all $k \ge 0$,
\begin{equation}\label{eq:mds}
\mathbb{E}[\xi_i^{k} \mid \mathcal{F}^{k-1}] = 0, \qquad \mathbb{E}[\|\xi_i^{k}\|_2^2 \mid \mathcal{F}^{k-1}] \le \sigma^2,
\end{equation}
with conditional independence across good workers ($x^0$ is deterministic, so the $k=0$ instance is unconditional); in particular, each sequence $(\xi_i^{k})_{k\ge0}$ is a martingale-difference sequence. We denote the ideal (good-worker) average momentum and the ideal momentum error by
\[
\bar m^{k} := \frac{1}{G}\sum_{i\in\mathcal{G}} m_i^{k}, \qquad
\bar e^{k} := \bar m^{k} - \nabla f(x^{k}),
\]
abbreviate $g_k := \mathbb{E}\|\nabla f(x^k)\|_*$, and recall the noise scales $V_G, V_P$ defined in Section~\ref{sec:theory}. The initialization of Algorithm~\ref{alg:byz-muon} gives $m_i^0 = \nabla f_i(x^0) + \xi_i^0$ for $i \in \mathcal{G}$, hence $\bar m^0 = \nabla f(x^0) + \bar\xi^0$ and $\bar e^0 = \bar\xi^0$. Since $x^0$ is deterministic, $g_0 = \|\nabla f(x^0)\|_*$.

\begin{remark}[Relation to second-moment Euclidean aggregators]\label{rem:aggregators}
The aggregation rules used in practice are analyzed by \citet{karimireddy2021learning, Karimireddy2020ByzantineRobustLO} under a second-moment Euclidean guarantee, stated with a per-pair dispersion premise and extended to the average pairwise premise we use here, $\frac{1}{G(G-1)}\sum_{i \neq l} \mathbb{E}\|z_i - z_l\|_2^2 \le \bar{\rho}^2$ implying $\mathbb{E}\|\hat{z} - \bar{z}\|_2^2 \le c\delta \bar{\rho}^2$, by \citet{gorbunov2022variance}. Plain Krum, geometric median, and coordinate-wise median do not satisfy this criterion without qualification; bucketed variants have guarantees under the stated corruption, bucket-size, and dimension conditions \citep{Karimireddy2020ByzantineRobustLO}. The cited centered-clipping guarantee uses a dispersion-dependent radius and is not agnostic. By Jensen's inequality and Assumption~\ref{asn:norm-equivalence}, any such aggregator obeys a first-moment dual-norm guarantee of the form $\mathbb{E}\|\hat{z} - \bar{z}\|_* \le \kappa\, \mathbb{E}\|\hat{z} - \bar{z}\|_2 \le \kappa\sqrt{\mathbb{E}\|\hat{z} - \bar{z}\|_2^2} \le \kappa\sqrt{c\delta}\,\bar{\rho}$, a dual-norm output-error bound under a separate second-moment dispersion premise, rather than a literal instance of Definition~\ref{def:aragg}. Under this distinct premise, the corresponding guarantees are proved in Appendix~\ref{apx:second-moment}: carrying the dispersion recursion in Euclidean second moments (Lemma~\ref{lem:dispersion2}), under Assumption~\ref{asn:stoch-grad} (already second-moment) and the second-moment additive heterogeneity Assumption~\ref{asn:het2}, yields Theorem~\ref{thm:second-moment} (non-convex) and Corollary~\ref{cor:starconvex2} (star-convex): the heterogeneity floor $4c\delta\zeta$ becomes $4\sqrt{2}\,\sqrt{c\delta}\,\kappa\zeta_2$ and the Byzantine variance term $2c\delta V_P$ becomes $2\sqrt{2}\,\sqrt{c\delta}\, V_P$ (the conversion factor $\kappa$ enters once, through Jensen); the relative heterogeneity component is discussed in Remark~\ref{rem:relative2}.

We also caution that the first-moment factor $c\delta$ of Definition~\ref{def:aragg} cannot, in general, be made small. Consider $(1-\delta)n$ inputs at $0$ and $\delta n$ inputs at a point $D \neq 0$. If the good set is the zero cluster, the dispersion is $r = 0$ and the definition forces $\hat{z} = 0$ almost surely. If instead the good set consists of $(1-2\delta)n$ of the zeros together with the $\delta n$ inputs at $D$ --- the same input multiset, so the aggregator's output is unchanged --- then $\bar{z} = \frac{\delta}{1-\delta} D$, while the average pairwise dispersion tends to $\frac{2\delta(1-2\delta)}{(1-\delta)^2}\,\|D\|_*$ as $n \to \infty$; the definition then requires
\[
\frac{\delta}{1-\delta}\|D\|_* \le c\delta \cdot \frac{2\delta(1-2\delta)}{(1-\delta)^2}\,\|D\|_*,
\qquad \text{i.e.,} \qquad
c\delta \ge \frac{1-\delta}{2(1-2\delta)} \ge \frac{1}{2}.
\]
The first-moment guarantee is, however, \emph{realizable} in any finite-dimensional normed space, albeit with a factor that does not vanish as $\delta\to0$ through positive values. Indeed, for $0<\delta<1/2$, choose an exact geometric median of all inputs in the dual norm,
\[
\hat z\in\arg\min_{u\in\mathcal X}\sum_{i=1}^n\|u-z_i\|_*.
\]
For each realization, let $b=n-G$, $\bar z=G^{-1}\sum_{i\in\mathcal G}z_i$, and
$r_{\mathrm{obs}}=[G(G-1)]^{-1}\sum_{i\ne j\in\mathcal G}\|z_i-z_j\|_*$.
Comparing the geometric-median objective at $\hat z$ and $\bar z$ and applying the triangle inequality separately to good and Byzantine inputs yields
\[
(G-b)\|\hat z-\bar z\|_*
\le 2\sum_{i\in\mathcal G}\|z_i-\bar z\|_*
\le 2(G-1)r_{\mathrm{obs}}.
\]
Because $b/G\le\delta/(1-\delta)$, we obtain
$\|\hat z-\bar z\|_*\le[2(1-\delta)/(1-2\delta)]r_{\mathrm{obs}}$
and hence, after expectation, Definition~\ref{def:aragg} with
$c\delta=2(1-\delta)/(1-2\delta)$. At $\delta=0.2$, this sufficient factor is $8/3$, so the non-convex condition $4c\delta B<1$ requires $B<3/32$. For $\delta=0$, the server can instead take the arithmetic mean, which has zero error. Thus the main results have a concrete, agnostic first-moment instantiation; exact dual-norm geometric medians may nevertheless be computationally demanding, particularly for the nuclear norm. The Euclidean smoothed RFA implementation used in our experiments is not asserted to attain this precise guarantee. The separate second-moment route in Appendix~\ref{apx:second-moment} gives a more favorable dependence on the Byzantine fraction under its stronger assumptions. Sharpening the constants achievable in non-Euclidean geometries (e.g., avoiding the dimension-dependent factor $\kappa$ for the nuclear norm arising in the Muon geometry) remains an interesting direction.
\end{remark}

\begin{samepage}
\begin{lemma}[Smoothed geometric median with an objective certificate]\label{lem:smoothed-median}
Let $n=G+b$ with $G>b$, and write $r_{\mathrm{obs}}=[G(G-1)]^{-1}\sum_{i\ne j\in\mathcal G}\|z_i-z_j\|_*$. Define $h_\nu(t)=t$ for $t\ge\nu$ and $h_\nu(t)=t^2/(2\nu)+\nu/2$ otherwise, and $F_\nu(u)=\sum_{i=1}^n h_\nu(\|u-z_i\|_*)$. If $F_\nu(\hat z)\le\inf_u F_\nu(u)+\tau$, then
\[
\|\hat z-\bar z\|_*\le \frac{2(G-1)}{G-b}\,r_{\mathrm{obs}}+\frac{n\nu/2+\tau}{G-b}.
\]
Consequently, a uniform certificate $\tau\le\bar\tau$ gives the additive residual $\varepsilon^{\mathrm{agg}}=(n\nu/2+\bar\tau)/(G-b)$ in Remark~\ref{rem:approx-aggregation}; the multiplicative factor is bounded by $2(1-\delta)/(1-2\delta)$ when $G\ge(1-\delta)n$.
\end{lemma}
\begin{proof}
Writing $F(u)=\sum_i\|u-z_i\|_*$, the inequalities $0\le h_\nu(t)-t\le\nu/2$ imply $F(\hat z)\le F(\bar z)+n\nu/2+\tau$. The triangle inequality, applied separately to good and Byzantine inputs, gives $(G-b)\|\hat z-\bar z\|_*\le 2\sum_{i\in\mathcal G}\|z_i-\bar z\|_*+n\nu/2+\tau$. Finally, $\sum_{i\in\mathcal G}\|z_i-\bar z\|_*\le(G-1)r_{\mathrm{obs}}$.
\end{proof}
\end{samepage}
The extensions in Remark~\ref{rem:approx-aggregation} follow by inserting the residual into Lemma~\ref{lem:descent_exp_rigorous}: it contributes $2\eta\varepsilon_k^{\mathrm{agg}}$ per step, while the momentum and heterogeneity recursions are unchanged. The star-convex recursion is modified by the same additive term.
For the Euclidean smoothing used by our implementation, the same argument applies in $\ell_2$. In the product Muon geometry of Section~\ref{sec:theory}, $\|v\|_2\le\|v\|_*\le\kappa\|v\|_2$, so conversion to the dual norm multiplies both terms of the Euclidean bound by at most $\kappa$. In the Euclidean case the median minimizer and every Weiszfeld iterate lie in the convex hull of the messages. Writing $\bar F_\nu=F_\nu/n$, convexity permits the computable choice $\tau=n\big[\langle\nabla\bar F_\nu(\hat z),\hat z\rangle-\min_i\langle\nabla\bar F_\nu(\hat z),z_i\rangle\big]$. A fixed number of Weiszfeld iterations is not itself an objective certificate; the controlled experiment in Appendix~\ref{apx:experiment-details} uses this stopping rule and records the resulting residual.

We now state the supporting lemmas and then the main theorem.

\begin{lemma}[Trust-region optimality]
\label{lem:tr_opt}
Let $z \in \mathbb{R}^d$ and $\eta>0$. Consider the trust-region linear subproblem
\[
z_+ := \arg\min_{\|x-z\|\le \eta} \langle v, x \rangle
\]
for some $v \in \mathbb{R}^d$. Let $\|\cdot\|$ be any norm with dual $\|\cdot\|_*$. Then
\[
\langle v, z - z_+ \rangle = \eta \|v\|_*.
\]
\end{lemma}

\begin{proof}
First, we reparameterize the problem by shifting it. Let $u := x-z$. The optimization problem is then equivalent to finding $u_+$ such that
\[
u_+ := \arg\min_{\|u\|\le \eta} \langle v, u \rangle,
\]
where $z_+ = z + u_+$.

Next, recall that the dual norm is defined as $\|v\|_* := \max_{\|u\|\le 1} \langle v, u \rangle$. By symmetry and scaling the constraint set by $\eta$, the minimum value over the ball satisfies
\[
\langle v, u_+ \rangle = \min_{\|u\|\le \eta} \langle v, u \rangle = - \eta \|v\|_*.
\]

Finally, we substitute this back to compute the desired inner product. Since $z - z_+ = - u_+$, we obtain
\[
\langle v, z - z_+ \rangle = - \langle v, u_+ \rangle = \eta \|v\|_*.
\]
When the minimizer is not unique (e.g., for a rank-deficient $v$ in the operator-norm geometry), every minimizer attains the same optimal value $-\eta\|v\|_*$, so the identity holds for any (measurable) selection. This completes the proof.
\end{proof}

\begin{lemma}[Descent bound]
\label{lem:descent_tr_rigorous}
Let $f:\mathbb{R}^d \to \mathbb{R}$ be $L$-smooth with respect to a norm $\|\cdot\|$ and dual $\|\cdot\|_*$ (Assumption~\ref{asn:lipschitz} holds). Let $x^{k+1}$ be obtained from the trust-region linear update
\[
x^{k+1} = \arg\min_{\|x-x^k\|\le \eta} \langle m^{k}, x \rangle,
\]
where $m^{k}$ is any vector (e.g., a robustly aggregated momentum). Then
\[
f(x^{k+1}) \le f(x^k) - \eta \|\nabla f(x^{k})\|_* + 2\eta \|\nabla f(x^{k}) - m^{k}\|_* + \frac{L \eta^2}{2}.
\]
\end{lemma}
\begin{proof}
Using the $L$-smoothness of $f$ and the trust-region constraint $\|x^{k+1}-x^k\| \le \eta$, we start with the standard upper bound:
\[
f(x^{k+1}) \le f(x^k) + \langle \nabla f(x^k), x^{k+1}-x^k \rangle + \frac{L\eta^2}{2}.
\]
To analyze the linear term, we decompose the inner product by adding and subtracting $m^{k}$:
\[
\langle \nabla f(x^k), x^{k+1}-x^k \rangle
= \langle m^{k}, x^{k+1}-x^k \rangle + \langle \nabla f(x^k)-m^{k}, x^{k+1}-x^k \rangle.
\]
Applying Lemma~\ref{lem:tr_opt} with $v=m^{k}$ and $z=x^k$, the minimization over the trust region yields
\[
\langle m^{k}, x^{k+1}-x^k \rangle = - \langle m^{k}, x^k - x^{k+1} \rangle = - \eta \|m^{k}\|_*.
\]
We can lower bound the norm of the momentum using the reverse triangle inequality, $$\|m^{k}\|_* \ge \|\nabla f(x^{k})\|_* - \|\nabla f(x^{k}) - m^{k}\|_*.$$ Multiplying this by $-\eta$ gives
\[
- \eta \|m^{k}\|_* \le -\eta \|\nabla f(x^{k})\|_* + \eta \|\nabla f(x^{k}) - m^{k}\|_*.
\]
Meanwhile, the error term in the decomposition is bounded using the definition of the dual norm and the trust-region constraint:
\[
\langle \nabla f(x^k)-m^{k}, x^{k+1}-x^k \rangle \le \|\nabla f(x^k)-m^{k}\|_* \cdot \|x^{k+1}-x^k\| \le \eta \|\nabla f(x^k)-m^{k}\|_*.
\]
Finally, substituting these bounds back into the smoothness inequality yields
\begin{align*}
f(x^{k+1}) &\le f(x^k) - \eta \|\nabla f(x^{k})\|_* + \eta \|\nabla f(x^{k}) - m^{k}\|_* + \eta \|\nabla f(x^k)-m^{k}\|_* + \frac{L\eta^2}{2} \\
&= f(x^k) - \eta \|\nabla f(x^{k})\|_* + 2\eta \|\nabla f(x^{k}) - m^{k}\|_* + \frac{L\eta^2}{2},
\end{align*}
which completes the proof.
\end{proof}
The server's robust aggregate momentum is $m^{k} := \mathrm{ARAGG}(m_1^{k},\dots,m_n^{k})$ for every $k \ge 0$, and the ideal average momentum is $\bar m^{k} := \frac{1}{G}\sum_{i\in \mathcal{G}} m_i^{k}$. We first record an exact representation of the good-worker momenta as convex combinations of past stochastic gradients.

\begin{lemma}[Momentum representation]\label{lem:momentum-repr}
For every good worker $i \in \mathcal{G}$ and every $k \ge 0$, the iterates of Algorithm~\ref{alg:byz-muon} (and of Algorithm~\ref{alg:byz-muon-decay}) satisfy
\[
m_i^{k} = \sum_{s=0}^{k} w_{k,s}\left(\nabla f_i(x^{s}) + \xi_i^{s}\right),
\qquad
w_{k,0} := q^{k}, \quad w_{k,s} := \alpha q^{k-s} \ \ (1 \le s \le k),
\]
and the weights form a convex combination: $w_{k,s} \ge 0$ and $\sum_{s=0}^{k} w_{k,s} = 1$ (throughout, we use the convention $0^0 := 1$, which covers the case $\alpha = 1$, $q = 0$). Moreover,
\begin{equation}\label{eq:weight-l2}
\sum_{s=0}^{k} w_{k,s}^2 \;\le\; q^{2k} + \frac{\alpha}{2-\alpha}.
\end{equation}
\end{lemma}
\begin{proof}
We proceed by induction over $k$. For $k=0$ the claim reads $m_i^0 = \nabla f_i(x^0)+\xi_i^0 = g_i(x^0;\xi_i^0)$, which is exactly the initialization. Assume the representation holds for some $k \ge 0$. Then, by the momentum update,
\begin{align*}
m_i^{k+1} = q\, m_i^{k} + \alpha\, g_i^{k+1}
= \sum_{s=0}^{k} q\, w_{k,s} \left(\nabla f_i(x^s)+\xi_i^s\right) + \alpha\left(\nabla f_i(x^{k+1})+\xi_i^{k+1}\right),
\end{align*}
and the new weights are $q\, w_{k,0} = q^{k+1} = w_{k+1,0}$, $\;q\, w_{k,s} = \alpha q^{(k+1)-s} = w_{k+1,s}$ for $1\le s\le k$, and $\alpha = w_{k+1,k+1}$, which is the representation at $k+1$. The weights are nonnegative and sum to
\[
q^k + \alpha\sum_{s=1}^k q^{k-s} = q^k + \alpha\,\frac{1-q^k}{1-q} = q^k + (1-q^k) = 1.
\]
Finally, using $1-q^2 = (1-q)(1+q) = \alpha(2-\alpha)$,
\[
\sum_{s=0}^{k} w_{k,s}^2 = q^{2k} + \alpha^2 \sum_{s=1}^{k} q^{2(k-s)} \le q^{2k} + \frac{\alpha^2}{1-q^2} = q^{2k} + \frac{\alpha}{2-\alpha}. \qedhere
\]
\end{proof}

\begin{lemma}[Weighted noise bounds]\label{lem:noise}
Let Assumptions~\ref{asn:stoch-grad} and~\ref{asn:norm-equivalence} hold, fix $k\ge0$, and let $w_{k,s}$ be the weights of Lemma~\ref{lem:momentum-repr}. Then
\begin{align}
\mathbb{E}\bigg\|\sum_{s=0}^{k} w_{k,s}\,\bar\xi^{s}\bigg\|_* &\le \frac{\kappa\sigma}{\sqrt{G}}\left(q^{k} + \sqrt{\frac{\alpha}{2-\alpha}}\right) = \frac{\kappa\sigma}{\sqrt{G}}\, q^{k} + V_G, \label{eq:noise-avg}\\
\mathbb{E}\bigg\|\sum_{s=0}^{k} w_{k,s}\,(\xi_i^{s}-\xi_j^{s})\bigg\|_* &\le \sqrt{2}\,\kappa\sigma\left(q^{k} + \sqrt{\frac{\alpha}{2-\alpha}}\right) = \sqrt{2}\,\kappa\sigma\, q^{k} + V_P \qquad \text{for all } i\neq j \in\mathcal{G}. \label{eq:noise-pair}
\end{align}
\end{lemma}
\begin{proof}
We first carry out all second-moment computations in the Euclidean norm, where conditional independence can be exploited, and only afterwards convert to the dual norm via Assumption~\ref{asn:norm-equivalence}. We emphasize that the analogous expansion performed directly in a general dual norm would be invalid: $\|\cdot\|_*^2$ does not expand into inner-product cross terms unless the norm is induced by the inner product (in particular, not for the nuclear norm).

\emph{Step 1 (variance of the averaged noise).} Fix $s \ge 0$. By conditional independence across good workers and the conditional zero mean in \eqref{eq:mds}, the cross terms vanish:
\[
\mathbb{E}\big[\|\bar\xi^{s}\|_2^2 \,\big|\, \mathcal{F}^{s-1}\big]
= \frac{1}{G^2}\sum_{i,j\in\mathcal{G}} \mathbb{E}\big[\langle \xi_i^{s},\xi_j^{s}\rangle \,\big|\, \mathcal{F}^{s-1}\big]
= \frac{1}{G^2}\sum_{i\in\mathcal{G}} \mathbb{E}\big[\|\xi_i^{s}\|_2^2 \,\big|\, \mathcal{F}^{s-1}\big]
\le \frac{\sigma^2}{G}.
\]
\emph{Step 2 (orthogonality across iterations).} For $s < t$, the vector $\bar\xi^{s}$ is $\mathcal{F}^{t-1}$-measurable, so by the tower property
\[
\mathbb{E}\langle \bar\xi^{s}, \bar\xi^{t}\rangle = \mathbb{E}\big\langle \bar\xi^{s},\, \mathbb{E}[\bar\xi^{t}\mid\mathcal{F}^{t-1}]\big\rangle = 0.
\]
Hence, expanding the Euclidean square and using~\eqref{eq:weight-l2},
\[
\mathbb{E}\bigg\|\sum_{s=0}^{k} w_{k,s}\bar\xi^{s}\bigg\|_2^2 = \sum_{s=0}^{k} w_{k,s}^2\, \mathbb{E}\|\bar\xi^{s}\|_2^2 \le \frac{\sigma^2}{G}\sum_{s=0}^k w_{k,s}^2 \le \frac{\sigma^2}{G}\left(q^{2k}+\frac{\alpha}{2-\alpha}\right).
\]
\emph{Step 3 (conversion to the dual norm).} By Assumption~\ref{asn:norm-equivalence} and Jensen's inequality,
\[
\mathbb{E}\bigg\|\sum_{s=0}^k w_{k,s}\bar\xi^{s}\bigg\|_* \le \kappa\, \mathbb{E}\bigg\|\sum_{s=0}^k w_{k,s}\bar\xi^{s}\bigg\|_2 \le \kappa \sqrt{\mathbb{E}\bigg\|\sum_{s=0}^k w_{k,s}\bar\xi^{s}\bigg\|_2^2} \le \frac{\kappa\sigma}{\sqrt{G}}\sqrt{q^{2k}+\frac{\alpha}{2-\alpha}},
\]
and \eqref{eq:noise-avg} follows from $\sqrt{a+b}\le\sqrt{a}+\sqrt{b}$ for $a, b \ge 0$. The pairwise bound \eqref{eq:noise-pair} follows by the identical argument applied to the differences $\xi_i^{s}-\xi_j^{s}$, whose conditional second moments satisfy
\[
\mathbb{E}\big[\|\xi_i^{s}-\xi_j^{s}\|_2^2\mid\mathcal{F}^{s-1}\big] = \mathbb{E}\big[\|\xi_i^{s}\|_2^2\mid\mathcal{F}^{s-1}\big] + \mathbb{E}\big[\|\xi_j^{s}\|_2^2\mid\mathcal{F}^{s-1}\big] \le 2\sigma^2,
\]
again because the cross term vanishes by conditional independence, and which are martingale differences across iterations by the same tower-property argument.
\end{proof}

\begin{lemma}[Ideal momentum error]
\label{lem:error_recursion_final}
Let Assumptions~\ref{asn:stoch-grad}--\ref{asn:norm-equivalence} hold. Fix $k \ge 0$ and suppose that the iterates satisfy $\|x^{s}-x^{s-1}\|\le d$ almost surely for all $1 \le s \le k$. Then
\[
\mathbb{E} \|\bar{e}^k\|_* \;\le\; \frac{q\,(1-q^{k})\,L d}{\alpha} + \frac{\kappa\sigma}{\sqrt{G}}\,q^{k} + V_G
\;\le\; \frac{q L d}{\alpha} + \frac{\kappa\sigma}{\sqrt{G}}\,q^{k} + V_G.
\]
For Algorithm~\ref{alg:byz-muon} the trust-region constraint gives $d=\eta$.
\end{lemma}

\begin{proof}
\emph{Step 1 (error recursion).} Averaging the momentum update over $i\in\mathcal{G}$ gives, for $k\ge1$,
\[
\bar m^{k} = q\,\bar m^{k-1} + \alpha\big(\nabla f(x^{k}) + \bar\xi^{k}\big).
\]
Therefore,
\begin{align*}
\bar e^{k} = \bar m^{k} - \nabla f(x^{k})
&= q\,\bar m^{k-1} - (1-\alpha)\nabla f(x^{k}) + \alpha\bar\xi^{k} \\
&= q\big(\bar m^{k-1} - \nabla f(x^{k-1})\big) + q\big(\nabla f(x^{k-1}) - \nabla f(x^{k})\big) + \alpha\bar\xi^{k} \\
&= q\,\bar e^{k-1} + q\big(\nabla f(x^{k-1}) - \nabla f(x^{k})\big) + \alpha\bar\xi^{k}.
\end{align*}
\emph{Step 2 (unrolling).} Since $\bar e^{0} = \bar\xi^{0}$ by the initialization, unrolling the recursion yields, for every $k\ge0$,
\[
\bar e^{k} = \sum_{s=1}^{k} q^{\,k-s+1}\big(\nabla f(x^{s-1}) - \nabla f(x^{s})\big) + \sum_{s=0}^{k} w_{k,s}\,\bar\xi^{s},
\]
where the noise terms collect exactly the weights of Lemma~\ref{lem:momentum-repr}: the initial noise $\bar\xi^0$ carries weight $q^{k}=w_{k,0}$ and, for $s \ge 1$, the noise $\bar\xi^{s}$ carries weight $\alpha q^{k-s}=w_{k,s}$.

\emph{Step 3 (bounding the two parts).} For the drift part, Assumption~\ref{asn:lipschitz} and the step bound give $\|\nabla f(x^{s-1})-\nabla f(x^{s})\|_* \le L\|x^{s-1}-x^{s}\| \le L d$, hence by the triangle inequality
\[
\bigg\|\sum_{s=1}^{k} q^{\,k-s+1}\big(\nabla f(x^{s-1}) - \nabla f(x^{s})\big)\bigg\|_* \le L d \sum_{s=1}^{k} q^{\,k-s+1} = L d\,\frac{q(1-q^{k})}{\alpha}.
\]
The noise part is bounded by \eqref{eq:noise-avg} of Lemma~\ref{lem:noise}. Combining the two bounds and using $1 - q^k \le 1$ completes the proof.
\end{proof}

\begin{lemma}[Descent in expectation with aggregation error]
\label{lem:descent_exp_rigorous}
Let $f$ be $L$-smooth with respect to a norm $\|\cdot\|$ and dual $\|\cdot\|_*$ (Assumption~\ref{asn:lipschitz} holds). For any $k\ge0$, the iterates of Algorithm~\ref{alg:byz-muon} satisfy
\begin{equation}
\mathbb{E}[f(x^{k+1})] \le \mathbb{E}[f(x^k)] - \eta \mathbb{E}\|\nabla f(x^{k})\|_* + 2\eta\,\mathbb{E}\|\bar{e}^{k}\|_* + 2\eta\,\mathbb{E}\|\bar{m}^{k} - m^{k}\|_* + \frac{L \eta^2}{2},
\end{equation}
where $\bar{e}^{k} := \bar{m}^{k} - \nabla f(x^k)$ is the ideal momentum error, and the expectation is over all randomness of the algorithm (the $f$- and gradient-terms are integrable since $\|x^k - x^0\| \le k\eta$ deterministically; the aggregation-error term is nonnegative, so its expectation is well defined, and it is finite by Lemma~\ref{lem:agg_error_rigorous}).
\end{lemma}

\begin{proof}
We begin by invoking the descent bound established in Lemma~\ref{lem:descent_tr_rigorous}. Applied to the step from $x^k$ to $x^{k+1}$, the pointwise inequality states
\[
f(x^{k+1}) \le f(x^k) - \eta \|\nabla f(x^{k})\|_* + 2\eta \|\nabla f(x^{k}) - m^{k}\|_* + \frac{L \eta^2}{2}.
\]
Taking the total expectation over all randomness up to time $k+1$ yields the base inequality
\begin{equation}
\label{eq:descent_exp_base}
\mathbb{E}[f(x^{k+1})] \le \mathbb{E}[f(x^k)] - \eta \mathbb{E}\|\nabla f(x^{k})\|_* + 2\eta\,\mathbb{E}\|\nabla f(x^{k}) - m^{k}\|_* + \frac{L \eta^2}{2}.
\end{equation}
Next, we isolate the error term $\mathbb{E}\|\nabla f(x^{k}) - m^{k}\|_*$. Using the triangle inequality and introducing the ideal average momentum $\bar{m}^{k}$, we decompose this error as
\begin{equation}
\label{eq:grad_err_decomp1}
\|\nabla f(x^{k}) - m^{k}\|_* \le \|\nabla f(x^{k}) - \bar{m}^{k}\|_* + \|\bar{m}^{k} - m^{k}\|_* = \|\bar{e}^k\|_* + \|\bar{m}^{k} - m^{k}\|_*.
\end{equation}
Finally, substituting this decomposition back into the main expectation inequality \eqref{eq:descent_exp_base} leads directly to the result:
\begin{align*}
\mathbb{E}[f(x^{k+1})] &\le \mathbb{E}[f(x^k)] - \eta \mathbb{E}\|\nabla f(x^{k})\|_* + 2\eta\,\mathbb{E}\left[\|\bar{e}^{k}\|_* + \|\bar{m}^{k} - m^{k}\|_*\right] + \frac{L \eta^2}{2} \\
&= \mathbb{E}[f(x^k)] - \eta \mathbb{E}\|\nabla f(x^{k})\|_* + 2\eta\,\mathbb{E}\|\bar{e}^{k}\|_* + 2\eta\,\mathbb{E}\|\bar{m}^{k} - m^{k}\|_* + \frac{L \eta^2}{2}.
\end{align*}
This completes the proof.
\end{proof}

\begin{lemma}[Momentum dispersion and aggregation error]
\label{lem:agg_error_rigorous}
Let Assumptions~\ref{asn:stoch-grad}, \ref{asn:norm-equivalence}, and~\ref{asn:bounded-heterogeneity} hold, and let the server aggregate satisfy Definition~\ref{def:aragg} with $G \ge \max\{2, (1-\delta)n\}$. Define the average pairwise momentum dispersion and the relative-heterogeneity history
\[
r_k := \frac{1}{G(G-1)}\sum_{\substack{i,j\in\mathcal{G}\\ i\neq j}} \mathbb{E}\|m_i^{k} - m_j^{k}\|_*,
\qquad
S_k := \alpha \sum_{s=1}^{k} q^{\,k-s}\, g_s \quad (\text{so that } S_0 = 0).
\]
Then, for every $k \ge 0$,
\begin{equation}\label{eq:dispersion}
r_k \;\le\; 2\zeta + 2 B S_k + \Big(\sqrt{2}\,\kappa\sigma + 2B\|\nabla f(x^0)\|_*\Big)\, q^{k} + V_P,
\end{equation}
and consequently, since the good momenta satisfy the premise of Definition~\ref{def:aragg} with dispersion level $r_k$,
\begin{equation}\label{eq:aggerr}
\mathbb{E}\|m^{k}-\bar m^{k}\|_* \;\le\; c\delta\, r_k \;\le\; c\delta\left[\,2\zeta + 2 B S_k + \Big(\sqrt{2}\,\kappa\sigma + 2B\|\nabla f(x^0)\|_*\Big)\, q^{k} + V_P\right].
\end{equation}
The history sequence obeys the recursion $S_k = q\, S_{k-1} + \alpha\, g_k$ for $k\ge1$. The bounds hold for Algorithm~\ref{alg:byz-muon} and Algorithm~\ref{alg:byz-muon-decay} alike, since they involve only the momentum recursion and not the form of the iterate update.
\end{lemma}

\begin{proof}
\emph{Step 1 (pairwise decomposition).} By Lemma~\ref{lem:momentum-repr}, for any two good workers $i\neq j\in\mathcal{G}$,
\[
m_i^{k} - m_j^{k} = \sum_{s=0}^{k} w_{k,s}\big(\nabla f_i(x^{s})-\nabla f_j(x^{s})\big) + \sum_{s=0}^{k} w_{k,s}\big(\xi_i^{s}-\xi_j^{s}\big),
\]
so that, by the triangle inequality,
\[
\mathbb{E}\|m_i^{k}-m_j^{k}\|_* \le \sum_{s=0}^{k} w_{k,s}\, \mathbb{E}\|\nabla f_i(x^{s})-\nabla f_j(x^{s})\|_* + \mathbb{E}\bigg\|\sum_{s=0}^{k} w_{k,s}(\xi_i^{s}-\xi_j^{s})\bigg\|_*.
\]

\emph{Step 2 (heterogeneity term).} For every fixed $x \in \mathcal{X}$, inserting $\nabla f(x)$ and applying the triangle inequality followed by Assumption~\ref{asn:bounded-heterogeneity},
\[
\frac{1}{G(G-1)}\sum_{\substack{i,j\in\mathcal{G}\\ i \neq j}} \|\nabla f_i(x)-\nabla f_j(x)\|_* \le \frac{2}{G}\sum_{i\in\mathcal{G}} \|\nabla f_i(x)-\nabla f(x)\|_* \le 2\zeta + 2B\|\nabla f(x)\|_*.
\]
We emphasize that only this \emph{average} pairwise bound is available: Assumption~\ref{asn:bounded-heterogeneity} controls the average deviation over good workers and does not imply a bound for each individual pair, which is why Definition~\ref{def:aragg} is formulated with the average pairwise dispersion. Applying this bound pointwise at $x = x^{s}$, taking expectations, and summing against the weights,
\begin{align*}
\frac{1}{G(G-1)}\sum_{\substack{i,j\in\mathcal{G}\\ i\neq j}}\ \sum_{s=0}^{k} w_{k,s}\, \mathbb{E}\|\nabla f_i(x^{s})-\nabla f_j(x^{s})\|_*
&\le \sum_{s=0}^{k} w_{k,s}\,\big(2\zeta + 2B g_s\big) \\
&= 2\zeta + 2B\Big(q^{k}\, \|\nabla f(x^0)\|_* + S_k\Big),
\end{align*}
where we used $\sum_{s=0}^k w_{k,s}=1$, the weight values $w_{k,0}=q^{k}$ and $w_{k,s}=\alpha q^{k-s}$ for $s \ge 1$, and $g_0 = \|\nabla f(x^0)\|_*$.

\emph{Step 3 (noise term).} By \eqref{eq:noise-pair} of Lemma~\ref{lem:noise}, for every pair $i \neq j$,
\[
\mathbb{E}\bigg\|\sum_{s=0}^{k} w_{k,s}(\xi_i^{s}-\xi_j^{s})\bigg\|_* \le \sqrt{2}\,\kappa\sigma\, q^{k} + V_P.
\]

Averaging the bound of Step 1 over all ordered pairs $i \neq j$ and combining Steps 2 and 3 yields \eqref{eq:dispersion}. The good momenta $\{m_i^k\}_{i\in\mathcal{G}}$ therefore satisfy the premise of Definition~\ref{def:aragg} with dispersion level $r_k$, and the aggregation guarantee gives \eqref{eq:aggerr}. Finally, splitting off the $s=k$ term shows $S_k = \alpha g_k + q\,\alpha\sum_{s=1}^{k-1} q^{\,k-1-s} g_s = qS_{k-1} + \alpha g_k$ for $k \ge 1$.
\end{proof}
\begin{theorem}[Non-convex convergence; formal version of Theorem~\ref{thm:main-nonconvex}]\label{thm:nonconvex-apx}
Suppose we are given a $(\delta, c)$-robust aggregator satisfying Definition~\ref{def:aragg}, and $n$ workers of which the subset $\mathcal{G}$ of size $G \ge \max\{2,\, n(1 - \delta)\}$ faithfully follows the algorithm, with
\[
4 c \delta B < 1.
\]
Further, for any good worker $i \in \mathcal{G}$ let $f_i$ be a possibly non-convex function, and let Assumptions~\ref{asn:stoch-grad}, \ref{asn:lipschitz}, \ref{asn:norm-equivalence}, and~\ref{asn:bounded-heterogeneity} hold. Then the iterates of Algorithm~\ref{alg:byz-muon} satisfy
\begin{align*}
\big(1 - 4 c \delta B\big)\,\frac{1}{K} \sum_{k=0}^{K-1}\mathbb{E}\|\nabla f(x^{k})\|_* \le\ & \frac{f(x^0) - f_{\inf}}{\eta K} + \frac{L\eta}{2} \left(1 + \frac{4(1-\alpha)}{\alpha}\right) \\
 & + 2 V_G + 2 c\delta V_P + 4c \delta \zeta + \frac{C_0}{\alpha K},
\end{align*}
where
\[
C_0 := \frac{2\kappa\sigma}{\sqrt{G}} + 2\sqrt{2}\,c\delta \kappa\sigma + 4 c\delta B \|\nabla f(x^0)\|_*.
\]
Since the average dominates the minimum and equals the expectation over a uniformly sampled iterate, the same bound holds for $\min_{0 \le k < K}\mathbb{E}\|\nabla f(x^k)\|_*$ and for $\mathbb{E}\|\nabla f(\hat{x})\|_*$ with $\hat{x} \sim \mathrm{Unif}\{x^0,\dots,x^{K-1}\}$ drawn independently of the algorithm.
\end{theorem}

\begin{proof}
    Fix $k \in \{0, \dots, K-1\}$. We begin by invoking the descent inequality established in Lemma~\ref{lem:descent_exp_rigorous}:
    \begin{equation*}
        \mathbb{E}[f(x^{k+1})] \le \mathbb{E}[f(x^k)] - \eta\, g_k
        + 2\eta\,\mathbb{E}\|\bar{e}^{k}\|_* + 2\eta\,\mathbb{E}\|m^{k}-\bar{m}^{k}\|_* + \frac{L\eta^2}{2}.
    \end{equation*}
    The trust-region constraint of Algorithm~\ref{alg:byz-muon} gives $\|x^{s}-x^{s-1}\| \le \eta$ for all $s \ge 1$, so Lemma~\ref{lem:error_recursion_final} with $d = \eta$ bounds the ideal momentum error:
    \[
    \mathbb{E}\|\bar{e}^{k}\|_* \le \frac{qL\eta}{\alpha} + \frac{\kappa\sigma}{\sqrt{G}}\, q^{k} + V_G.
    \]
    Lemma~\ref{lem:agg_error_rigorous} bounds the aggregation error:
    \[
    \mathbb{E}\|m^{k}-\bar{m}^{k}\|_* \le c\delta\left[2\zeta + 2BS_k + \Big(\sqrt{2}\kappa\sigma + 2B\|\nabla f(x^0)\|_*\Big) q^{k} + V_P\right].
    \]
    Substituting both bounds into the descent inequality and rearranging,
    \begin{align}
    \eta\, g_k
    &\le \mathbb{E}[f(x^{k})] - \mathbb{E}[f(x^{k+1})] + \frac{L\eta^2}{2} + \frac{2qL\eta^2}{\alpha} + 2\eta V_G + 4\eta c\delta \zeta + 2\eta c\delta V_P \notag\\
    &\quad + \left[\frac{2\eta\kappa\sigma}{\sqrt{G}} + 2\eta c\delta\Big(\sqrt{2}\kappa\sigma + 2B\|\nabla f(x^0)\|_*\Big)\right] q^{k}
    + 4\eta c\delta B\, S_k. \label{eq:onestep-nc}
    \end{align}
    We now sum \eqref{eq:onestep-nc} over $k=0,\dots,K-1$ and treat the resulting terms in turn.

    \emph{(i) Telescoping.} Since $f(x^{K}) \ge f_{\inf}$,
    \[
    \sum_{k=0}^{K-1}\big(\mathbb{E}[f(x^{k})] - \mathbb{E}[f(x^{k+1})]\big) = f(x^{0}) - \mathbb{E}[f(x^{K})] \le f(x^0) - f_{\inf}.
    \]

    \emph{(ii) Geometric transients.} All coefficients multiplying $q^{k}$ in \eqref{eq:onestep-nc} are nonnegative, so, using $\sum_{k=0}^{K-1} q^{k} \le \frac{1}{1-q} = \frac{1}{\alpha}$,
    \begin{align*}
    \sum_{k=0}^{K-1}\left[\frac{2\eta\kappa\sigma}{\sqrt{G}} + 2\eta c\delta\Big(\sqrt{2}\kappa\sigma + 2B\|\nabla f(x^0)\|_*\Big)\right] q^{k}
    &\le \frac{\eta}{\alpha}\left[\frac{2\kappa\sigma}{\sqrt{G}} + 2\sqrt{2}\, c\delta\kappa\sigma + 4c\delta B\|\nabla f(x^0)\|_*\right] \\
    &= \frac{\eta C_0}{\alpha}.
    \end{align*}

    \emph{(iii) History term.} Exchanging the order of summation and using the geometric series bound $\sum_{k \ge s} q^{k-s} \le \frac{1}{\alpha}$,
    \[
    \sum_{k=0}^{K-1} S_k = \alpha \sum_{k=1}^{K-1}\sum_{s=1}^{k} q^{\,k-s}\, g_s = \alpha \sum_{s=1}^{K-1} g_s \sum_{k=s}^{K-1} q^{\,k-s} \le \sum_{s=1}^{K-1} g_s \le \sum_{k=0}^{K-1} g_k,
    \]
    where the last step uses $g_0 \ge 0$. Hence the total contribution of the history term is at most $4\eta c\delta B \sum_{k=0}^{K-1} g_k$.

    Substituting (i)--(iii) into the summed inequality gives
    \[
    \eta \sum_{k=0}^{K-1} g_k \le f(x^0) - f_{\inf} + K\left[\frac{L\eta^2}{2} + \frac{2qL\eta^2}{\alpha} + 2\eta V_G + 4\eta c\delta\zeta + 2\eta c\delta V_P\right] + \frac{\eta C_0}{\alpha} + 4\eta c\delta B \sum_{k=0}^{K-1} g_k.
    \]
    Moving the last term to the left-hand side and dividing by $\eta K$ yields the claimed bound, after noting that
    \[
    \frac{L\eta}{2} + \frac{2qL\eta}{\alpha} = \frac{L\eta}{2}\left(1 + \frac{4q}{\alpha}\right) = \frac{L\eta}{2}\left(1 + \frac{4(1-\alpha)}{\alpha}\right).
    \]
    The final claims about the minimum and the uniformly sampled iterate follow since
    \[
    \min_{0\le k<K} \mathbb{E}\|\nabla f(x^k)\|_*
    \le \frac{1}{K}\sum_{k=0}^{K-1} \mathbb{E}\|\nabla f(x^k)\|_*
    = \mathbb{E}\|\nabla f(\hat{x})\|_* ,
    \]
    where $\hat{x}$ is uniform on $\{x^0,\dots,x^{K-1}\}$ and independent of the algorithm's randomness.
\end{proof}

\begin{remark}[On the heterogeneity floor]\label{rem:heterogeneity-floor}
A fixed positive absolute heterogeneity term $\zeta$ produces the residual $4c\delta\zeta/(1-4c\delta B)$ in our \emph{upper bound}, which cannot be removed by tuning $\eta$ or $\alpha$ alone; the term vanishes if $\zeta=0$. More generally, heterogeneity-dependent error is unavoidable for some Byzantine learning problems: \citet{Karimireddy2020ByzantineRobustLO} prove an expected function-gap lower bound of order $\Omega(\delta\zeta_2^2/\mu)$ for a class of $1$-smooth, $\mu$-strongly convex problems with second-moment Euclidean heterogeneity $\mathbb{E}_{i\sim\mathcal{G}}\|\nabla f_i-\nabla f\|_2^2\le\zeta_2^2$. That statement concerns \emph{expected suboptimality}, not directly our expected \emph{first moment} of the gradient dual norm; a lower bound on a second moment does not alone imply a corresponding lower bound on the first moment for arbitrary randomized outputs. Under the second-moment aggregator assumptions (Theorem~\ref{thm:second-moment}), our \emph{upper} Byzantine heterogeneity term has order $\sqrt{c\delta}\,\kappa\zeta_2$, consistent with the square-root dependence suggested by those lower bounds, but this comparison does not establish optimality for our stationarity criterion. The exact dual-norm geometric median provides a realizable first-moment aggregator, although $c\delta$ cannot generally vanish with positive $\delta$ under Definition~\ref{def:aragg} (Remark~\ref{rem:aggregators}). Momentum damps the stochastic contributions $V_G$ and $V_P$ through $\sqrt{\alpha}$, whereas the additive heterogeneity term carries no such factor.
\end{remark}

\section{Analysis for Star-Convex Functions}\label{apx:starconvex-proof}
In this appendix we prove Theorem~\ref{thm:main-starconvex} for Algorithm~\ref{alg:byz-muon-decay}. Relative to Appendix~\ref{apx:nonconvex-proof}, the only change in the method is the center of the trust region, which is shifted from $x^k$ to $(1-\beta)x^k$; the momentum updates and their initialization are unchanged, so the notation and the momentum lemmas of Appendix~\ref{apx:nonconvex-proof} carry over. Throughout this appendix we impose the standing parameter conditions
\begin{equation}\label{eq:sc-params}
0 < \beta < \alpha \le 1, \qquad \eta \ge \beta\max\{\|x^0\|, \|x^*\|\},
\end{equation}
where $x^*$ is a minimizer of problem~\eqref{eq:main} whose existence is granted by Assumption~\ref{asn:star-cvx}, and we write $\Delta_k := \mathbb{E}[f(x^{k}) - f(x^*)]$.

We start with the geometric properties of the weight-decay trust region; the statement and the proof follow \citet{kovalev2025understanding}.
\begin{lemma}[Weight-decay geometry]\label{lem:aux-inequalities}
    Let \eqref{eq:sc-params} hold, and define the comparison point $y^k := \beta x^* + (1-\beta)x^k$. Then, for all $k \ge 0$, the iterates of Algorithm~\ref{alg:byz-muon-decay} satisfy $\beta\|x^k\| \leq \eta$, and consequently
    \[
    \|y^k - (1 - \beta)x^k\| \leq \eta, \quad
    \|y^k - x^k\| \leq 2\eta, \quad
    \|y^k - x^{k+1}\| \leq 2\eta, \quad
    \|x^{k+1} - x^k\| \leq 2\eta.
    \]
    In particular, $y^k$ is feasible for the trust-region subproblem at iteration $k$, and the iterates remain in the bounded set $\{x : \|x\| \le \max\{\|x^0\|, \eta/\beta\}\}$.
\end{lemma}
\begin{proof}
We first show $\beta\|x^k\|\le\eta$ by induction. The base case $\beta\|x^0\| \le \eta$ holds by \eqref{eq:sc-params}. Assuming $\beta\|x^k\|\le\eta$, the trust-region constraint $\|x^{k+1}-(1-\beta)x^k\|\le\eta$ and the triangle inequality give
\[
\beta\|x^{k+1}\| \le \beta\|x^{k+1}-(1-\beta)x^k\| + \beta(1-\beta)\|x^k\| \le \beta\eta + (1-\beta)\eta = \eta.
\]
The four displayed inequalities now follow. First, $\|y^k - (1-\beta)x^k\| = \beta\|x^*\| \le \eta$ by \eqref{eq:sc-params}; hence $y^k$ is feasible. Second, $\|y^k - x^k\| = \beta\|x^* - x^k\| \le \beta\|x^*\| + \beta\|x^k\| \le 2\eta$. Third, $\|y^k - x^{k+1}\| \le \|y^k - (1-\beta)x^k\| + \|x^{k+1} - (1-\beta)x^k\| \le \eta + \eta = 2\eta$, using feasibility of $x^{k+1}$. Fourth, $\|x^{k+1}-x^k\| \le \|x^{k+1}-(1-\beta)x^k\| + \beta\|x^k\| \le 2\eta$.
\end{proof}
\begin{lemma}[Star-convex descent bound]
\label{lem:scvx-descent}
    Let Assumptions~\ref{asn:lipschitz} and~\ref{asn:star-cvx} and condition~\eqref{eq:sc-params} hold, and let $x^k$ satisfy $\beta\|x^k\| \le \eta$ (as holds for every iterate of Algorithm~\ref{alg:byz-muon-decay} by Lemma~\ref{lem:aux-inequalities}). Let $x^{k+1}$ be obtained from the trust-region linear update
    \[
    x^{k+1} = \arg\min_{\|x-(1-\beta)x^k\|\le \eta} \langle m^{k}, x \rangle,
    \]
    where $m^{k}$ is any vector (e.g., a robustly aggregated momentum). Then
    \begin{align*}
        f(x^{k+1}) \le \beta f(x^*) + (1-\beta)f(x^k) + 4L\eta^2 + 2\eta \|\nabla f(x^k)-m^{k}\|_*.
    \end{align*}
\end{lemma}
\begin{proof}
    Using the $L$-smoothness of $f$ and the bound $\|x^{k+1}-x^k\| \le 2\eta$ from Lemma~\ref{lem:aux-inequalities} (whose derivation of the four distance bounds uses only $\beta\|x^k\| \le \eta$, $\beta\|x^*\| \le \eta$, and feasibility of $x^{k+1}$ --- all available under the present hypotheses), we start with the standard upper bound:
    \[
    f(x^{k+1}) \le f(x^k) + \langle \nabla f(x^k), x^{k+1}-x^k \rangle + 2L\eta^2.
    \]
    To analyze the linear term, we decompose the inner product by adding and subtracting $m^{k}$:
    \[
    \langle \nabla f(x^k), x^{k+1}-x^k \rangle
    = \langle m^{k}, x^{k+1}-x^k \rangle + \langle \nabla f(x^k)-m^{k}, x^{k+1}-x^k \rangle.
    \]
    By Lemma~\ref{lem:aux-inequalities}, the comparison point $y^k = \beta x^* + (1-\beta)x^k$ satisfies $\|y^k - (1 - \beta)x^k\| \leq \eta$, i.e., it is feasible for the subproblem defining $x^{k+1}$, so optimality of $x^{k+1}$ gives $\langle m^{k}, x^{k+1}\rangle \le \langle m^{k}, y^k\rangle$ and hence
    \[\langle m^{k}, x^{k+1}-x^k \rangle\le \langle m^{k}, y^k-x^k \rangle.\]
    Substituting this into the decomposition and \emph{regrouping the two error inner products into a single one},
    \begin{align*}
    \langle \nabla f(x^k), x^{k+1}-x^k \rangle
    &\le \langle m^{k}, y^k-x^k \rangle + \langle \nabla f(x^k)-m^{k}, x^{k+1}-x^k \rangle \\
    &= \langle \nabla f(x^k), y^k-x^k \rangle + \langle m^{k}-\nabla f(x^k), y^k-x^k \rangle \\
    &\qquad + \langle \nabla f(x^k)-m^{k}, x^{k+1}-x^k \rangle \\
    &= \langle \nabla f(x^k), y^k-x^k \rangle + \langle m^{k}-\nabla f(x^k), y^k-x^{k+1} \rangle \\
    &\le \langle \nabla f(x^k), y^k-x^k \rangle + \|\nabla f(x^k)-m^{k}\|_* \cdot \|y^k-x^{k+1}\| \\
    &\le \langle \nabla f(x^k), y^k-x^k \rangle + 2\eta\, \|\nabla f(x^k)-m^{k}\|_*,
    \end{align*}
    where the second equality collects the two inner products against $m^{k}-\nabla f(x^k)$, the first inequality is the definition of the dual norm, and the last step uses $\|y^k - x^{k+1}\| \le 2\eta$ from Lemma~\ref{lem:aux-inequalities}. (Bounding the two error inner products separately would produce the coefficient $4\eta$; the regrouping yields the sharper constant $2\eta$, following the argument of \citet{kovalev2025understanding}.)
    Next, using the smoothness lower bound $f(y^k) \ge f(x^k) + \langle \nabla f(x^k), y^k-x^k \rangle - \frac{L}{2}\|y^k-x^k\|^2$ and $\|y^k - x^k\| \le 2\eta$ from Lemma~\ref{lem:aux-inequalities}, we obtain
    \[f(x^k) + \langle \nabla f(x^k), y^k-x^k \rangle \le f(y^k) + \frac{L}{2}\|y^k-x^k\|^2 \le f(y^k) + 2L\eta^2.\]
    Substituting these bounds back into the smoothness inequality yields
    \begin{align*}
        f(x^{k+1}) \le f(y^k) + 4L\eta^2 + 2\eta \|\nabla f(x^k)-m^{k}\|_*.
    \end{align*}
    Using the star-convexity property (Assumption~\ref{asn:star-cvx}) at $y^k = \beta x^* + (1-\beta)x^k$, we obtain
    \begin{align*}
        f(x^{k+1}) \le \beta f(x^*) + (1-\beta)f(x^k) + 4L\eta^2 + 2\eta \|\nabla f(x^k)-m^{k}\|_*. \qedhere
    \end{align*}
\end{proof}
Since the momentum updates and their initialization are identical to those of Algorithm~\ref{alg:byz-muon}, Lemmas~\ref{lem:momentum-repr}, \ref{lem:noise}, and~\ref{lem:agg_error_rigorous} apply verbatim to Algorithm~\ref{alg:byz-muon-decay}: their proofs involve only the momentum recursion, the stochastic assumptions, and the heterogeneity assumption, and never use the form of the iterate update. The ideal-momentum lemma also carries over, with the step bound enlarged by the weight decay.

\begin{corollary}[Ideal momentum error under weight decay]\label{cor:error-recursion-star}
Let Assumptions~\ref{asn:stoch-grad}--\ref{asn:norm-equivalence} and~\ref{asn:star-cvx}, together with condition~\eqref{eq:sc-params}, hold. Then, for all $k \ge 0$, the iterates of Algorithm~\ref{alg:byz-muon-decay} satisfy
\[
\mathbb{E} \|\bar{e}^k\|_* \le \frac{2qL\eta}{\alpha} + \frac{\kappa\sigma}{\sqrt{G}}\, q^{k} + V_G.
\]
\end{corollary}
\begin{proof}
By Lemma~\ref{lem:aux-inequalities}, the iterates satisfy $\|x^{s}-x^{s-1}\| \le 2\eta$ for all $s \ge 1$, so Lemma~\ref{lem:error_recursion_final} applies with $d = 2\eta$.
\end{proof}

\begin{lemma}[Descent in expectation with aggregation error]
\label{lem:descent_exp_star}
Let Assumptions~\ref{asn:lipschitz} and~\ref{asn:star-cvx} and condition~\eqref{eq:sc-params} hold. For any $k\ge0$, the sequence generated by Algorithm~\ref{alg:byz-muon-decay} satisfies
\begin{equation}
\mathbb{E}[f(x^{k+1})] \le \beta f(x^*) + (1-\beta)\mathbb{E}[f(x^k)] + 2\eta\,\mathbb{E}\|\bar{e}^{k}\|_* + 2\eta\,\mathbb{E}\|\bar{m}^{k} - m^{k}\|_* + 4L \eta^2,
\end{equation}
where $\bar{e}^{k} = \bar{m}^{k} - \nabla f(x^k)$ denotes the ideal momentum error, and the expectation is over all randomness of the algorithm.
\end{lemma}

\begin{proof}
We start by utilizing the descent guarantee provided by Lemma~\ref{lem:scvx-descent}. Applying this lemma to the update from $x^k$ to $x^{k+1}$ gives the following pointwise bound:
\begin{align*}
    f(x^{k+1}) \le \beta f(x^*) + (1-\beta)f(x^k) + 4L\eta^2 + 2\eta \|\nabla f(x^k)-m^{k}\|_*.
\end{align*}
Computing the full expectation with respect to all randomness up to time $k+1$ results in the base inequality
\begin{align*}
    \mathbb{E}[f(x^{k+1})] \le \beta f(x^*) + (1-\beta)\mathbb{E}[f(x^k)] + 4L\eta^2 + 2\eta \mathbb{E}\|\nabla f(x^k)-m^{k}\|_*.
\end{align*}
We now address the error term $\mathbb{E}\|\nabla f(x^{k}) - m^{k}\|_*$. By adding and subtracting the ideal average momentum $\bar{m}^{k}$ and applying the triangle inequality, we can decompose this error as follows:
\begin{equation*}
\|\nabla f(x^{k}) - m^{k}\|_* \le \|\nabla f(x^{k}) - \bar{m}^{k}\|_* + \|\bar{m}^{k} - m^{k}\|_* = \|\bar{e}^k\|_* + \|\bar{m}^{k} - m^{k}\|_*.
\end{equation*}
Inserting this decomposition into the previous expectation inequality yields the final result:
\begin{align*}
\mathbb{E}[f(x^{k+1})] &\le \beta f(x^*) + (1-\beta)\mathbb{E}[f(x^k)] + 2\eta\,\mathbb{E}\left[\|\bar{e}^{k}\|_* + \|\bar{m}^{k} - m^{k}\|_*\right] + 4L\eta^2 \\
&= \beta f(x^*) + (1-\beta)\mathbb{E}[f(x^k)] + 2\eta\,\mathbb{E}\|\bar{e}^{k}\|_* + 2\eta\,\mathbb{E}\|\bar{m}^{k} - m^{k}\|_* + 4L\eta^2.
\end{align*}
This concludes the proof.
\end{proof}
The final ingredient relates the gradient dual norm to the function gap; it holds for any $L$-smooth function with a global minimizer over the (unconstrained) vector space $\mathcal{X}$ and requires no convexity.

\begin{lemma}[Gradient norm and function gap]
\label{lem:grad-gap}
    Let Assumption~\ref{asn:lipschitz} hold and let $x^*$ be a global minimizer of $f$ over $\mathcal{X}$. Then, for every $x \in \mathcal{X}$,
    \[
        \dnorm{\nabla f(x)}^2 \leq 2L\big(f(x) - f(x^*)\big).
    \]
    Consequently, $g_k = \mathbb{E}\dnorm{\nabla f(x^k)} \le \sqrt{2L\Delta_k}$ for all $k \ge 0$.
\end{lemma}

\begin{proof}
    Fix $x \in \mathcal{X}$. By the definition of the dual norm and the compactness of the unit ball of $\|\cdot\|$ in the finite-dimensional space $\mathcal{X}$, there exists $s \in \mathcal{X}$ with
    \[
    \|s\| \le 1 \qquad \text{and} \qquad \langle \nabla f(x), s\rangle = \dnorm{\nabla f(x)}.
    \]
    Consider the steepest-descent point $y := x - \frac{\dnorm{\nabla f(x)}}{L}\, s$. Assumption~\ref{asn:lipschitz} implies the quadratic upper bound $f(y) \le f(x) + \langle \nabla f(x), y - x\rangle + \frac{L}{2}\|y-x\|^2$, whence
    \[
    f(y) \le f(x) - \frac{\dnorm{\nabla f(x)}^2}{L} + \frac{L}{2}\cdot\frac{\dnorm{\nabla f(x)}^2}{L^2}\,\|s\|^2 \le f(x) - \frac{\dnorm{\nabla f(x)}^2}{2L}.
    \]
    Since $x^*$ is a global minimizer, $f(x^*) \le f(y)$, which rearranges to the pointwise claim. The expectation version follows from Jensen's inequality applied to the concave square root:
    \[
    \mathbb{E}\dnorm{\nabla f(x^k)} \le \sqrt{\mathbb{E}\dnorm{\nabla f(x^k)}^2} \le \sqrt{2L\,\mathbb{E}[f(x^k)-f(x^*)]} = \sqrt{2L\Delta_k}. \qedhere
    \]
\end{proof}
\begin{theorem}[Star-convex convergence; formal version of Theorem~\ref{thm:main-starconvex}]\label{thm:starconvex-apx}
    Suppose we are given a $(\delta,c)$-robust aggregator satisfying Definition~\ref{def:aragg}, and $n$ workers of which the subset $\mathcal{G}$ of size $G \ge \max\{2,\, n(1-\delta)\}$ faithfully follows the algorithm. Let Assumptions~\ref{asn:stoch-grad}, \ref{asn:lipschitz}, \ref{asn:norm-equivalence}, \ref{asn:star-cvx}, and~\ref{asn:bounded-heterogeneity} hold, and let the parameters satisfy condition~\eqref{eq:sc-params}. Define
    \[
    K_B := \frac{4c\delta\eta\,\alpha B(1-\beta)}{\alpha-\beta}, \qquad
    \widetilde{D} := 2\eta\kappa\sigma\left(\frac{1}{\sqrt{G}} + \sqrt{2}\,c\delta\right) + 4c\delta\eta B\,\|\nabla f(x^0)\|_*,
    \]
    \[
    C_\star := \frac{4L\eta^2}{\alpha} + 2\eta V_G + 4c\delta\eta\zeta + 2c\delta\eta V_P.
    \]
    Then, for every $K \ge 0$, the iterates of Algorithm~\ref{alg:byz-muon-decay} satisfy
    \[
    \Delta_K \;\le\; \left(1-\frac{\beta}{2}\right)^{K}\left(\Delta_0 + \frac{\widetilde{D}}{\alpha - \beta/2}\right) + \frac{2}{\beta}\left(C_\star + \frac{L K_B^2}{\beta}\right).
    \]
\end{theorem}
\begin{proof}
\emph{Step 1 (one-step recursion).} Fix $k \ge 0$. Substituting into Lemma~\ref{lem:descent_exp_star} the momentum-error bound of Corollary~\ref{cor:error-recursion-star} and the aggregation-error bound~\eqref{eq:aggerr} of Lemma~\ref{lem:agg_error_rigorous}, and subtracting $f(x^*)$ from both sides, we obtain
\begin{align*}
    \Delta_{k+1} &\le (1-\beta)\Delta_k + 4L\eta^2 + 2\eta\left[\frac{2qL\eta}{\alpha} + \frac{\kappa\sigma}{\sqrt{G}}q^{k} + V_G\right] \\
    &\quad + 2\eta c\delta\left[2\zeta + 2BS_k + \Big(\sqrt{2}\kappa\sigma + 2B\|\nabla f(x^0)\|_*\Big)q^{k} + V_P\right] \\
    &= (1-\beta)\Delta_k + C_\star + \widetilde{D}\, q^{k} + 4c\delta\eta B\, S_k,
\end{align*}
where we collected the constant terms using
\[
4L\eta^2 + \frac{4qL\eta^2}{\alpha} = \frac{4L\eta^2(\alpha + q)}{\alpha} = \frac{4L\eta^2}{\alpha},
\]
so that indeed $C_\star = \frac{4L\eta^2}{\alpha} + 2\eta V_G + 4c\delta\eta\zeta + 2c\delta\eta V_P$, and the coefficients of $q^k$ sum to $\widetilde{D}$.

\emph{Step 2 (Lyapunov function).} Since $\alpha > \beta$ by \eqref{eq:sc-params}, we may define
\[
\Gamma := \frac{4c\delta\eta B\, q}{\alpha - \beta} \;\ge\; 0,
\qquad
\Psi_0 := \Delta_0, \qquad \Psi_{k} := \Delta_{k} + \Gamma S_{k-1} \ \ (k \ge 1),
\]
where $S_k$ is the relative-heterogeneity history of Lemma~\ref{lem:agg_error_rigorous} (recall $S_0 = 0$, so $\Psi_1 = \Delta_1$). A direct computation shows that
\[
4c\delta\eta B + \Gamma = 4c\delta\eta B\left(1 + \frac{q}{\alpha-\beta}\right) = \frac{4c\delta\eta B(1-\beta)}{\alpha-\beta} =: \Lambda,
\]
and this constant satisfies the two identities
\begin{equation}\label{eq:lyap-identities}
q\,\Lambda = (1-\beta)\,\Gamma, \qquad \alpha\,\Lambda = K_B.
\end{equation}

\emph{Step 3 (contraction of the potential).} For $k \ge 1$, adding $\Gamma S_k$ to both sides of the one-step recursion and substituting the recursion $S_k = q S_{k-1} + \alpha g_k$ from Lemma~\ref{lem:agg_error_rigorous},
\begin{align*}
\Psi_{k+1} = \Delta_{k+1} + \Gamma S_k
&\le (1-\beta)\Delta_k + \big(4c\delta\eta B + \Gamma\big) S_k + C_\star + \widetilde{D} q^{k} \\
&= (1-\beta)\Delta_k + \Lambda\big(q S_{k-1} + \alpha g_k\big) + C_\star + \widetilde{D} q^{k} \\
&\overset{\eqref{eq:lyap-identities}}{=} (1-\beta)\big(\Delta_k + \Gamma S_{k-1}\big) + K_B\, g_k + C_\star + \widetilde{D} q^{k} \\
&= (1-\beta)\Psi_k + K_B\, g_k + C_\star + \widetilde{D} q^{k}.
\end{align*}
By Lemma~\ref{lem:grad-gap} and Young's inequality $xy \le \frac{\beta}{2}x^2 + \frac{y^2}{2\beta}$ applied with $x = \sqrt{\Delta_k}$ and $y = K_B\sqrt{2L}$,
\[
K_B\, g_k \le K_B\sqrt{2L\Delta_k} \le \frac{\beta}{2}\Delta_k + \frac{L K_B^2}{\beta} \le \frac{\beta}{2}\Psi_k + \frac{L K_B^2}{\beta},
\]
where the last step uses $\Delta_k \le \Psi_k$ ($\Gamma, S_{k-1} \ge 0$). Therefore, for all $k \ge 1$,
\begin{equation}\label{eq:potential-step}
\Psi_{k+1} \le \left(1 - \frac{\beta}{2}\right)\Psi_k + C_\star + \frac{L K_B^2}{\beta} + \widetilde{D}\, q^{k}.
\end{equation}
For $k = 0$, the one-step recursion with $S_0 = 0$ gives directly $\Psi_1 = \Delta_1 \le (1-\beta)\Delta_0 + C_\star + \widetilde{D} = (1-\beta)\Psi_0 + C_\star + \widetilde{D}\,q^0$, so \eqref{eq:potential-step} also holds at $k=0$ (the discarded terms are nonnegative). Hence \eqref{eq:potential-step} holds for every $k \ge 0$.

\emph{Step 4 (unrolling).} Write $\theta := 1-\frac{\beta}{2}$ and note $\theta - q = (1-\frac{\beta}{2}) - (1-\alpha) = \alpha - \frac{\beta}{2} > 0$ by \eqref{eq:sc-params}. Unrolling \eqref{eq:potential-step} from $k = 0$ to $K-1$,
\[
\Psi_K \le \theta^{K}\,\Psi_0 + \left(C_\star + \frac{L K_B^2}{\beta}\right)\sum_{j=0}^{K-1} \theta^{j} + \widetilde{D}\sum_{k=0}^{K-1} \theta^{\,K-1-k}\, q^{k}.
\]
The first geometric sum is bounded by $\sum_{j\ge0} \theta^j = \frac{1}{1-\theta} = \frac{2}{\beta}$. The cross sum evaluates to
\[
\sum_{k=0}^{K-1} \theta^{\,K-1-k} q^{k} = \frac{\theta^{K} - q^{K}}{\theta - q} \le \frac{\theta^{K}}{\alpha - \beta/2}.
\]
Since $\Psi_0 = \Delta_0$ and $\Delta_K \le \Psi_K$, we conclude
\[
\Delta_K \le \left(1-\frac{\beta}{2}\right)^{K}\left(\Delta_0 + \frac{\widetilde{D}}{\alpha-\beta/2}\right) + \frac{2}{\beta}\left(C_\star + \frac{L K_B^2}{\beta}\right). \qedhere
\]
\end{proof}

\begin{remark}[No restriction on $\delta B$; role of the condition $\alpha > \beta$]
In contrast to the non-convex case, no upper bound on $c\delta B$ is required: the relative-heterogeneity history $S_k$ is fully absorbed by the weight-decay contraction via Young's inequality, and enters the final bound only through the quadratic neighborhood term $2LK_B^2/\beta^2$. The condition $\alpha > \beta$ is needed only to form the potential ($\Gamma \ge 0$); when $B = 0$ or $\delta = 0$, one has $\Gamma = K_B = 0$, the potential reduces to $\Psi_k = \Delta_k$, and the requirement $\alpha > \beta$ can be relaxed to $\alpha > \beta/2$ (needed for the transient cross sum). At $\delta = 0$, taking the aggregator to be the plain average (so that $m^{k} = \bar{m}^{k}$ exactly and the aggregation-error term is dropped from the start, without invoking Definition~\ref{def:aragg} or Lemma~\ref{lem:agg_error_rigorous}), the resulting bound --- then valid even for a single worker, $G = 1$ --- recovers the structure of Theorem~4 of \citet{kovalev2025understanding} up to absolute constants.
\end{remark}

\section{Guarantees under Second-Moment Euclidean Aggregators}\label{apx:second-moment}

The results of Appendices~\ref{apx:nonconvex-proof} and~\ref{apx:starconvex-proof} are stated under the first-moment aggregation property of Definition~\ref{def:aragg}, which Remark~\ref{rem:aggregators} shows to be an idealized assumption: its factor satisfies $c\delta \ge \nicefrac{1}{2}$ in the worst case over inputs. In this appendix we prove the guarantees that hold under the \emph{worst-case-realizable} route: the second-moment Euclidean aggregation property satisfied by bucketed Krum, geometric median, and coordinate-wise median, under their respective conditions \citep{Karimireddy2020ByzantineRobustLO, gorbunov2022variance}, converted to the dual norm by Jensen's inequality and Assumption~\ref{asn:norm-equivalence}. The price of realizability is the aggregation factor $\kappa\sqrt{c\delta}$ in place of $c\delta$, together with a second-moment heterogeneity assumption; the structure of the bounds is otherwise unchanged.

\begin{assumption}[Second-moment robust aggregator]\label{asn:aragg2}
Let $\delta < \nicefrac{1}{2}$ and $c \ge 0$. For any random inputs $\{z_1,\dots,z_n\} \subset \mathcal{X}$ with a good subset $\mathcal{G} \subseteq [n]$ of size $G = |\mathcal{G}| \ge \max\{2,\, (1-\delta)n\}$ satisfying
$\frac{1}{G(G-1)}\sum_{\substack{i,l\in\mathcal{G}\\ i\neq l}} \mathbb{E}\|z_i - z_l\|_2^2 \le \bar{\rho}^2$
for some $\bar{\rho} \ge 0$, the aggregator output satisfies
\[
\mathbb{E}\|\hat{z} - \bar{z}\|_2^2 \;\le\; c\delta\, \bar{\rho}^2, \qquad \bar{z} = \frac{1}{G}\sum_{i\in\mathcal{G}} z_i,
\]
and is computed without knowledge of $\bar{\rho}^2$.
\end{assumption}
This agnostic second-moment property is available for suitable bucketed Krum, geometric-median, and coordinate-wise-median constructions under their respective restrictions \citep{Karimireddy2020ByzantineRobustLO, gorbunov2022variance}. The centered-clipping result cited there depends on a dispersion-calibrated radius and is not an agnostic instance of Assumption~\ref{asn:aragg2}.

\begin{assumption}[Second-moment additive heterogeneity]\label{asn:het2}
There exists $\zeta_2 \ge 0$ such that, for all $x \in \mathcal{X}$,
\[
\frac{1}{G} \sum_{i \in \mathcal{G}} \|\nabla f_i(x) - \nabla f(x)\|_2^2 \;\le\; \zeta_2^2.
\]
\end{assumption}
This is the additive ($B_2 = 0$) case of the second-moment heterogeneity condition of \citet{Karimireddy2020ByzantineRobustLO}; the relative component is discussed in Remark~\ref{rem:relative2} below.

\begin{lemma}[Second-moment momentum dispersion]\label{lem:dispersion2}
Let Assumptions~\ref{asn:stoch-grad} and~\ref{asn:het2} hold. Then, for every $k \ge 0$, the good-worker momenta of Algorithm~\ref{alg:byz-muon} (equivalently, of Algorithm~\ref{alg:byz-muon-decay}) satisfy
\[
\frac{1}{G(G-1)}\sum_{\substack{i,j\in\mathcal{G}\\ i\neq j}} \mathbb{E}\|m_i^{k} - m_j^{k}\|_2^2
\;\le\; \bar{\rho}_k^2 := 8\zeta_2^2 + 4\sigma^2 q^{2k} + \frac{4\alpha\sigma^2}{2-\alpha}.
\]
\end{lemma}
\begin{proof}
Fix $i \neq j \in \mathcal{G}$. By Lemma~\ref{lem:momentum-repr},
\[
m_i^{k} - m_j^{k} = \underbrace{\sum_{s=0}^{k} w_{k,s}\,\big(\nabla f_i(x^{s}) - \nabla f_j(x^{s})\big)}_{=:\,D_{ij}} \;+\; \underbrace{\sum_{s=0}^{k} w_{k,s}\,\big(\xi_i^{s} - \xi_j^{s}\big)}_{=:\,N_{ij}},
\]
and $\mathbb{E}\|m_i^k - m_j^k\|_2^2 \le 2\,\mathbb{E}\|D_{ij}\|_2^2 + 2\,\mathbb{E}\|N_{ij}\|_2^2$.

\emph{Deterministic part.} The weights $w_{k,s}$ form a convex combination (Lemma~\ref{lem:momentum-repr}) and $\|\cdot\|_2^2$ is convex, so Jensen's inequality gives $\|D_{ij}\|_2^2 \le \sum_{s=0}^{k} w_{k,s}\, \|\nabla f_i(x^{s})-\nabla f_j(x^{s})\|_2^2$. For every fixed $x \in \mathcal{X}$, inserting $\nabla f(x)$ and using $\|a-b\|_2^2 \le 2\|a\|_2^2 + 2\|b\|_2^2$,
\[
\frac{1}{G(G-1)}\sum_{\substack{i,j\in\mathcal{G}\\ i\neq j}} \|\nabla f_i(x)-\nabla f_j(x)\|_2^2
\le \frac{4}{G}\sum_{i\in\mathcal{G}}\|\nabla f_i(x)-\nabla f(x)\|_2^2 \le 4\zeta_2^2,
\]
where the first step uses $\sum_{i \neq j}\big(2\|\nabla f_i(x) - \nabla f(x)\|_2^2 + 2\|\nabla f_j(x) - \nabla f(x)\|_2^2\big) = 4(G-1)\sum_{i}\|\nabla f_i(x)-\nabla f(x)\|_2^2$, and the second is Assumption~\ref{asn:het2}. Applying this pointwise at $x = x^{s}$, taking expectations, and averaging over ordered pairs,
\[
\frac{1}{G(G-1)}\sum_{i\neq j} \mathbb{E}\|D_{ij}\|_2^2 \le \sum_{s=0}^{k} w_{k,s} \cdot 4\zeta_2^2 = 4\zeta_2^2.
\]

\emph{Stochastic part.} Exactly as in Steps 1--2 of the proof of Lemma~\ref{lem:noise}, conditional independence across good workers and the martingale property across iterations give
\[
\mathbb{E}\|N_{ij}\|_2^2 = \sum_{s=0}^{k} w_{k,s}^2\, \mathbb{E}\|\xi_i^{s} - \xi_j^{s}\|_2^2 \le 2\sigma^2 \sum_{s=0}^{k} w_{k,s}^2 \le 2\sigma^2\Big(q^{2k} + \frac{\alpha}{2-\alpha}\Big),
\]
using \eqref{eq:weight-l2}. Combining the two parts, $2\cdot 4\zeta_2^2 + 2\cdot 2\sigma^2\big(q^{2k} + \frac{\alpha}{2-\alpha}\big) = \bar{\rho}_k^2$.
\end{proof}

\begin{corollary}[Aggregation error in the dual norm]\label{cor:aggerr2}
Let additionally Assumptions~\ref{asn:norm-equivalence} and~\ref{asn:aragg2} hold, with $G \ge \max\{2,\, (1-\delta)n\}$. Then, for every $k \ge 0$,
\[
\mathbb{E}\|m^{k} - \bar{m}^{k}\|_* \;\le\; \kappa\sqrt{c\delta}\;\bar{\rho}_k \;\le\; \sqrt{c\delta}\left(2\sqrt{2}\,\kappa\zeta_2 + \sqrt{2}\, V_P\right) + 2\kappa\sigma\sqrt{c\delta}\; q^{k}.
\]
\end{corollary}
\begin{proof}
By Lemma~\ref{lem:dispersion2}, the good momenta satisfy the premise of Assumption~\ref{asn:aragg2} with dispersion level $\bar{\rho}_k$, so $\mathbb{E}\|m^k - \bar{m}^k\|_2^2 \le c\delta\,\bar{\rho}_k^2$. By Assumption~\ref{asn:norm-equivalence} and Jensen's inequality, $\mathbb{E}\|m^k - \bar{m}^k\|_* \le \kappa\,\mathbb{E}\|m^k - \bar{m}^k\|_2 \le \kappa\sqrt{\mathbb{E}\|m^k - \bar{m}^k\|_2^2} \le \kappa\sqrt{c\delta}\,\bar{\rho}_k$. Finally, $\sqrt{a+b+c} \le \sqrt{a}+\sqrt{b}+\sqrt{c}$ gives $\bar{\rho}_k \le 2\sqrt{2}\,\zeta_2 + 2\sigma q^{k} + 2\sigma\sqrt{\alpha/(2-\alpha)}$, and $2\kappa\sigma\sqrt{\alpha/(2-\alpha)} = \sqrt{2}\,V_P$.
\end{proof}

\begin{theorem}[Non-convex convergence with second-moment aggregators]\label{thm:second-moment}
Suppose the server aggregate satisfies Assumption~\ref{asn:aragg2}, and $n$ workers of which the subset $\mathcal{G}$ of size $G \ge \max\{2,\, n(1-\delta)\}$ faithfully follows the algorithm. Let Assumptions~\ref{asn:stoch-grad}, \ref{asn:lipschitz}, \ref{asn:norm-equivalence}, and~\ref{asn:het2} hold, and let $f_{\inf} := \inf_x f(x) > -\infty$. Then the iterates of Algorithm~\ref{alg:byz-muon} satisfy
\begin{align*}
\frac{1}{K} \sum_{k=0}^{K-1}\mathbb{E}\|\nabla f(x^{k})\|_* \;\le\;\ & \frac{f(x^0) - f_{\inf}}{\eta K} + \frac{L\eta}{2}\left(1 + \frac{4(1-\alpha)}{\alpha}\right) + 2 V_G \\
& + \sqrt{c\delta}\left(4\sqrt{2}\,\kappa\zeta_2 + 2\sqrt{2}\, V_P\right) + \frac{\bar{C}_0}{\alpha K},
\end{align*}
where $\bar{C}_0 := \frac{2\kappa\sigma}{\sqrt{G}} + 4\sqrt{c\delta}\,\kappa\sigma$. The same bound holds for $\min_{0\le k < K}\mathbb{E}\|\nabla f(x^k)\|_*$ and for a uniformly sampled iterate. No upper-bound condition on $\delta$ beyond $\delta < \nicefrac{1}{2}$ is required.
\end{theorem}
\begin{proof}
Fix $k \in \{0,\dots,K-1\}$. Lemma~\ref{lem:descent_exp_rigorous} applies (its finiteness parenthetical is supplied here by Corollary~\ref{cor:aggerr2}), and we substitute the momentum-error bound of Lemma~\ref{lem:error_recursion_final} (with $d = \eta$) and the aggregation bound of Corollary~\ref{cor:aggerr2}:
\begin{align*}
\eta\, g_k \le\ & \mathbb{E}[f(x^{k})] - \mathbb{E}[f(x^{k+1})] + \frac{L\eta^2}{2}
+ 2\eta\left[\frac{qL\eta}{\alpha} + \frac{\kappa\sigma}{\sqrt{G}}\, q^{k} + V_G\right] \\
& + 2\eta\left[\sqrt{c\delta}\big(2\sqrt{2}\kappa\zeta_2 + \sqrt{2} V_P\big) + 2\kappa\sigma\sqrt{c\delta}\, q^{k}\right].
\end{align*}
Summing over $k = 0,\dots,K-1$ and repeating steps (i)--(ii) of the proof of Theorem~\ref{thm:nonconvex-apx} --- the telescoping gives $f(x^0)-f_{\inf}$, and the nonnegative $q^{k}$-coefficients sum against $\sum_k q^{k} \le \frac{1}{\alpha}$ to $\frac{\eta}{\alpha}\big(\frac{2\kappa\sigma}{\sqrt{G}} + 4\sqrt{c\delta}\kappa\sigma\big) = \frac{\eta \bar{C}_0}{\alpha}$ --- there is no gradient-history term to absorb (Assumption~\ref{asn:het2} has no relative component), so dividing by $\eta K$ and using $\frac{L\eta}{2} + \frac{2qL\eta}{\alpha} = \frac{L\eta}{2}\big(1+\frac{4(1-\alpha)}{\alpha}\big)$ yields the claim.
\end{proof}

\begin{corollary}[Star-convex convergence with second-moment aggregators]\label{cor:starconvex2}
Let, in addition, Assumption~\ref{asn:star-cvx} hold, and let the parameters satisfy $0 < \beta < 1$, $\alpha > \beta/2$, and $\eta \ge \beta\max\{\|x^0\|, \|x^*\|\}$. Then the iterates of Algorithm~\ref{alg:byz-muon-decay} satisfy, for every $K \ge 0$,
\[
\Delta_K \;\le\; \Big(1-\frac{\beta}{2}\Big)^{K}\left(\Delta_0 + \frac{\bar{D}}{\alpha - \beta/2}\right) + \frac{2}{\beta}\,\bar{C}_4,
\]
where $\bar{D} := \eta\bar{C}_0 = 2\eta\kappa\sigma\big(\tfrac{1}{\sqrt{G}} + 2\sqrt{c\delta}\big)$ and $\bar{C}_4 := \frac{4L\eta^2}{\alpha} + 2\eta V_G + \sqrt{c\delta}\,\eta\big(4\sqrt{2}\,\kappa\zeta_2 + 2\sqrt{2}\, V_P\big)$.
\end{corollary}
\begin{proof}
Lemma~\ref{lem:dispersion2} and Corollary~\ref{cor:aggerr2} hold verbatim for Algorithm~\ref{alg:byz-muon-decay}, since they use only the momentum recursion. Although Lemma~\ref{lem:descent_exp_star} and Corollary~\ref{cor:error-recursion-star} are stated under condition~\eqref{eq:sc-params}, their proofs (and those of Lemmas~\ref{lem:aux-inequalities} and~\ref{lem:scvx-descent} that they invoke) use only the $\beta$- and $\eta$-parts of that condition, namely $0 < \beta < 1$ and $\eta \ge \beta\max\{\|x^0\|, \|x^*\|\}$ --- the requirement $\alpha > \beta$ enters only the Lyapunov construction of Theorem~\ref{thm:starconvex-apx}, which is not needed here --- so they apply under the present hypotheses. Substituting Corollary~\ref{cor:error-recursion-star} and Corollary~\ref{cor:aggerr2} into Lemma~\ref{lem:descent_exp_star}, and using $4L\eta^2 + \frac{4qL\eta^2}{\alpha} = \frac{4L\eta^2}{\alpha}$ as in Theorem~\ref{thm:starconvex-apx},
\[
\Delta_{k+1} \le (1-\beta)\,\Delta_k + \bar{C}_4 + \bar{D}\, q^{k} \le \Big(1-\frac{\beta}{2}\Big)\Delta_k + \bar{C}_4 + \bar{D}\, q^{k}.
\]
No Lyapunov correction is needed: Assumption~\ref{asn:het2} has no relative component, so no gradient-history term arises (formally, $\Gamma = K_B = 0$, and the condition $\alpha > \beta$ of Theorem~\ref{thm:starconvex-apx} relaxes to $\alpha > \beta/2$). Unrolling exactly as in Step 4 of the proof of Theorem~\ref{thm:starconvex-apx}, with $\theta = 1 - \beta/2$ and cross sum $\sum_{k=0}^{K-1}\theta^{K-1-k}q^{k} = \frac{\theta^K - q^K}{\alpha - \beta/2} \le \frac{\theta^K}{\alpha-\beta/2}$, gives the claim.
\end{proof}

\begin{remark}[The relative heterogeneity component]\label{rem:relative2}
Under the full second-moment condition of \citet{Karimireddy2020ByzantineRobustLO}, $\frac{1}{G}\sum_{i}\|\nabla f_i(x) - \nabla f(x)\|_2^2 \le \zeta_2^2 + B_2^2\|\nabla f(x)\|_2^2$ with $B_2 > 0$, our results extend as follows. Suppose the gradients are bounded by $M$ on the algorithm's reachable set: $\{x : \|x - x^0\| \le K\eta\}$ for Algorithm~\ref{alg:byz-muon} (whose iterates lie there almost surely, since each step has length at most $\eta$), or $\{x : \|x\| \le \max\{\|x^0\|, \eta/\beta\}\}$ for Algorithm~\ref{alg:byz-muon-decay} (Lemma~\ref{lem:aux-inequalities}). Since Lemma~\ref{lem:dispersion2} applies the heterogeneity condition only at the iterates $x^{s}$, the condition then acts as Assumption~\ref{asn:het2} with $\zeta_2$ replaced by $\sqrt{\zeta_2^2 + B_2^2 M^2} \le \zeta_2 + B_2 M$, and Theorem~\ref{thm:second-moment} and Corollary~\ref{cor:starconvex2} apply verbatim under that substitution. Absorbing $B_2$ into a multiplicative prefactor, as the factor $(1 - 4c\delta B)$ does for the first-moment analysis of Theorem~\ref{thm:nonconvex-apx}, appears to require a descent inequality in \emph{squared} gradient norms, as in \citet{Karimireddy2020ByzantineRobustLO}: the trust-region step yields first-moment descent $-\eta\,\mathbb{E}\|\nabla f(x^k)\|_*$, and a second-moment gradient history cannot, in general, be controlled by a first-moment average (Jensen's inequality runs in the opposite direction). Obtaining the sharp relative-heterogeneity prefactor under first-moment trust-region descent is an interesting open problem.
\end{remark}

\section{Improved Analysis under Second-Order Smoothness}\label{apx:second-order}

In this appendix we show that, under an additional second-order smoothness assumption, the first-order drift term $\frac{2(1-\alpha)L\eta}{\alpha}$ in Theorem~\ref{thm:main-nonconvex} is replaced by $\frac{(1-\alpha)H\eta^2}{\alpha^2}$ plus a Byzantine cross term of order $c\delta B L\eta/\alpha$ by evaluating the worker gradients at server-broadcast \emph{extrapolated} points, following the implicit-gradient-transport technique of \citet{cutkosky2020momentum} in the non-Euclidean trust-region form of \citet{kovalev2025understanding} (their Algorithm~3 and Theorem~5). The extrapolation is a purely algebraic modification of the query points and composes with robust aggregation with an additional term in the dispersion analysis (the conversion term $\frac{2BqL\eta}{\alpha}$ in Lemma~\ref{lem:dispersion-extra} below, which becomes the Byzantine cross term $\frac{4c\delta B qL\eta}{\alpha}$ of Theorem~\ref{thm:second-order} and vanishes when $\delta = 0$ or $B = 0$); in particular, the Byzantine error floor is unchanged.

\begin{assumption}[Second-order smoothness]\label{asn:hessian}
The function $f$ is twice differentiable and its Hessian is $H$-Lipschitz with respect to the norm $\|\cdot\|$, in the sense that
\[
\big\|\big(\nabla^2 f(x) - \nabla^2 f(x')\big)(x - x')\big\|_* \;\le\; H\,\|x - x'\|^2 \qquad \text{for all } x, x' \in \mathcal{X},
\]
where $H > 0$ is the Hessian Lipschitz constant.
\end{assumption}
This is Assumption~(A5) of \citet{kovalev2025understanding}. It suffices to impose the per-worker analogue for each $f_i$, $i \in \mathcal{G}$: since $\nabla^2 f = \frac{1}{G}\sum_{i\in\mathcal{G}} \nabla^2 f_i$, the average $f$ inherits the property with the same constant by the triangle inequality.

\begin{lemma}[Second-order Taylor bound]\label{lem:taylor}
Under Assumption~\ref{asn:hessian}, for all $x, y \in \mathcal{X}$,
\[
\big\|\nabla f(y) - \nabla f(x) - \nabla^2 f(x)(y - x)\big\|_* \;\le\; \frac{H}{2}\,\|y - x\|^2.
\]
\end{lemma}
\begin{proof}
Write $\Delta := y - x$. By the fundamental theorem of calculus applied to $t \mapsto \nabla f(x + t\Delta)$,
\[
\nabla f(y) - \nabla f(x) - \nabla^2 f(x)\Delta = \int_0^1 \big(\nabla^2 f(x + t\Delta) - \nabla^2 f(x)\big)\Delta \, \mathrm{d}t.
\]
For each $t \in (0, 1]$, applying Assumption~\ref{asn:hessian} with the pair $(x + t\Delta,\, x)$, whose difference is $t\Delta$, gives $\|(\nabla^2 f(x + t\Delta) - \nabla^2 f(x))(t\Delta)\|_* \le H t^2 \|\Delta\|^2$, i.e., $\|(\nabla^2 f(x + t\Delta) - \nabla^2 f(x))\Delta\|_* \le H t \|\Delta\|^2$ after dividing by $t$ (the bound also holds trivially at $t = 0$). The same argument applied to pairs $(x + t\Delta, x + t'\Delta)$ shows that the integrand is Lipschitz in $t$, hence continuous, so the integral representation and the triangle inequality for the integral are valid. Integrating, $\int_0^1 Ht\|\Delta\|^2\,\mathrm{d}t = \frac{H}{2}\|\Delta\|^2$, which proves the claim.
\end{proof}

\paragraph{Algorithm.}
The extrapolated variant is stated as Algorithm~\ref{alg:byz-muon-extra}. Relative to Algorithm~\ref{alg:byz-muon}, the only change is that the server additionally broadcasts the extrapolated point
\begin{equation}\label{eq:extrapolation}
\bar{x}^{k+1} := x^{k} + \frac{1}{\alpha}\big(x^{k+1} - x^{k}\big), \qquad \bar{x}^0 := x^0,
\end{equation}
and the good workers evaluate their stochastic gradients at $\bar{x}^{k+1}$ instead of $x^{k+1}$; this is the fixed extrapolation parameter $1/\alpha$ (denoted $\gamma$ by \citet{kovalev2025understanding}). Since $\bar{x}^{k+1}$ is a deterministic function of $x^k$ and $x^{k+1}$, it is $\mathcal{F}^{k}$-measurable. Throughout this appendix, the noises are defined at the query points, $\xi_i^{k} := g_i(\bar{x}^{k}; \xi_i^{k}) - \nabla f_i(\bar{x}^{k})$, and Assumption~\ref{asn:stoch-grad} is imposed at the query points $\bar{x}^{k}$, i.e., $\mathbb{E}[\xi_i^{k} \mid \mathcal{F}^{k-1}] = 0$ and $\mathbb{E}[\|\xi_i^{k}\|_2^2 \mid \mathcal{F}^{k-1}] \le \sigma^2$, with conditional independence across good workers. We take $\beta = 0$ (no weight decay), matching Theorem~5 of \citet{kovalev2025understanding}.

\begin{algorithm}[H]
\caption{Byzantine-Robust Muon with Extrapolation}
\label{alg:byz-muon-extra}
\begin{algorithmic}[1]
    \STATE {\bfseries Input:} $(\delta,c)$-ARAgg, trust-region radius $\eta > 0$, worker momentum weight $\alpha \in (0,1]$, initial point $x^0 \in \mathcal{X}$; set $\bar{x}^0 \leftarrow x^0$
    \STATE Each good worker $i \in \mathcal{G}$ initializes $m_i^{0} \leftarrow g_i(\bar{x}^0;\,\xi_i^{0})$; each Byzantine worker $i \in \mathcal{B}$ sends an arbitrary vector $m_i^0$
    \STATE Server computes $m^{0} \leftarrow \mathrm{ARAGG}(m_1^{0},\dots,m_n^{0})$
    \FOR{$k = 0,1,\dots, K-1$}
        \STATE Server solves trust-region linear subproblem:
        \STATE \hskip1em $x^{k+1} \;=\; \underset{x \in \mathcal{X}}{\operatorname{argmin}}\ \langle m^{k}, x \rangle \quad\text{s.t.}\quad \|x - x^k\| \le \eta$
        \STATE Server broadcasts the extrapolated point $\bar{x}^{k+1} \leftarrow x^{k} + \frac{1}{\alpha}(x^{k+1} - x^{k})$
        \FOR{worker $i \in [n]$ {\bfseries in parallel}}
            \STATE Good worker $i \in \mathcal{G}$: compute $g_i^{k+1} \leftarrow g_i(\bar{x}^{k+1};\,\xi_i^{k+1})$, update $m_i^{k+1} \leftarrow (1-\alpha)\,m_i^{k} + \alpha\,g_i^{k+1}$, send $m_i^{k+1}$
            \STATE Byzantine worker $i \in \mathcal{B}$: send an arbitrary vector $m_i^{k+1} \in \mathcal{X}$
        \ENDFOR
        \STATE Server computes robust aggregate momentum: $m^{k+1} \leftarrow \mathrm{ARAGG}(m_1^{k+1},\dots,m_n^{k+1})$
    \ENDFOR
    \STATE {\bfseries Output:} final model $x^K$
\end{algorithmic}
\end{algorithm}

\paragraph{Carrying over the momentum lemmas.}
The momentum recursion and its initialization are unchanged, so the induction in the proof of Lemma~\ref{lem:momentum-repr} applies verbatim with the query points $\bar{x}^s$ in place of $x^s$: for every $i \in \mathcal{G}$ and $k \ge 0$,
\begin{equation}\label{eq:repr-extra}
m_i^{k} = \sum_{s=0}^{k} w_{k,s}\left(\nabla f_i(\bar{x}^{s}) + \xi_i^{s}\right),
\end{equation}
with the same convex weights $w_{k,s}$ and the same bound~\eqref{eq:weight-l2}. Likewise, Lemma~\ref{lem:noise} applies verbatim, since its proof uses only the conditional moment properties of the noises (now imposed at the query points) and the weight bound. Lemmas~\ref{lem:tr_opt}, \ref{lem:descent_tr_rigorous}, and~\ref{lem:descent_exp_rigorous} also apply verbatim, since the $x$-update of Algorithm~\ref{alg:byz-muon-extra} coincides with that of Algorithm~\ref{alg:byz-muon} (for Algorithm~\ref{alg:byz-muon-extra}, the finiteness of the aggregation-error expectation in Lemma~\ref{lem:descent_exp_rigorous} is supplied by Lemma~\ref{lem:dispersion-extra} below in place of Lemma~\ref{lem:agg_error_rigorous}); in particular, $\|x^{s} - x^{s-1}\| \le \eta$ for all $s \ge 1$ and, by~\eqref{eq:extrapolation},
\begin{equation}\label{eq:extra-displacement}
\|\bar{x}^{s} - x^{s}\| = \Big\|\Big(\tfrac{1}{\alpha} - 1\Big)\big(x^{s} - x^{s-1}\big)\Big\| = \frac{q}{\alpha}\,\|x^{s} - x^{s-1}\| \le \frac{q\eta}{\alpha} \qquad (s \ge 1).
\end{equation}

\begin{lemma}[Ideal momentum error under extrapolation]\label{lem:error-extra}
Let Assumptions~\ref{asn:stoch-grad} (at the query points), \ref{asn:norm-equivalence}, and~\ref{asn:hessian} hold. Then, for all $k \ge 0$, the iterates of Algorithm~\ref{alg:byz-muon-extra} satisfy
\[
\mathbb{E}\|\bar{e}^{k}\|_* \;\le\; \frac{q H \eta^2}{2\alpha^2} + \frac{\kappa\sigma}{\sqrt{G}}\, q^{k} + V_G.
\]
\end{lemma}
\begin{proof}
\emph{Step 1 (error recursion).} Averaging the momentum update over $i \in \mathcal{G}$ gives, for $k \ge 1$, $\bar{m}^{k} = q\,\bar{m}^{k-1} + \alpha\big(\nabla f(\bar{x}^{k}) + \bar{\xi}^{k}\big)$, hence
\begin{align*}
\bar{e}^{k} = \bar{m}^{k} - \nabla f(x^{k})
&= q\big(\bar{m}^{k-1} - \nabla f(x^{k-1})\big) + \alpha\,\bar{\xi}^{k} + T^{k}
= q\,\bar{e}^{k-1} + \alpha\,\bar{\xi}^{k} + T^{k},
\end{align*}
where the per-step drift is
\[
T^{k} := \alpha \nabla f(\bar{x}^{k}) + q \nabla f(x^{k-1}) - \nabla f(x^{k}).
\]

\emph{Step 2 (exact first-order cancellation).} Fix $s \ge 1$ and expand both gradients in $T^{s}$ around $x^{s}$ using Lemma~\ref{lem:taylor}:
\begin{align*}
T^{s} &= \alpha\left[\nabla f(x^{s}) + \nabla^2 f(x^{s})(\bar{x}^{s} - x^{s}) + \varrho_1\right] \\
&\qquad + q\left[\nabla f(x^{s}) + \nabla^2 f(x^{s})(x^{s-1} - x^{s}) + \varrho_2\right] - \nabla f(x^{s}) \\
&= (\alpha + q - 1)\,\nabla f(x^{s}) + \nabla^2 f(x^{s})\big[\alpha(\bar{x}^{s} - x^{s}) + q(x^{s-1} - x^{s})\big] + \alpha \varrho_1 + q \varrho_2,
\end{align*}
with $\|\varrho_1\|_* \le \frac{H}{2}\|\bar{x}^{s} - x^{s}\|^2$ and $\|\varrho_2\|_* \le \frac{H}{2}\|x^{s-1} - x^{s}\|^2$. The zeroth-order coefficient vanishes since $\alpha + q = 1$. For the first-order coefficient, the extrapolation rule~\eqref{eq:extrapolation} gives $\alpha\bar{x}^{s} = \alpha x^{s-1} + x^{s} - x^{s-1} = x^{s} - q x^{s-1}$, whence
\[
\alpha(\bar{x}^{s} - x^{s}) + q(x^{s-1} - x^{s}) = \alpha\bar{x}^{s} + q x^{s-1} - x^{s} = 0.
\]
Therefore, using~\eqref{eq:extra-displacement} and $\|x^{s} - x^{s-1}\| \le \eta$,
\[
\|T^{s}\|_* \le \alpha\,\frac{H}{2}\cdot\frac{q^2\eta^2}{\alpha^2} + q\,\frac{H}{2}\,\eta^2
= \frac{H q \eta^2}{2}\cdot\frac{q + \alpha}{\alpha}
= \frac{H q \eta^2}{2\alpha}.
\]

\emph{Step 3 (unrolling).} Since $\bar{e}^{0} = \bar{\xi}^{0}$ by the initialization, unrolling the recursion of Step 1 yields
\[
\bar{e}^{k} = \sum_{s=1}^{k} q^{\,k-s}\, T^{s} + \sum_{s=0}^{k} w_{k,s}\, \bar{\xi}^{s},
\]
where the noise terms carry exactly the weights of Lemma~\ref{lem:momentum-repr}. The drift part is bounded by
\[
\sum_{s=1}^{k} q^{\,k-s}\,\|T^{s}\|_* \le \frac{Hq\eta^2}{2\alpha}\sum_{s=1}^{k} q^{\,k-s} \le \frac{Hq\eta^2}{2\alpha}\cdot\frac{1}{\alpha} = \frac{qH\eta^2}{2\alpha^2},
\]
and the noise part by \eqref{eq:noise-avg} of Lemma~\ref{lem:noise}. Combining the two bounds completes the proof.
\end{proof}

\begin{lemma}[Dispersion and aggregation error under extrapolation]\label{lem:dispersion-extra}
Let Assumptions~\ref{asn:stoch-grad} (at the query points), \ref{asn:lipschitz}, \ref{asn:norm-equivalence}, and~\ref{asn:bounded-heterogeneity} hold, and let the server aggregate satisfy Definition~\ref{def:aragg} with $G \ge \max\{2, (1-\delta)n\}$. Then, with $r_k$ and $S_k$ as in Lemma~\ref{lem:agg_error_rigorous}, the iterates of Algorithm~\ref{alg:byz-muon-extra} satisfy, for every $k \ge 0$,
\[
\mathbb{E}\|m^{k}-\bar m^{k}\|_* \;\le\; c\delta\, r_k \;\le\; c\delta\left[\,2\zeta + 2 B S_k + \frac{2 B q L\eta}{\alpha} + \Big(\sqrt{2}\,\kappa\sigma + 2B\|\nabla f(x^0)\|_*\Big)\, q^{k} + V_P\right].
\]
\end{lemma}
\begin{proof}
Steps 1 and 3 of the proof of Lemma~\ref{lem:agg_error_rigorous} apply verbatim with the representation~\eqref{eq:repr-extra}: the pairwise decomposition and the noise bound do not involve the query points beyond the conditional moment properties. In Step 2, the average pairwise heterogeneity bound is applied pointwise at $x = \bar{x}^{s}$, yielding
\[
\frac{1}{G(G-1)}\sum_{\substack{i,j\in\mathcal{G}\\ i\neq j}}\ \sum_{s=0}^{k} w_{k,s}\, \mathbb{E}\|\nabla f_i(\bar{x}^{s})-\nabla f_j(\bar{x}^{s})\|_*
\le \sum_{s=0}^{k} w_{k,s}\,\big(2\zeta + 2B \bar{g}_s\big),
\]
where $\bar{g}_s := \mathbb{E}\|\nabla f(\bar{x}^{s})\|_*$ and $\bar{g}_0 = g_0 = \|\nabla f(x^0)\|_*$ since $\bar{x}^0 = x^0$. By Assumption~\ref{asn:lipschitz} (inherited by $f$) and \eqref{eq:extra-displacement}, for $s \ge 1$,
\[
\bar{g}_s \le g_s + L\,\mathbb{E}\|\bar{x}^{s} - x^{s}\| \le g_s + \frac{qL\eta}{\alpha}.
\]
Hence
\[
\sum_{s=0}^{k} w_{k,s}\,\big(2\zeta + 2B \bar{g}_s\big)
\le 2\zeta + 2B\Big(q^{k} g_0 + S_k + \frac{qL\eta}{\alpha}\sum_{s=1}^{k}\alpha q^{k-s}\Big)
\le 2\zeta + 2B q^{k} g_0 + 2B S_k + \frac{2BqL\eta}{\alpha},
\]
using $\sum_{s=1}^{k}\alpha q^{k-s} = 1 - q^{k} \le 1$. Combining with Steps 1 and 3 as before and invoking Definition~\ref{def:aragg} completes the proof.
\end{proof}

\begin{theorem}[Non-convex convergence under second-order smoothness]\label{thm:second-order}
Suppose we are given a $(\delta, c)$-robust aggregator satisfying Definition~\ref{def:aragg}, and $n$ workers of which the subset $\mathcal{G}$ of size $G \ge \max\{2,\, n(1-\delta)\}$ faithfully follows the algorithm, with $4c\delta B < 1$. Let Assumptions~\ref{asn:stoch-grad} (at the query points), \ref{asn:lipschitz}, \ref{asn:norm-equivalence}, \ref{asn:bounded-heterogeneity}, and~\ref{asn:hessian} hold, and let $f_{\inf} := \inf_x f(x) > -\infty$. Then the iterates of Algorithm~\ref{alg:byz-muon-extra} satisfy
\begin{align*}
\big(1 - 4 c \delta B\big)\,\frac{1}{K} \sum_{k=0}^{K-1}\mathbb{E}\|\nabla f(x^{k})\|_* \le\ & \frac{f(x^0) - f_{\inf}}{\eta K} + \frac{L\eta}{2} + \frac{q H\eta^2}{\alpha^2} + \frac{4c\delta B\, qL\eta}{\alpha} \\
& + 2 V_G + 2 c\delta V_P + 4c \delta \zeta + \frac{C_0}{\alpha K},
\end{align*}
where $C_0 = \frac{2\kappa\sigma}{\sqrt{G}} + 2\sqrt{2}\,c\delta \kappa\sigma + 4 c\delta B \|\nabla f(x^0)\|_*$ is the constant of Theorem~\ref{thm:nonconvex-apx}. The same bound holds for $\min_{0 \le k < K}\mathbb{E}\|\nabla f(x^k)\|_*$ and for a uniformly sampled iterate.
\end{theorem}
\begin{proof}
Fix $k \in \{0,\dots,K-1\}$. Lemma~\ref{lem:descent_exp_rigorous} applies verbatim to Algorithm~\ref{alg:byz-muon-extra}, and we substitute the momentum-error bound of Lemma~\ref{lem:error-extra} and the aggregation bound of Lemma~\ref{lem:dispersion-extra}:
\begin{align*}
\eta\, g_k
&\le \mathbb{E}[f(x^{k})] - \mathbb{E}[f(x^{k+1})] + \frac{L\eta^2}{2} + 2\eta\left[\frac{qH\eta^2}{2\alpha^2} + \frac{\kappa\sigma}{\sqrt{G}}\, q^{k} + V_G\right] \\
&\quad + 2\eta c\delta\left[2\zeta + 2BS_k + \frac{2BqL\eta}{\alpha} + \Big(\sqrt{2}\kappa\sigma + 2B\|\nabla f(x^0)\|_*\Big) q^{k} + V_P\right].
\end{align*}
We sum over $k = 0, \dots, K-1$ and repeat steps (i)--(iii) of the proof of Theorem~\ref{thm:nonconvex-apx} without change: the telescoping gives $f(x^0) - f_{\inf}$; the geometric transients (whose coefficients are nonnegative and identical to those of Theorem~\ref{thm:nonconvex-apx}) give $\frac{\eta C_0}{\alpha}$; and the history term gives $4\eta c\delta B \sum_{k=0}^{K-1} g_k$, which is moved to the left-hand side, producing the factor $(1 - 4c\delta B)$. The remaining per-step constants are
\[
\frac{L\eta^2}{2} + \frac{qH\eta^3}{\alpha^2} + \frac{4c\delta B qL\eta^2}{\alpha} + 2\eta V_G + 4\eta c\delta \zeta + 2\eta c\delta V_P,
\]
and dividing by $\eta K$ yields the claimed bound.
\end{proof}

\begin{remark}[Improvement, complexity, and the star-convex case]\label{rem:second-order}
Relative to Theorem~\ref{thm:nonconvex-apx}, the first-order drift contribution $\frac{2qL\eta}{\alpha}$ is replaced by $\frac{qH\eta^2}{\alpha^2}$ plus the Byzantine cross term $\frac{4c\delta B qL\eta}{\alpha}$ of the same order in $\eta/\alpha$ (which is smaller than the removed term by the factor $2c\delta B < \nicefrac{1}{2}$ and vanishes when $\delta = 0$ or $B = 0$); the Byzantine floor $4c\delta\zeta + 2c\delta V_P$ and the transient $C_0/(\alpha K)$ are unchanged, so extrapolation improves the optimization terms but not the adversarial ones. At $\delta = 0$, taking the aggregator to be the plain average (so that $m^{k} = \bar{m}^{k}$ exactly and Lemma~\ref{lem:dispersion-extra} is not needed), the resulting bound --- then valid even for a single worker, $G = 1$ --- matches the structure of Theorem~5 of \citet{kovalev2025understanding}, and tuning $\alpha = \Theta\big(\min\{1, G(\varepsilon/\kappa\sigma)^2\}\big)$ and $\eta = \Theta\big(\min\{\varepsilon/L,\ \alpha\sqrt{\varepsilon/H}\}\big)$ yields the improved iteration complexity $K = \mathcal{O}(\varepsilon^{-3.5})$ for reaching $\frac{1}{K}\sum_k \mathbb{E}\|\nabla f(x^k)\|_* \le \varepsilon$ (up to problem constants), compared to $\mathcal{O}(\varepsilon^{-4})$ under first-order smoothness \citep{cutkosky2020momentum, kovalev2025understanding}. The extrapolation also composes with weight decay exactly as in Algorithm~3 of \citet{kovalev2025understanding}: since the per-step cancellation in Lemma~\ref{lem:error-extra} is purely algebraic, the star-convex analysis of Appendix~\ref{apx:starconvex-proof} carries over with the step bound $2\eta$ (Lemma~\ref{lem:aux-inequalities}), replacing the term $\frac{4L\eta^2}{\alpha}$ in $C_\star$ by $4L\eta^2 + \frac{4qH\eta^3}{\alpha^2}$ and adding the cross term $\frac{8c\delta B q L \eta^2}{\alpha}$; as in the centralized case \citep{kovalev2025understanding}, this does not improve the dominant $\mathcal{O}(L\eta^2/\beta)$ neighborhood term, so we do not state it as a separate theorem.
\end{remark}

\section{Extensions to Inexact Muon}\label{apx:inexact}

The results of Appendices~\ref{apx:nonconvex-proof}--\ref{apx:second-order} analyze an idealized method in which the trust-region linear subproblem is solved \emph{exactly} --- for the Muon geometry, this requires an exact polar factor, which can be characterized by an SVD. In practice, Muon computes the orthogonalized direction approximately, by a small fixed number of Newton--Schulz iterations \citep{jordan6muon} or refined schemes such as PolarExpress \citep{amsel2025polar}. Following the inexactness framework of \citet{shulgin2025beyond}, who give the first analysis of the inexact (non-Byzantine) Muon update under a realistic additive LMO error model, we extend \emph{all} results of this paper to the inexact setting. Throughout this appendix, an unadorned $\delta$ continues to denote the Byzantine fraction of Definition~\ref{def:aragg}; the LMO inexactness is always written $\delta_k$ or $\bar{\delta}$ (\citet{shulgin2025beyond} write $\delta$ for the latter). We impose the following oracle model.

\begin{assumption}[Inexact LMO; \citealp{shulgin2025beyond}]\label{asn:inexact}
At every iteration $k$, instead of solving the trust-region subproblem exactly, the server updates
\begin{equation}\label{eq:inexact-update}
x^{k+1} = c^{k} + \eta\, \hat{d}^{\,k}, \qquad c^{k} := (1-\beta)x^{k} \quad (\text{with } \beta = 0 \text{ in the absence of weight decay}),
\end{equation}
where $\hat{d}^{\,k}$ satisfies $\|\hat{d}^{\,k} - d^{k}\| \le \delta_k$ for some exact solution $d^{k} \in \arg\min_{\|d\| \le 1} \langle m^{k}, d\rangle$ and inexactness levels $\delta_k \in [0, \bar{\delta}\,]$ with $\bar{\delta} < 1$. The output need not be feasible: only $\|\hat{d}^{\,k}\| \le 1 + \delta_k$ is guaranteed.
\end{assumption}
Approximation schemes can be analyzed under Assumption~\ref{asn:inexact} when an appropriate error bound is established \citep[Theorem~4.3]{amsel2025polar}; see \citet[Appendix~A]{shulgin2025beyond}. The five-step Newton--Schulz implementation in our experiments is not certified here to satisfy a uniform $\bar{\delta}<1$. At $\bar{\delta} = 0$ the update \eqref{eq:inexact-update} coincides with the exact trust-region step, and every result below reduces to its exact counterpart. As in \citet{shulgin2025beyond}, the condition $\bar{\delta} < 1$ is sufficient for uniform progress over the permitted oracle-error class, but is not necessary for each realized direction to descend. The filtration convention is unchanged: $\hat{d}^{\,k}$ is computed by the server from $m^k$, with any internal randomness of the approximation scheme included in $\mathcal{F}^{k}$ and assumed independent of the round-$(k+1)$ stochastic-gradient draws (vacuous for the deterministic Newton--Schulz and PolarExpress iterations), so $x^{k+1}$ remains $\mathcal{F}^{k}$-measurable and Assumption~\ref{asn:stoch-grad} and Lemma~\ref{lem:noise} are unaffected.

\begin{lemma}[Inexact LMO step]\label{lem:inexact-step}
Under Assumption~\ref{asn:inexact}, for every $k \ge 0$:
\emph{(i)} $\langle m^{k}, x^{k+1} - c^{k}\rangle \le -(1-\delta_k)\,\eta\,\|m^{k}\|_*$;
\emph{(ii)} $\|x^{k+1} - c^{k}\| \le (1+\delta_k)\,\eta$;
\emph{(iii)} for every $y$ with $\|y - c^{k}\| \le \eta$, $\;\langle m^{k}, x^{k+1}\rangle \le \langle m^{k}, y\rangle + \delta_k\,\eta\,\|m^{k}\|_*$.
\end{lemma}
\begin{proof}
(i) By Lemma~\ref{lem:tr_opt}, $\langle m^k, d^k\rangle = -\|m^k\|_*$, and by H\"older's inequality $\langle m^k, \hat{d}^{\,k} - d^k\rangle \le \|m^k\|_* \|\hat{d}^{\,k} - d^k\| \le \delta_k \|m^k\|_*$; hence $\langle m^k, x^{k+1} - c^k\rangle = \eta\langle m^k, \hat{d}^{\,k}\rangle \le -(1-\delta_k)\eta\|m^k\|_*$.
(ii) $\|\hat{d}^{\,k}\| \le \|d^k\| + \delta_k \le 1 + \delta_k$.
(iii) By H\"older, $\langle m^k, y - c^k\rangle \ge -\eta\|m^k\|_*$; subtracting this from (i) gives $\langle m^k, x^{k+1} - c^k\rangle - \langle m^k, y - c^k\rangle \le \delta_k\eta\|m^k\|_*$.
\end{proof}

\paragraph{Carrying over the momentum lemmas.}
The momentum recursion, its initialization, and the aggregation step are untouched by Assumption~\ref{asn:inexact}, so Lemmas~\ref{lem:momentum-repr}, \ref{lem:noise}, \ref{lem:agg_error_rigorous}, and~\ref{lem:dispersion2} (with Corollary~\ref{cor:aggerr2}) apply verbatim. Lemma~\ref{lem:error_recursion_final} applies with the enlarged step bound: by Lemma~\ref{lem:inexact-step}(ii), $\|x^{s} - x^{s-1}\| \le (1+\bar{\delta})\eta$ when $\beta = 0$, and $\|x^{s} - x^{s-1}\| \le 2(1+\bar{\delta})\eta$ under weight decay (Lemma~\ref{lem:inexact-geometry} below), so its conclusion holds with $d = (1+\bar{\delta})\eta$ and $d = 2(1+\bar{\delta})\eta$, respectively. The integrability convention of Lemma~\ref{lem:descent_exp_rigorous} holds with $\|x^{k} - x^{0}\| \le k(1+\bar{\delta})\eta$.

\subsection{Non-convex case}\label{apx:inexact-nonconvex}

\begin{lemma}[Inexact descent bound]\label{lem:inexact-descent}
Let Assumptions~\ref{asn:lipschitz} and~\ref{asn:inexact} hold with $\beta = 0$. Then, for every $k \ge 0$,
\[
f(x^{k+1}) \le f(x^{k}) - (1-\bar{\delta})\,\eta\,\|\nabla f(x^{k})\|_* + 2\,\eta\,\|\nabla f(x^{k}) - m^{k}\|_* + \frac{L(1+\bar{\delta})^2\eta^2}{2}.
\]
Notably, the coefficient of the momentum-error term is $2\eta$, exactly as in the exact case (Lemma~\ref{lem:descent_tr_rigorous}): inexactness deflates the progress term and inflates the quadratic term, but does not amplify the momentum error.
\end{lemma}
\begin{proof}
Smoothness and Lemma~\ref{lem:inexact-step}(ii) give $f(x^{k+1}) \le f(x^k) + \langle \nabla f(x^k), x^{k+1}-x^k\rangle + \frac{L(1+\bar{\delta})^2\eta^2}{2}$. Decomposing and applying Lemma~\ref{lem:inexact-step}(i) (with $c^k = x^k$) and H\"older,
\[
\langle \nabla f(x^k), x^{k+1}-x^k\rangle \le -(1-\delta_k)\eta\|m^k\|_* + (1+\delta_k)\eta\,\|\nabla f(x^k)-m^k\|_*.
\]
The reverse triangle inequality yields $-(1-\delta_k)\eta\|m^k\|_* \le -(1-\delta_k)\eta\|\nabla f(x^k)\|_* + (1-\delta_k)\eta\|\nabla f(x^k)-m^k\|_*$, and collecting terms proves the claim: the negative term uses $1-\delta_k \ge 1-\bar{\delta}$, while the two error contributions sum to exactly $\big[(1-\delta_k) + (1+\delta_k)\big]\eta = 2\eta$.
\end{proof}

\begin{theorem}[Inexact non-convex convergence]\label{thm:inexact-nonconvex}
Let the assumptions of Theorem~\ref{thm:nonconvex-apx} hold together with Assumption~\ref{asn:inexact}, and let
\[
4 c\delta B < 1-\bar{\delta}.
\]
Then the iterates of the inexact update \eqref{eq:inexact-update} (with $\beta = 0$) satisfy
\begin{align*}
\big[(1-\bar{\delta}) - 4 c \delta B\big]\, \frac{1}{K}\sum_{k=0}^{K-1}\mathbb{E}\|\nabla f(x^{k})\|_*
\le\ & \frac{f(x^0)-f_{\inf}}{\eta K} + \frac{L(1+\bar{\delta})^2\eta}{2} + \frac{2(1+\bar{\delta})\, q L\eta}{\alpha} \\
& + 2 V_G + 4c\delta\zeta + 2 c\delta V_P + \frac{C_{0}}{\alpha K},
\end{align*}
where $C_{0}$ is the constant of Theorem~\ref{thm:nonconvex-apx}. At $\bar{\delta} = 0$ this is exactly Theorem~\ref{thm:nonconvex-apx}; in particular, the Byzantine and transient terms in the numerator are unchanged, although the smaller progress denominator can enlarge their contribution to the resulting stationarity bound.
\end{theorem}
\begin{proof}
Take total expectations of Lemma~\ref{lem:inexact-descent} and decompose $\|\nabla f(x^k)-m^k\|_* \le \|\bar{e}^k\|_* + \|m^k - \bar{m}^k\|_*$ as in Lemma~\ref{lem:descent_exp_rigorous}. Substituting Lemma~\ref{lem:error_recursion_final} with $d = (1+\bar{\delta})\eta$ and Lemma~\ref{lem:agg_error_rigorous},
\begin{align*}
(1-\bar{\delta})\,\eta\, g_k \le\ & \mathbb{E}[f(x^k)] - \mathbb{E}[f(x^{k+1})] + \frac{L(1+\bar{\delta})^2\eta^2}{2}
+ 2\eta\left[\frac{q L (1+\bar{\delta})\eta}{\alpha} + \frac{\kappa\sigma}{\sqrt{G}} q^k + V_G\right] \\
& + 2\eta\, c\delta\left[2\zeta + 2BS_k + \big(\sqrt{2}\kappa\sigma + 2B\|\nabla f(x^0)\|_*\big)q^k + V_P\right].
\end{align*}
Summing over $k = 0,\dots,K-1$ and repeating steps (i)--(iii) of the proof of Theorem~\ref{thm:nonconvex-apx} verbatim --- telescoping, $\sum_k q^k \le 1/\alpha$ on the nonnegative $q^k$-coefficients (which assemble to $\eta C_{0}/\alpha$, exactly as there), and $\sum_k S_k \le \sum_k g_k$ --- the history contributes $4c\delta B\,\eta \sum_k g_k$, which is moved to the left-hand side to produce the factor $(1-\bar{\delta}) - 4c\delta B$; dividing by $\eta K$ yields the claim.
\end{proof}

\subsection{Star-convex case}\label{apx:inexact-star}

\begin{lemma}[Inexact weight-decay geometry]\label{lem:inexact-geometry}
Let $0 < \beta < 1$, $\eta \ge \beta\max\{\|x^0\|, \|x^*\|\}$, and Assumption~\ref{asn:inexact} hold, and define $y^{k} := \beta x^* + (1-\beta)x^{k}$. Then, for all $k \ge 0$: $\beta\|x^{k}\| \le (1+\bar{\delta})\eta$; $\|y^{k} - (1-\beta)x^{k}\| \le \eta$ (so $y^k$ is feasible for the exact ball); $\|y^{k} - x^{k}\| \le (2+\bar{\delta})\eta$; $\|y^{k} - x^{k+1}\| \le (2+\bar{\delta})\eta$; and $\|x^{k+1} - x^{k}\| \le 2(1+\bar{\delta})\eta$.
\end{lemma}
\begin{proof}
Induction for the first claim: $\beta\|x^0\| \le \eta \le (1+\bar{\delta})\eta$, and, using Lemma~\ref{lem:inexact-step}(ii), $\beta\|x^{k+1}\| \le \beta\|x^{k+1} - (1-\beta)x^k\| + (1-\beta)\beta\|x^k\| \le \beta(1+\bar{\delta})\eta + (1-\beta)(1+\bar{\delta})\eta = (1+\bar{\delta})\eta$. Then $\|y^k - (1-\beta)x^k\| = \beta\|x^*\| \le \eta$; $\|y^k - x^k\| \le \beta\|x^*\| + \beta\|x^k\| \le (2+\bar{\delta})\eta$; $\|y^k - x^{k+1}\| \le \|y^k - (1-\beta)x^k\| + \|x^{k+1} - (1-\beta)x^k\| \le \eta + (1+\bar{\delta})\eta = (2+\bar{\delta})\eta$; and $\|x^{k+1} - x^k\| \le \|x^{k+1} - (1-\beta)x^k\| + \beta\|x^k\| \le 2(1+\bar{\delta})\eta$.
\end{proof}

\begin{lemma}[Inexact star-convex descent bound]\label{lem:inexact-scvx}
Let Assumptions~\ref{asn:lipschitz}, \ref{asn:star-cvx}, and~\ref{asn:inexact} hold, with $0 < \beta < 1$, $\eta \ge \beta\max\{\|x^0\|,\|x^*\|\}$. Then, for every $k \ge 0$,
\[
f(x^{k+1}) \le \beta f(x^*) + (1-\beta) f(x^{k}) + 4L(1+\bar{\delta})^2\eta^2 + 2(1+\bar{\delta})\,\eta\,\|\nabla f(x^{k}) - m^{k}\|_* + \bar{\delta}\,\eta\,\|\nabla f(x^{k})\|_*.
\]
\end{lemma}
\begin{proof}
Smoothness and $\|x^{k+1}-x^k\| \le 2(1+\bar{\delta})\eta$ (Lemma~\ref{lem:inexact-geometry}) give the quadratic term $2L(1+\bar{\delta})^2\eta^2$. Since $y^k$ is feasible for the exact ball, Lemma~\ref{lem:inexact-step}(iii) gives $\langle m^k, x^{k+1} - x^k\rangle \le \langle m^k, y^k - x^k\rangle + \delta_k\eta\|m^k\|_*$, and the regrouping of the proof of Lemma~\ref{lem:scvx-descent} yields
\[
\langle \nabla f(x^k), x^{k+1}-x^k\rangle \le \langle \nabla f(x^k), y^k - x^k\rangle + \langle m^k - \nabla f(x^k),\, y^k - x^{k+1}\rangle + \bar{\delta}\eta\|m^k\|_*.
\]
By H\"older and Lemma~\ref{lem:inexact-geometry}, the middle term is at most $(2+\bar{\delta})\eta\|\nabla f(x^k)-m^k\|_*$, and $\|m^k\|_* \le \|\nabla f(x^k)\|_* + \|\nabla f(x^k)-m^k\|_*$, so the error terms total $\big[(2+\bar{\delta}) + \bar{\delta}\big]\eta\,\|\nabla f(x^k)-m^k\|_* + \bar{\delta}\eta\|\nabla f(x^k)\|_* = 2(1+\bar{\delta})\eta\,\|\nabla f(x^k)-m^k\|_* + \bar{\delta}\eta\|\nabla f(x^k)\|_*$. The lower Taylor bound with $\|y^k - x^k\| \le (2+\bar{\delta})\eta \le 2(1+\bar{\delta})\eta$ gives $f(x^k) + \langle\nabla f(x^k), y^k - x^k\rangle \le f(y^k) + 2L(1+\bar{\delta})^2\eta^2$, and star-convexity at $y^k$ concludes.
\end{proof}

\begin{theorem}[Inexact star-convex convergence]\label{thm:inexact-starconvex}
Let the assumptions of Theorem~\ref{thm:starconvex-apx} hold together with Assumption~\ref{asn:inexact}. Define
\[
K_{B,\bar{\delta}} := \frac{4(1+\bar{\delta})\,c\delta\eta\,\alpha B(1-\beta)}{\alpha-\beta}, \quad
D_{\bar{\delta}} := 2(1+\bar{\delta})\,\eta\left(\frac{\kappa\sigma}{\sqrt{G}} + \sqrt{2}\,c\delta\kappa\sigma + 2c\delta B\|\nabla f(x^0)\|_*\right),
\]
\[
C^{\star}_{\bar{\delta}} := \frac{4L(1+\bar{\delta})^2\eta^2}{\alpha} + 2(1+\bar{\delta})\,\eta\big(V_G + 2c\delta\zeta + c\delta V_P\big).
\]
Then, for every $K \ge 0$, the iterates of the inexact update \eqref{eq:inexact-update} satisfy
\[
\Delta_K \le \Big(1-\frac{\beta}{2}\Big)^{K}\left(\Delta_0 + \frac{D_{\bar{\delta}}}{\alpha-\beta/2}\right) + \frac{2}{\beta}\left(C^{\star}_{\bar{\delta}} + \frac{L\big(K_{B,\bar{\delta}} + \bar{\delta}\eta\big)^2}{\beta}\right).
\]
At $\bar{\delta} = 0$ this is exactly Theorem~\ref{thm:starconvex-apx}.
\end{theorem}
\begin{proof}
Taking expectations of Lemma~\ref{lem:inexact-scvx}, decomposing the momentum error, and substituting Lemma~\ref{lem:error_recursion_final} with $d = 2(1+\bar{\delta})\eta$ and Lemma~\ref{lem:agg_error_rigorous}, the one-step recursion becomes (using $4L(1+\bar{\delta})^2\eta^2 + 2(1+\bar{\delta})\eta \cdot \frac{2q L(1+\bar{\delta})\eta}{\alpha} = \frac{4L(1+\bar{\delta})^2\eta^2}{\alpha}$, since $\alpha + q = 1$)
\[
\Delta_{k+1} \le (1-\beta)\Delta_k + C^{\star}_{\bar{\delta}} + D_{\bar{\delta}}\, q^k + 4(1+\bar{\delta})c\delta\eta B\, S_k + \bar{\delta}\eta\, g_k.
\]
With $\Gamma_{\bar{\delta}} := \frac{4(1+\bar{\delta})c\delta\eta B q}{\alpha-\beta}$ and $\Lambda_{\bar{\delta}} := 4(1+\bar{\delta})c\delta\eta B + \Gamma_{\bar{\delta}} = \frac{4(1+\bar{\delta})c\delta\eta B(1-\beta)}{\alpha-\beta}$, the identities $q\Lambda_{\bar{\delta}} = (1-\beta)\Gamma_{\bar{\delta}}$ and $\alpha\Lambda_{\bar{\delta}} = K_{B,\bar{\delta}}$ hold exactly as in \eqref{eq:lyap-identities}, so with $\Psi_k := \Delta_k + \Gamma_{\bar{\delta}} S_{k-1}$ ($\Psi_0 = \Delta_0$), Step~3 of the proof of Theorem~\ref{thm:starconvex-apx} gives, for all $k \ge 1$,
\[
\Psi_{k+1} \le (1-\beta)\Psi_k + \big(K_{B,\bar{\delta}} + \bar{\delta}\eta\big) g_k + C^{\star}_{\bar{\delta}} + D_{\bar{\delta}}\, q^k.
\]
The same recursion holds at $k = 0$: since $S_0 = 0$, the one-step bound reads $\Psi_1 = \Delta_1 \le (1-\beta)\Psi_0 + \bar{\delta}\eta\, g_0 + C^{\star}_{\bar{\delta}} + D_{\bar{\delta}}$, and $\bar{\delta}\eta\, g_0 \le (K_{B,\bar{\delta}} + \bar{\delta}\eta)\, g_0$.
By Lemma~\ref{lem:grad-gap} and Young's inequality, $\big(K_{B,\bar{\delta}}+\bar{\delta}\eta\big) g_k \le \frac{\beta}{2}\Delta_k + \frac{L(K_{B,\bar{\delta}}+\bar{\delta}\eta)^2}{\beta} \le \frac{\beta}{2}\Psi_k + \frac{L(K_{B,\bar{\delta}}+\bar{\delta}\eta)^2}{\beta}$, and unrolling as in Step~4 of that proof completes the argument.
\end{proof}

\subsection{Second-moment aggregators and second-order smoothness}\label{apx:inexact-second}

\begin{theorem}[Inexact convergence with second-moment aggregators]\label{thm:inexact-second-moment}
Let Assumption~\ref{asn:inexact} hold in addition to the assumptions of Theorem~\ref{thm:second-moment} (resp.\ Corollary~\ref{cor:starconvex2}). Then:
\emph{(i)} with $\beta = 0$,
\begin{align*}
(1-\bar{\delta})\,\frac{1}{K}\sum_{k=0}^{K-1}\mathbb{E}\|\nabla f(x^{k})\|_*
\le\ & \frac{f(x^0)-f_{\inf}}{\eta K} + \frac{L(1+\bar{\delta})^2\eta}{2} + \frac{2(1+\bar{\delta})qL\eta}{\alpha} \\
& + 2 V_G + \sqrt{c\delta}\big(4\sqrt{2}\,\kappa\zeta_2 + 2\sqrt{2}\,V_P\big) + \frac{\bar{C}_{0}}{\alpha K},
\end{align*}
with $\bar{C}_{0}$ the constant of Theorem~\ref{thm:second-moment}, and no condition on $\delta$ beyond $\delta < \nicefrac{1}{2}$ (only $\bar{\delta} < 1$);
\emph{(ii)} under the star-convex conditions of Corollary~\ref{cor:starconvex2},
\[
\Delta_K \le \Big(1-\frac{\beta}{2}\Big)^{K}\left(\Delta_0 + \frac{\bar{D}_{\bar{\delta}}}{\alpha-\beta/2}\right) + \frac{2}{\beta}\left(\bar{C}_{4,\bar{\delta}} + \frac{L\bar{\delta}^2\eta^2}{\beta}\right),
\]
with $\bar{D}_{\bar{\delta}} := (1+\bar{\delta})\,\eta\bar{C}_{0} = 2(1+\bar{\delta})\eta\kappa\sigma\big(\tfrac{1}{\sqrt{G}} + 2\sqrt{c\delta}\big)$ and $\bar{C}_{4,\bar{\delta}} := \frac{4L(1+\bar{\delta})^2\eta^2}{\alpha} + 2(1+\bar{\delta})\eta\big[V_G + \sqrt{c\delta}\big(2\sqrt{2}\kappa\zeta_2 + \sqrt{2} V_P\big)\big]$.
At $\bar{\delta} = 0$ these are exactly Theorem~\ref{thm:second-moment} and Corollary~\ref{cor:starconvex2}.
\end{theorem}
\begin{proof}
Identical to the proofs of Theorem~\ref{thm:inexact-nonconvex} and Theorem~\ref{thm:inexact-starconvex}, with the aggregation bound of Lemma~\ref{lem:agg_error_rigorous} replaced throughout by Corollary~\ref{cor:aggerr2}, whose constant part is $\sqrt{c\delta}(2\sqrt{2}\kappa\zeta_2 + \sqrt{2}V_P)$ and transient part $2\kappa\sigma\sqrt{c\delta}\,q^k$; since Assumption~\ref{asn:het2} has no relative component, no gradient-history term arises in (i) (nothing is moved to the left-hand side, so the prefactor is $(1-\bar{\delta})$), and in (ii) the Lyapunov correction vanishes ($\Gamma_{\bar{\delta}} = K_{B,\bar{\delta}} = 0$, so $\alpha > \beta/2$ suffices), leaving only the Young absorption of the direct term $\bar{\delta}\eta g_k$, which contributes $\frac{L\bar{\delta}^2\eta^2}{\beta}$.
\end{proof}

\begin{theorem}[Inexact second-order convergence]\label{thm:inexact-second-order}
Let Assumption~\ref{asn:inexact} hold in addition to the assumptions of Theorem~\ref{thm:second-order}, with the extrapolated point computed from the actual (inexact) iterates as in \eqref{eq:extrapolation}, and let $4c\delta B < 1-\bar{\delta}$. Then
\begin{align*}
\big[(1-\bar{\delta}) - 4 c\delta B\big]\, \frac{1}{K}\sum_{k=0}^{K-1}\mathbb{E}\|\nabla f(x^{k})\|_*
\le\ & \frac{f(x^0)-f_{\inf}}{\eta K} + \frac{L(1+\bar{\delta})^2\eta}{2} + \frac{(1+\bar{\delta})^2 q H\eta^2}{\alpha^2} \\
& + \frac{4(1+\bar{\delta})\, c\delta B\, qL\eta}{\alpha}
+ 2 V_G + 4c\delta\zeta + 2 c\delta V_P + \frac{C_{0}}{\alpha K},
\end{align*}
with $C_{0}$ the constant of Theorem~\ref{thm:nonconvex-apx}. At $\bar{\delta} = 0$ this is exactly Theorem~\ref{thm:second-order}.
\end{theorem}
\begin{proof}
The extrapolation identity $\alpha\bar{x}^{s} + q x^{s-1} - x^{s} = 0$ is pure algebra on the actual iterates and is unaffected by inexactness, so the first-order cancellation in the proof of Lemma~\ref{lem:error-extra} holds verbatim. The step bound becomes $\|x^{s}-x^{s-1}\| \le (1+\bar{\delta})\eta$ (Lemma~\ref{lem:inexact-step}(ii) with $\beta = 0$), hence $\|\bar{x}^{s} - x^{s}\| \le \frac{q(1+\bar{\delta})\eta}{\alpha}$ by \eqref{eq:extra-displacement}, the per-step remainder becomes $\|T^{s}\|_* \le \frac{Hq(1+\bar{\delta})^2\eta^2}{2\alpha}$, and the unrolled drift $\frac{qH(1+\bar{\delta})^2\eta^2}{2\alpha^2}$; likewise the conversion term of Lemma~\ref{lem:dispersion-extra} becomes $\frac{2BqL(1+\bar{\delta})\eta}{\alpha}$. Substituting these bounds into the inexact descent Lemma~\ref{lem:inexact-descent} (whose derivation nowhere uses the location of the gradient queries) and repeating the summation of Theorem~\ref{thm:second-order} with the error coefficient $2\eta$ and progress coefficient $(1-\bar{\delta})\eta$ yields the claim, the history term being extracted exactly as in Theorem~\ref{thm:inexact-nonconvex}.
\end{proof}

\begin{remark}[Recovery, degradation, and parameter coupling]\label{rem:inexact}
All four results reduce to their exact counterparts at $\bar{\delta} = 0$, so the extension is strict, mirroring \citet{shulgin2025beyond}. The degradation pattern also mirrors theirs: progress is deflated by $(1-\bar{\delta})$ while the curvature and drift terms are inflated by powers of $(1+\bar{\delta})$, so their step-size/momentum coupling --- in the cited leading-order stochastic tuning, lower oracle precision favors a smaller radius $\eta$ and a larger new-gradient weight $\alpha$ (shorter momentum memory) --- is consistent with our bound, without prescribing a universal Byzantine optimum. Two effects are specific to our setting. First, in the non-convex results the Byzantine and transient terms in the numerator are unchanged by inexactness (the momentum-error coefficient in Lemma~\ref{lem:inexact-descent} is exactly $2\eta$, as in the exact case), and in the star-convex results the transient constant $D_{\bar{\delta}}$ and the noise/heterogeneity components of $C^{\star}_{\bar{\delta}}$ are inflated only by the bounded factor $(1+\bar{\delta}) < 2$, while the curvature component $4L(1+\bar{\delta})^2\eta^2/\alpha$ of $C^{\star}_{\bar{\delta}}$ is inflated by $(1+\bar{\delta})^2 < 4$, and the quadratic neighborhood term expands as $L(K_{B,\bar{\delta}} + \bar{\delta}\eta)^2/\beta = L\big((1+\bar{\delta})^2 K_B^2 + 2(1+\bar{\delta})K_B\,\bar{\delta}\eta + \bar{\delta}^2\eta^2\big)/\beta$ (with $K_{B,\bar{\delta}} = (1+\bar{\delta})K_B$): a $(1+\bar{\delta})^2 < 4$ inflation of the exact term $LK_B^2/\beta$ plus cross and additive terms that vanish with $\bar{\delta}\eta$, the additive $L\bar{\delta}^2\eta^2/\beta$ remainder being present even at $K_B = 0$ (cf.\ Theorem~\ref{thm:inexact-second-moment}(ii)) --- these star-convex numerator factors are bounded, but the non-convex bound can deteriorate arbitrarily near the threshold $1-\bar{\delta}-4c\delta B=0$. Second, the tolerable relative heterogeneity shrinks: the condition $4c\delta B < 1$ of Theorem~\ref{thm:nonconvex-apx} tightens to $4c\delta B < 1 - \bar{\delta}$, i.e., a less precise oracle tolerates strictly less relative heterogeneity, while the second-moment route (Theorem~\ref{thm:inexact-second-moment}) remains free of any such condition. Finally, when an approximation scheme admits a geometric error certificate in the number of inner iterations \citep{amsel2025polar}, its certified $\bar{\delta}$-degradations can be reduced at logarithmic extra cost per round; this statement does not certify the fixed five-step implementation.
\end{remark}

\end{document}